\documentclass[10pt,twocolumn,letterpaper]{article}

\usepackage{iccv}
\usepackage{times}
\usepackage{epsfig}
\usepackage{graphicx}
\usepackage{amsmath}
\usepackage{amssymb}
\usepackage{multirow}
\usepackage{subfigure}
\usepackage{authblk}
\newcommand{\wenhang}{\textcolor[rgb]{0,0,0}}
\newcommand{\yc}{\textcolor[rgb]{0,0,0}}

\usepackage[breaklinks=true,bookmarks=false]{hyperref}

\iccvfinalcopy 

\def\iccvPaperID{9385} 

\ificcvfinal\fi

\begin{document}

\title{Ref-NeuS: Ambiguity-Reduced Neural Implicit Surface Learning for Multi-View Reconstruction with Reflection}

\author{
 Wenhang Ge$^{1, 2}$ \quad Tao Hu$^4$ \quad  Haoyu Zhao$^1$ \quad Shu Liu$^5$ \quad Ying-Cong Chen$^{1, 2, 3,}$\thanks{Corresponding author.}\\
 $^1$HKUST(GZ) \qquad
 $^2$HKUST(GZ)-SmartMore Joint Lab \qquad
 $^3$HKUST \qquad
 $^4$CUHK \qquad
 $^5$SmartMore\\
 }

\maketitle
\ificcvfinal\thispagestyle{empty}\fi

\begin{abstract}
   Neural implicit surface learning has shown significant progress in multi-view 3D reconstruction, where an object is represented by multilayer perceptrons that provide continuous implicit surface representation and view-dependent radiance.
   However, current methods often fail to accurately reconstruct reflective surfaces, leading to severe ambiguity. 
   To overcome this issue, we propose Ref-NeuS, which aims to reduce ambiguity by attenuating the effect of reflective surfaces.
   Specifically, we utilize an anomaly detector to estimate an explicit reflection score with the guidance of multi-view context to localize reflective surfaces.
    Afterward, we design a reflection-aware photometric loss that adaptively reduces ambiguity by modeling rendered color as a Gaussian distribution, with the reflection score representing the variance. 
    We show that together with a reflection direction-dependent radiance, our model achieves high-quality surface reconstruction on reflective surfaces and outperforms the state-of-the-arts by a large margin. Besides, our model is also comparable on general surfaces.
    
\end{abstract}

\section{Introduction}

\begin{figure}[t]

\centering
\includegraphics[width=\columnwidth]{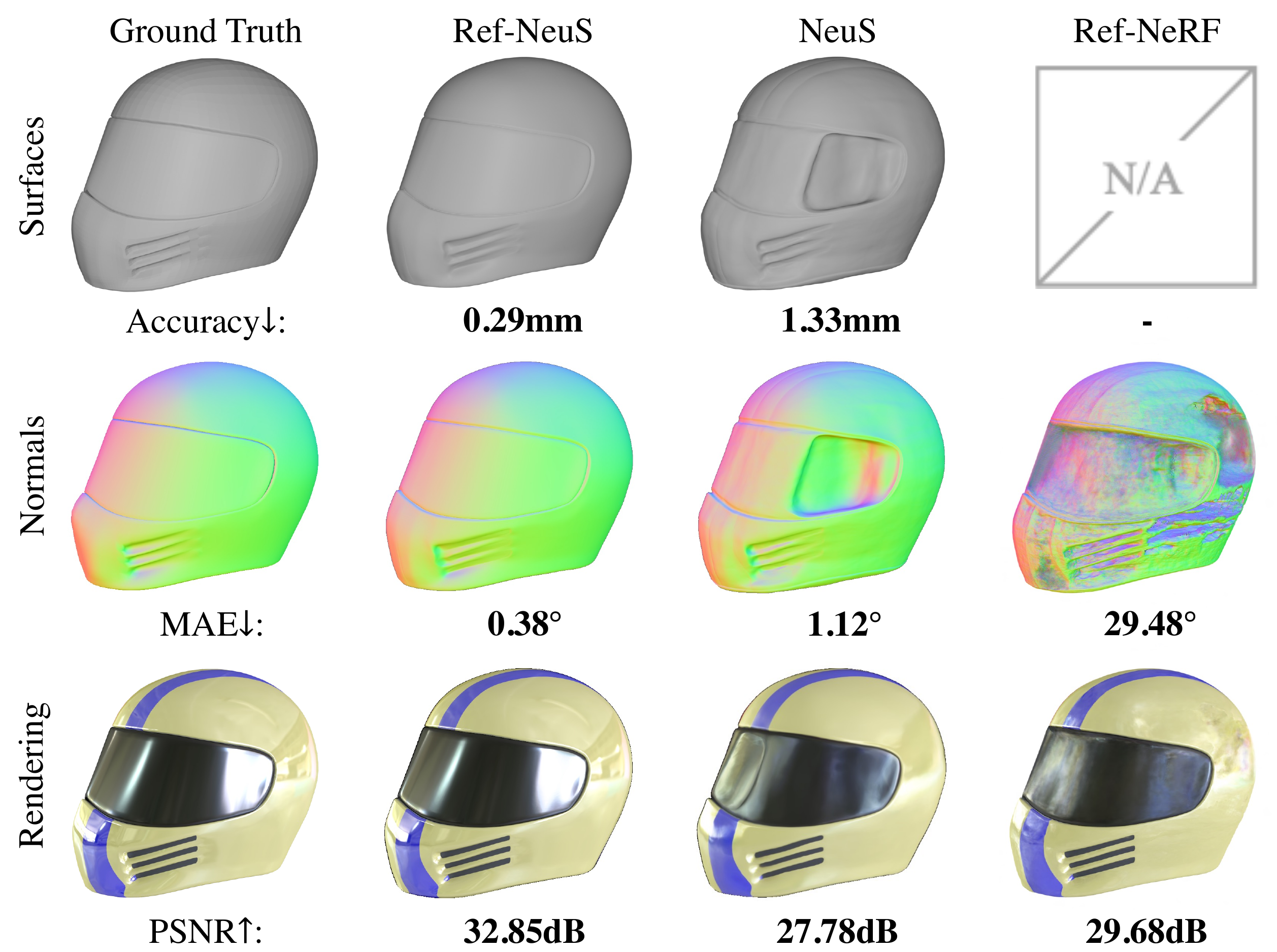}

\caption{Our ambiguity-reduced framework significantly improves explicit surfaces geometry, surface normals and rendering realism for modeling reflective surfaces compared to NeuS and Ref-NeRF. Reconstruction accuracy  (lower is better), surface normal mean angular error, i.e., MAE (lower is better) and PSNR (higher is better) are compared.}

\label{demo}
\end{figure}

3D reconstruction is a crucial task in computer vision that serves as the foundation for multiple fields such as computer-aided design \cite{kennard1969computer, falivene2019towards}, computer animation \cite{parent2012computer, lasseter1987principles}, and virtual reality \cite{schuemie2001research}. 
Among various 3D reconstruction techniques, image-based 3D reconstruction is particularly challenging, which aims to recover 3D structures from posed 2D images. 
Traditional multi-view stereo (MVS) approaches \cite{furukawa2009accurate, schonberger2016pixelwise, yao2020blendedmvs} generally require a multi-step pipeline with supervision, which can be cumbersome.  
Recently, neural implicit surface learning \cite{wang2021neus, volsdf, oechsle2021unisurf} has gained increasing attention due to its ability to achieve remarkable reconstruction quality with a neat formulation that supports end-to-end and unsupervised training.
However, as illustrated in Fig. \ref{demo}, 
existing methods tend to produce erroneous results in reflective surfaces. Since these approaches infer geometry information with multi-view consistency, which is compromised due to the ambiguous surface prediction on reflective surfaces. As a result, their practicality becomes limited in scenarios where reflection is unavoidable.

Several recent works \cite{boss2021nerd, srinivasan2021nerv, zhang2021physg, zhang2021nerfactor, verbin2022ref}  have investigated reflection modeling in neural radiance fields. 
These methods typically decompose the appearance of an object into several physical components, allowing reflection to be explicitly represented. 
This can lead to a better estimation of the 3D geometry by eliminating the influence of the reflection component. Nevertheless, physical decomposition can be highly ill-posed \cite{guo2014robust}, and inaccurate decomposition can significantly constrain performance. For example, as shown in Fig. \ref{demo}, the predicted normals in Ref-NeRF \cite{verbin2022ref} are not accurate enough, which leads to suboptimal performance.

In this paper, we propose a simple yet effective solution that does not rely on challenging physical decomposition. Instead, we suggest reducing ambiguity by introducing a reflection-aware photometric loss that adaptively lowers the weights fitting reflective surfaces based on the reflection score. By doing so, we avoid devastating multi-view consistency.
\wenhang{Additionally, inspired by Ref-NeRF \cite{verbin2022ref} and NeuralWarp \cite{darmon2022improving}, we show that we can further improve the geometry by substituting radiance dependency with reflection direction to obtain a more accurate radiance fields.}
As demonstrated in Fig. \ref{demo}, our model outperforms other competitive approaches in predicting surface geometry (top row) and surface normals (middle row). Besides, by estimating more accurate surface normals that determine the accuracy of reflection direction, we can also achieve promising rendering realism as an added benefit (bottom row).


Although the idea discussed above is simple, designing the reflection-aware photometric loss is non-trivial. 
One straightforward approach is to follow the NeRF-W \cite{martin2021nerfw}, in which the vanilla photometric loss is extended to a Bayesian learning framework \cite{kendall2017uncertainties}. It formulates radiance as a Gaussian distribution with the learned uncertainty representing the variance, expecting the uncertainty can localize transient components of an image in the wild and eliminate its influence for static components learning.
However, this method is not applicable in reflective scenes, since it learns the implicit uncertainty that only considers the information of a single ray and ignores the multi-view context.

To address this issue, we propose defining an explicit reflection score that leverages multi-view context obtained through pixel colors from multi-view images referring to the same surface point. First, we identify the visibility of all source views given a surface point. Next, we project the point to visible images to obtain pixel colors. Based on this, we use an anomaly detector to estimate the reflection score, which serves as the variance. By minimizing the negative log-likelihood of the Gaussian distribution of the  color, a large variance attenuates the importance. We further demonstrate that, by using reflection direction-dependent radiance, our model achieves promising results for multi-view reconstruction with a better radiance fields.

To summarize, our contributions are listed as follows.

\begin{itemize}
    \item \wenhang{To the best of our knowledge, we present the first neural implicit surface learning framework for reconstructing objects with reflective surfaces.}
    
    \item \yc{We propose a simple yet effective approach that enable neural implicit surface learning to handle reflective surfaces. Our approach can produce high-quality surface geometry and surface normals.}

    \item Extensive experiments on several datasets show that the proposed framework significantly outperforms the state-of-the-art methods on reflective surfaces.
\end{itemize}

\section{Related Works}

\subsection{Multi-View Stereo for 3D Reconstruction}

Multi-View Stereo (MVS) is a technique that aims to reconstruct fine-grained scene geometry from multi-view images. Traditional MVS methods can be classified into four categories based on the output scene representation: volumetric-based methods \cite{curless1996volumetric, seitz1999photorealistic}, mesh-based methods \cite{fua1995object}, point cloud-based methods \cite{furukawa2009accurate, lhuillier2005quasi}, and depth map-based methods \cite{campbell2008using, galliani2015massively, schonberger2016structure, schonberger2016pixelwise, xu2019multi}. Among them, depth map-based methods are the most flexible, estimating depth maps for each view by utilizing photometric consistency between reference and neighboring images \cite{furukawa2009accurate}, then fusing all the depth maps into dense point clouds. Surface reconstruction methods \cite{curless1996volumetric,kazhdan2013screened, labatut2007efficient}, such as screened Poisson Surface Reconstruction \cite{kazhdan2013screened}, are then employed to reconstruct surfaces from the point clouds.

Recent learning-based MVS methods use deep learning for improved reconstruction. SurfaceNet \cite{ji2017surfacenet} and LSM \cite{kar2017learning} were the first volumetric learning-based MVS pipelines proposed to regress surface voxels. More recently, MVSNet \cite{yao2018mvsnet} extracts deep image features and warps them into the reference camera frustum to construct a 3D cost volume via differentiable warping.

However, learning-based MVS methods may still produce unsatisfactory results in certain scenarios, such as surfaces with specular reflections, regions with low texture, and non-Lambertian regions. In these cases, photometric consistency across multi-view images is not guaranteed, which can lead to severe artifacts and missing parts in the reconstruction results.

\subsection{Neural Implicit Surface for 3D Reconstruction}


Recently, learning-based approaches using implicit surface representations have been proposed. In these representations, a neural network maps continuous points in 3D space to an occupancy field \cite{mescheder2019occupancy, peng2020convolutional} or a Signed Distance Function (SDF) \cite{park2019deepsdf}. These methods perform multi-view reconstruction with additional supervision corresponding to the occupancy value or SDF for each point. However, supervision for these methods is not always available with only multi-view 2D images, which limits their scalability.

The introduction of volumetric approaches in NeRF \cite{mildenhall2021nerf}, which combines classical volume rendering \cite{kajiya1984ray} with implicit functions for novel view synthesis, has attracted a lot of attention towards 3D reconstruction using neural implicit surface representations and volume rendering \cite{oechsle2021unisurf, volsdf, wang2021neus}. Unlike NeRF, which aims to render novel view images while keeping the geometry unconstrained, these methods define the surface in a more explicit manner and are therefore better suited for surface extraction. UNISURF \cite{oechsle2021unisurf} uses the occupancy field \cite{mescheder2019occupancy}, while IDR \cite{IDR}, VolSDF \cite{volsdf}, and NeuS \cite{wang2021neus} use the SDF field \cite{park2019deepsdf} as an implicit surface representation. Despite their promising performance on 3D reconstruction, these methods fall short of recovering the correct geometry of objects with reflections, leading to ambiguous surface optimization. Our approach builds upon NeuS \cite{wang2021neus}, but we believe it can be adapted to fit any volumetric neural implicit framework.

 \subsection{Modeling for Object with Reflection}
 We discuss rendering and reconstruction for objects with reflections. Recent works \cite{boss2021nerd, srinivasan2021nerv, zhang2021physg, zhang2021nerfactor, verbin2022ref} have investigated rendering view-dependent reflective appearance by decomposing a scene into shape, reflectance, and illumination for novel view synthesis and relighting. However, the recovered meshes are not explicitly validated, and the geometry is often unsatisfactory. Reconstruction, on the other hand, aims to recover explicit geometry, which is still under-explored due to the inherent challenges. For instance, PM-PMVS \cite{cheng2021multi} formulates the reconstruction task as joint energy minimization over surface geometry and reflectance, while nLMVS-Net \cite{yamashita2023nlmvs} formulates MVS as an end-to-end learnable network, leveraging surface normals as view-independent surface features for cost volume construction and filtering. However, none of these methods combines neural implicit surfaces with volume rendering for reconstruction.

\subsection{Warping-based Consistency Learning}

Warping-based consistency learning is widely used in both multi-view stereo \cite{wang2021patchmatchnet, xu2020marmvs, xu2019multi, mvsdf} and neural implicit surface learning \cite{darmon2022improving, fu2022geo} for 3D reconstruction by exploiting inter-image correspondence with differentiable warping operations. Generally, consistency learning in MVS-based pipelines is performed at the CNN feature level. For example, MVSDF \cite{mvsdf} warps the predicted surface point to its neighboring views and enforces pixel-wise feature consistency, while ACMM \cite{xu2019multi} warps the coarse predicted depth to form a multi-view aggregated geometric consistency cost to refine finer scales. Consistency learning in neural implicit surface-based pipelines, on the other hand, is typically performed at the image level. NeuralWarp \cite{darmon2022improving} warps the sampled points along a ray to source images to obtain their RGB values and optimizes them jointly with a radiance network, while Geo-NeuS \cite{fu2022geo} warps the gray-scale patch centered on the predicted surface point to its neighboring images to guarantee multi-view geometry consistency. However, they ignore view-dependent radiance and are limited when multi-view consistency is not reasonable due to reflection, which can lead to artifacts when minimizing patch similarity regardless of reflection. Furthermore, visibility identification is not well-handled, and both rely on cumbersome preprocessing to determine source images. Alternatively, we leverage inconsistency to reduce ambiguity for high-fidelity reconstruction.

\section{Approach} 

\begin{figure*}
    \centering
     \includegraphics[width=0.95\textwidth]{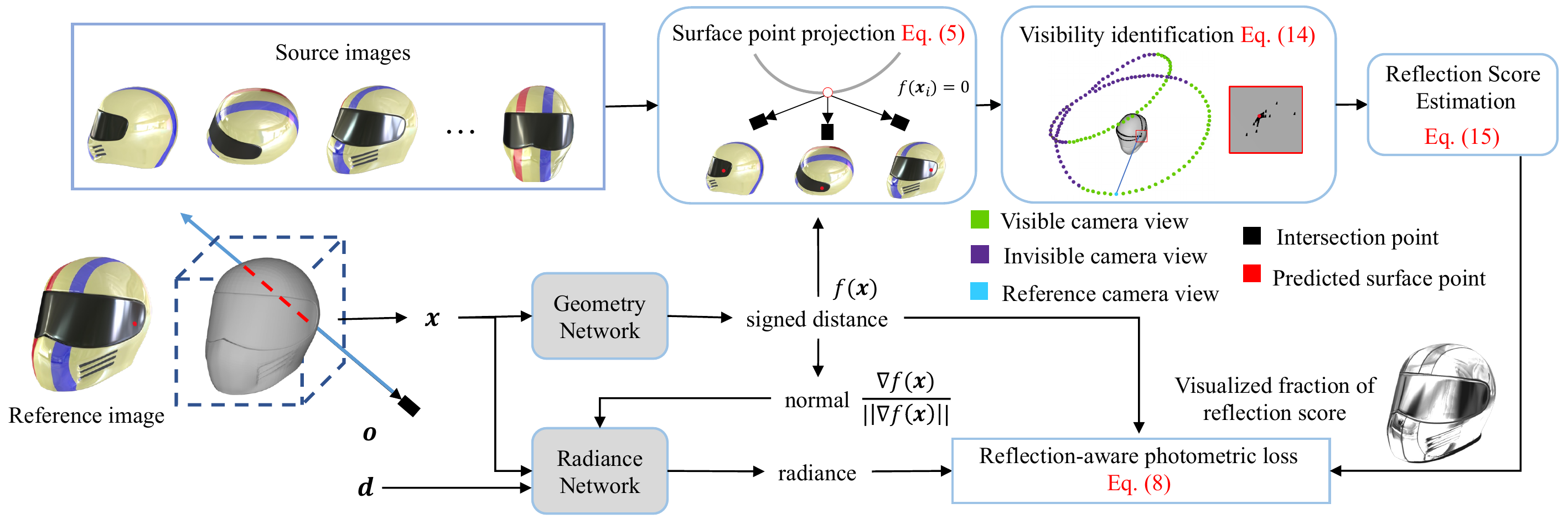}
     \caption{An overview of our framework. We propose a reflection-aware photometric loss to address the ambiguity induced by reflection. This loss attenuates the importance of reflective surfaces by lowering the weights of reflective pixels. Our loss relies on the estimated reflection score to localize these pixels.
    We use an  anomaly detector that leverages a multi-view context to obtain the reflection score. This context is captured by projecting surface points to visible source views with a visibility identification module.
     }
     \label{overview}
     \vspace{-0.3cm}
 \end{figure*}

Given $N$ calibrated multi-view images $\mathcal{X} = \left\{\mathbf{I}_{i} \right\}_{i=1}^{N}$ of an object with reflective surfaces, we aim to reconstruct the surfaces by neural implicit surface learning.
Section \ref{Preliminary} introduces NeuS, our baseline for reconstruction. 
Section \ref{Uncertainty} introduces a reflection-aware photometric loss. It reduces the influence of reflection  by formulating rendering color as a Gaussian distribution with an explicit variance estimation that considers multi-view context.
Section \ref{Refinement} discusses how we identify the visibility of source views to obtain unbiased reflection score. Section \ref{radiance} shows that together with a reflection direction-dependent radiance, our model achieves better geometry with a better radiance fields. Finally, Section \ref{optimization} presents our full optimization. An overview of our approach  is provided in Figure \ref{overview}.

\subsection{Volume Rendering with Implicit Surface}
\label{Preliminary}

Volume rendering \cite{kajiya1984ray} is used in NeRF \cite{mildenhall2021nerf} for
novel view synthesis. The idea is to represent the continuous attributes (i.e., density and radiance) of a 3D scene with neural networks. $\alpha$ compositing \cite{max1995optical} aggregates these attributes along a ray $\mathbf{r}$ to approximate the  pixel RGB values by:
\begin{equation}\label{render color}
    \hat{\mathbf{C}}(\mathbf{r})=\sum_{i=1}^P T_i \alpha_i \mathbf{c}_i, 
\end{equation}
where $T_i=\exp \left(-\sum_{j=1}^{i-1} \alpha_j \delta_j\right)$ and $\alpha_i=1-\exp \left(-\sigma_i \delta_i\right)$ denote the transmittance and alpha value of sampled point, respectively.   $\delta_i$
is the distance between neighboring sampled points. $P$ is the number of sampled points along a ray. $\sigma_i$ and $\mathbf{c}_i$ are predicted attributes by the neural networks conditioned on position $\mathbf{x}=(x, y, z)$ and view direction $\mathbf{d}=(\theta, \phi)$.
The training object $\mathcal{L}$ of NeRF is the mean square error between the ground-truth pixel color $\mathbf{C}(\mathbf{r})$ and the rendering color $\hat{\mathbf{C}}(\mathbf{r})$ formulated as
\begin{equation}
    \label{rgb_loss}
    \mathcal{L}_{\text{color}}=\sum_{\mathbf{r} \in \mathcal{R}}\|\mathbf{C}(\mathbf{r})-\hat{\mathbf{C}}(\mathbf{r})\|_2^2,
\end{equation}
where $\mathcal{R}$ is the set of all rays shooting from the camera center to image pixels. 

However, density-based volume rendering lacks a clear definition of the surface, which makes it difficult to extract precise geometry. Alternatively, Signed Distance Function (SDF) clearly defines the surface as the zero-level set, making SDF-based volume rendering more effective in surface reconstruction. Following NeuS \cite{wang2021neus}, the attributes of a 3D scene include signed distance and radiance parameterized by a geometry network $f$ and a radiance network $c$ by:
\begin{equation}\label{view-dep radiance}
    s = f(\mathbf{x}), \quad
    \mathbf{c} = c(\mathbf{x}, \mathbf{d}),
\end{equation}
where the geometry network $f$ maps a spatial position $\mathbf{x} $ to its signed distance $f( \mathbf{x} )$ to the object and the radiance network $c$ predicts the color conditioned on position $\mathbf{x} $ and view direction $\mathbf{d}$ to model the view-dependent radiance. To aggregate the signed distances and colors of sampled points along a ray $\mathbf{r}$ for pixel color approximation, we utilize volume rendering similar to that of NeRF. The key difference is the formulation of $\alpha_{i}$, which is calculated from the signed distance $f\left(\mathbf{x}\right)$ rather than density $\sigma_i$ as
\begin{equation}
    \alpha_{i}=\max \left(\frac{\Phi_{s}\left(f(\mathbf{x}_i) \right) -\Phi_{s}\left(f(\mathbf{x}_{i+1}) \right)}{\Phi_{s}\left(f(\mathbf{x}_i) \right)}, 0\right),
\end{equation}
where $\Phi_{s}(x)=\left(1+e^{-s x}\right)^{-1}$ and $1 / s$
 is a trainable parameter which indicates the standard deviation of $\Phi_{s}(x)$.

\subsection{Anomaly Detection for Reflection Score}
\label{Uncertainty}

For multi-view reconstruction, multi-view consistency is the promise for accurate surface reconstruction. However, for reflective pixels, the geometry network often predicts ambiguous surfaces, which devastates the multi-view consistency.
To overcome the issue, we propose to reduce the influence of reflective surfaces by a reflection-aware photometric loss, which adaptively lowers the weights assigned to reflective pixels. To achieve this, we first define the reflection score, which allows us to identify reflective pixels.

A naive solution is to treat the uncertainty defined in NeRF-W \cite{martin2021nerfw} as the reflection score. This approach models the radiance value of a scene as a Gaussian distribution and considers the predicted uncertainty as the variance. By minimizing the negative log-likelihood of the Gaussian distribution, a large variance reduces the importance of a pixel with high uncertainty. Ideally, the reflective pixels should be assigned a large variance to attenuate its influence for reconstruction.
However, the implicit uncertainty learned by the MLP is defined on a single ray without considering the multi-view context. Therefore, it may not accurately localize reflective surfaces without explicit supervision.

Similar to NeRF-W \cite{martin2021nerfw}, we also formulate rendering the color of a ray as a Gaussian distribution $\hat{\mathbf{C}}(\mathbf{r}) \sim (\overline{\mathbf{C}}(\mathbf{r}), \overline{\beta}^{2}(\mathbf{r}))$, where $\overline{\mathbf{C}}(\mathbf{r})$ and $\overline{\beta}^{2}(\mathbf{r})$ are the mean and variance, respectively. We adopt Eq. \eqref{render color} to query $\overline{\mathbf{C}}(\mathbf{r})$.
However, unlike NeRF-W, which only defines the implicit variance based on the information of a single ray, we explicitly define the variance based on the multi-view context. Specifically, we utilize multi-view pixel colors that refer to the same surface point to determine the variance.

To obtain the multi-view pixel colors, we project the surface point $\mathbf{x}$ onto all $N$ images $\left\{\mathbf{I}_{i}\right\}_{i=1}^{N}$ and use bilinear interpolation to obtain the corresponding pixel colors $\left\{\mathbf{C}_i(\mathbf{r})\right\}_{i=1}^{N}$.
For simplicity, omitting the subscript, the pixel color $\mathbf{C}$ is  obtained by
\begin{equation}\label{projection}
    \begin{split}
    \mathbf{\mathcal{G}} = \mathbf{K} \cdot  \left(\mathbf{R} \cdot \mathbf{x} + \mathbf{T}\right), \\
    \mathbf{C} = \operatorname{interp}(\mathbf{I}, \mathcal{G}),
    \end{split}
\end{equation}
where $\operatorname{interp}$ indicates bilinear interpolation, $\mathbf{K}$ denotes the internal calibration matrix, $\mathbf{R}$ denotes the rotation matrix, $\mathbf{T}$ denotes the translate matrix and $\cdot$ is matrix multiplication.

Considering the fact that only the local region of
partial images contains reflection, we treat reflection localization as an anomaly detection problem, with the expectation that reflective surfaces will be regarded as the anomaly and assigned a high reflection score. To this end, we  estimate a view-dependent reflection score $\overline{\beta}^{2}_{i}(\mathbf{r})$ by an anomaly detector empirically, which uses the Mahalanobis distance \cite{mahalanobis1936generalised} as the reflection score (i.e., variance) as follows:
\begin{equation}\label{ucc_view}
    \small
    \overline{\beta}^{2}_{i}(\mathbf{r}) = \gamma \frac{1}{N} \sum_{j=1}^{N} 
    \sqrt{
    \left( \mathbf{C}_i(\mathbf{r}) - \mathbf{C}_j(\mathbf{r})\right)^T \mathbf{\Sigma} ^{-1} 
    \left( \mathbf{C}_i(\mathbf{r}) - \mathbf{C}_j(\mathbf{r})\right)},
\end{equation}
where $\gamma$ is the a scale factor to control the scale of the reflection score and $\mathbf{\Sigma}$ is the empirical covariance matrix.  
As reflections do not dominate the majority of training images, if the currently rendered pixel color $\mathbf{C}_i(\mathbf{r})$ is contaminated by reflection, a larger reflection score is generated as most of the relative divergences become large.

Then,  \wenhang{we extend the photometric loss in Eq. \eqref{rgb_loss} to a reflection-aware one by minimizing the negative log-likelihood of distribution of rays $\mathbf{r}$ from a batch $\mathcal{R}$} similar to NeRF-W \cite{martin2021nerfw} and ActiveNeRF \cite{pan2022activenerf} as follows:
\begin{equation}
    \label{rew_rgb_loss_}
    \mathcal{L}_{\text{color}}= -\log p(\mathbf{\hat{C}}(\mathbf{r}))=  \sum_{\mathbf{r} \in \mathcal{R}} \frac{\|\mathbf{C}(\mathbf{r})-\overline{\mathbf{C}}(\mathbf{r})\|^{2}_{2}}{2\overline{\beta}^2(\mathbf{r})} + \frac{\log \overline{\beta}^2(\mathbf{r})}{2}.
\end{equation}
Since $\overline{\beta}^2(\mathbf{r})$ is estimated by Eq. \eqref{ucc_view} explicitly instead of implicitly learning by MLP, it is a constant and can be removed from the objective function. Besides, following previous works \cite{wang2021neus, volsdf, fu2022geo} for multi-view reconstruction, we use L1 loss instead of L2 loss. Finally, our reflection-aware photometric loss is quite simple formulated as 
\begin{equation}
    \label{rew_rgb_loss}
    \mathcal{L}_{\text{color}}=  \sum_{\mathbf{r} \in \mathcal{R}} \frac{|\mathbf{C}(\mathbf{r})-\overline{\mathbf{C}}(\mathbf{r})|}{\overline{\beta}^2(\mathbf{r})}.
\end{equation}
Note that we slightly abuse the term ${\overline{\beta}^2(\mathbf{r})}$, which should be corresponding to L2 loss.


\subsection{Visibility Identification for Reflection Score}
\label{Refinement}

The computation of reflection score using all pixel colors $\left\{ \mathbf{C}_j(\mathbf{r})\right\}_{j=1}^{N}$  assumes that the point on the surface has a valid projection on all source images. However, in practice, this assumption is not true due to self-occlusion. The projected pixel colors are meaningless if the point is invisible in the source images, and the corresponding pixel colors should not be used in Eq. \eqref{ucc_view}.

\begin{figure}[htbp]
    \centering
    \includegraphics[width=0.47\textwidth]{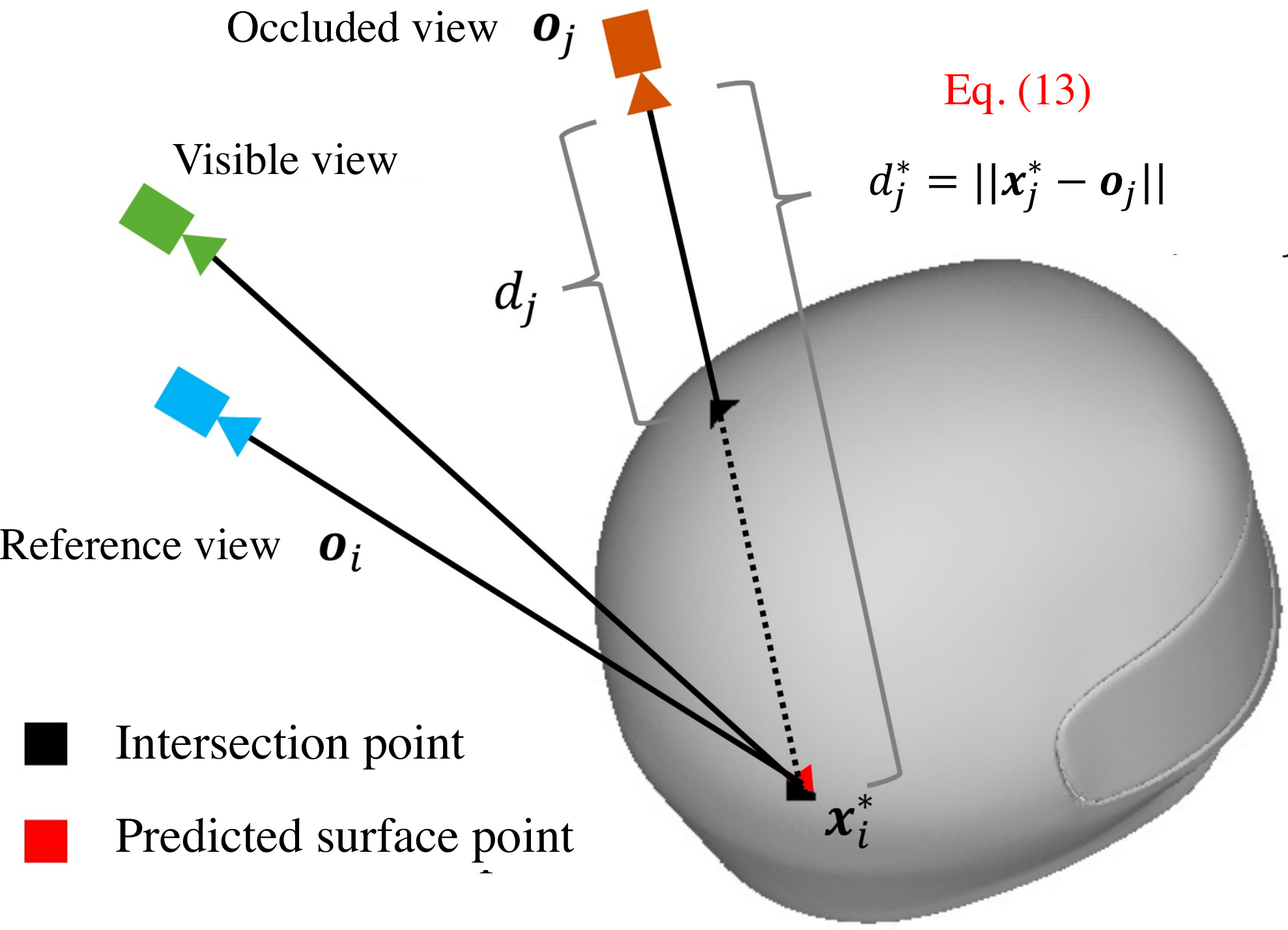}
    \caption{Illustration of visibility identification.}
    \label{example}
\end{figure}

To address this issue, we design a visibility identification module, which leverages intermediate reconstructed meshes to identify visibility, as illustrated in Fig. \ref{example}. Specifically, given a pixel $p_i$ corresponding to a ray $\mathbf{r}_i$, the implicit surfaces on ray $\mathbf{r}_i$ can be represented according to the signed distance of sampled points $\mathbf{x}$ as follows:
\begin{equation}
    \hat{S}_i=\{\mathbf{x} \mid f(\mathbf{x})=0 \}.
\end{equation}

Since an infinite number of points exist along a ray, we need to sample discrete points on this ray. Based on the sampled points and their signed distances, we can determine the intervals where surface points exist by
\begin{equation}
    T_i=\left\{\mathbf{x}_{j} \mid f\left(\mathbf{x}_{j}\right) \cdot  f\left(\mathbf{x}_{j+1}\right)<0\right\}.
\end{equation}
If the sign of the sampled point $f(\mathbf{x}_j)$ is different from the sign of the next sampled point $f(\mathbf{x}_{j+1})$, the interval $[\mathbf{x}_{j}, \mathbf{x}_{j+1}]$ intersects with the surface. The intersection point set $\hat{S}_i$ can be obtained by linear interpolation by
\begin{equation}
    \hat{S}_{i}=\left\{\mathbf{x} \mid \mathbf{x}=\frac{f(\mathbf{x}_{j}) \mathbf{x}_{j+1}-f(\mathbf{x}_{j+1}) \mathbf{x}_{j}}{f(\mathbf{x}_{j})-f(\mathbf{x}_{j+1})}, \mathbf{x}_{j} \in T_i\right\}.
\end{equation}
In practice, the ray $\mathbf{r}_i$ may intersect with the object at multiple surfaces. For our reflection score computation, only the first intersection is meaningful, and it is formulated as
\begin{equation}
    \mathbf{x}_{i}^{*}=\operatorname{argmin}\mathcal{D}(\mathbf{x}, \mathbf{o}_i),
\end{equation}
where $\mathbf{x} \in \hat{S}_{i}$ and $\mathcal{D}(\cdot, \mathbf{o}_i)$ indicate the distance between point $\mathbf{x}$ and the origin of the ray $\mathbf{r}_i$, respectively.

\begin{figure}
    \centering
    \begin{minipage}{0.48\textwidth}
    \includegraphics[width=0.24\columnwidth]{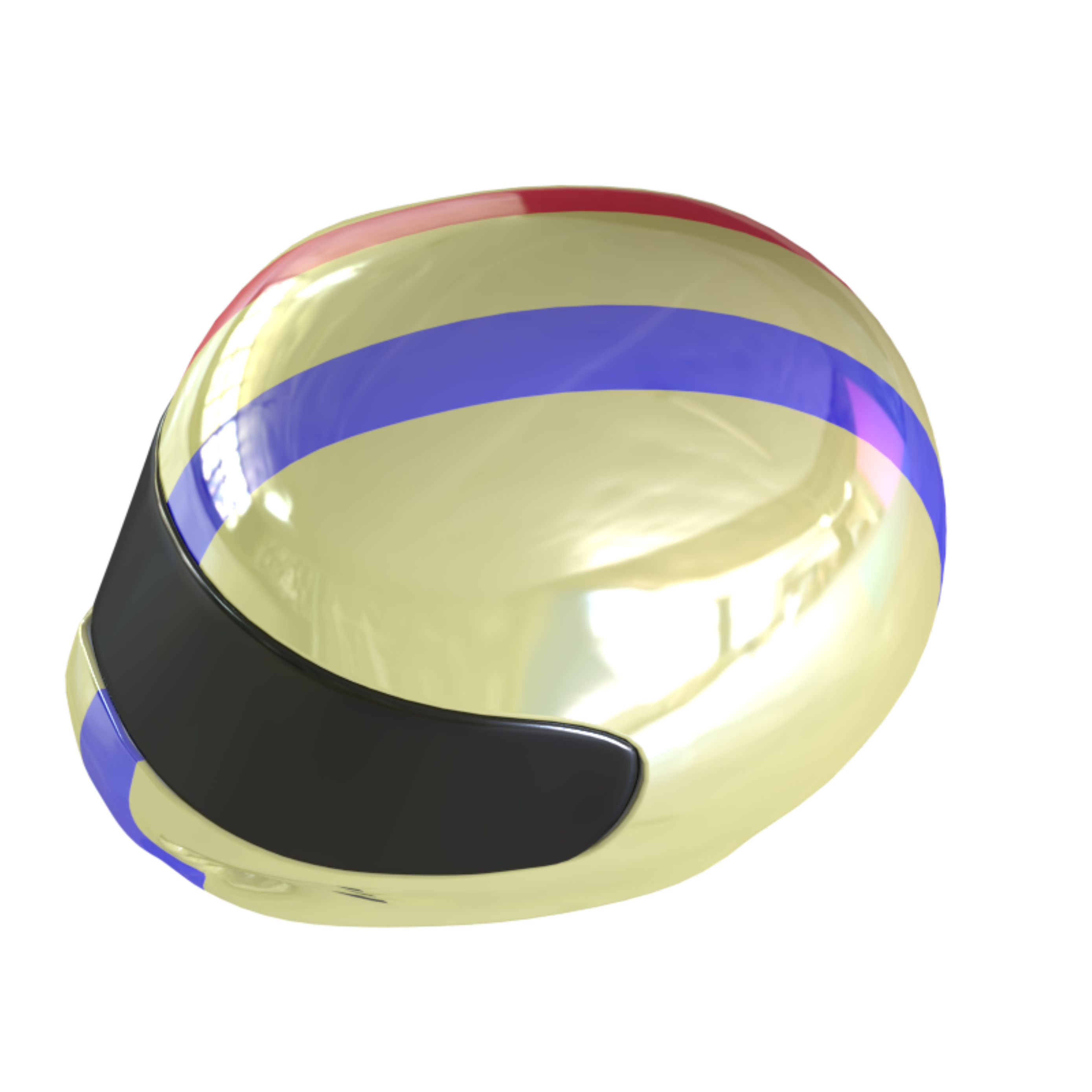}
    \includegraphics[width=0.24\columnwidth]{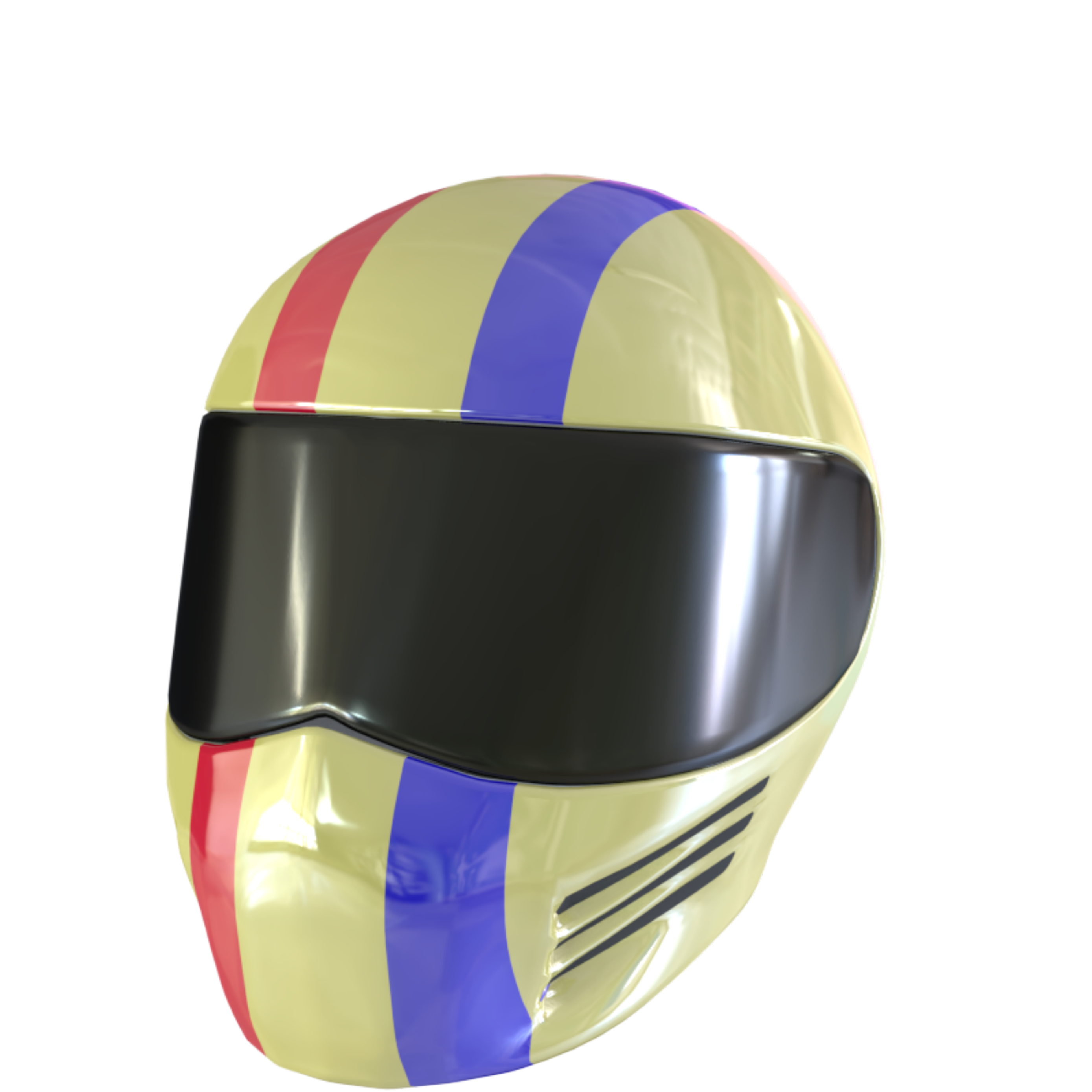}
    \includegraphics[width=0.24\columnwidth]{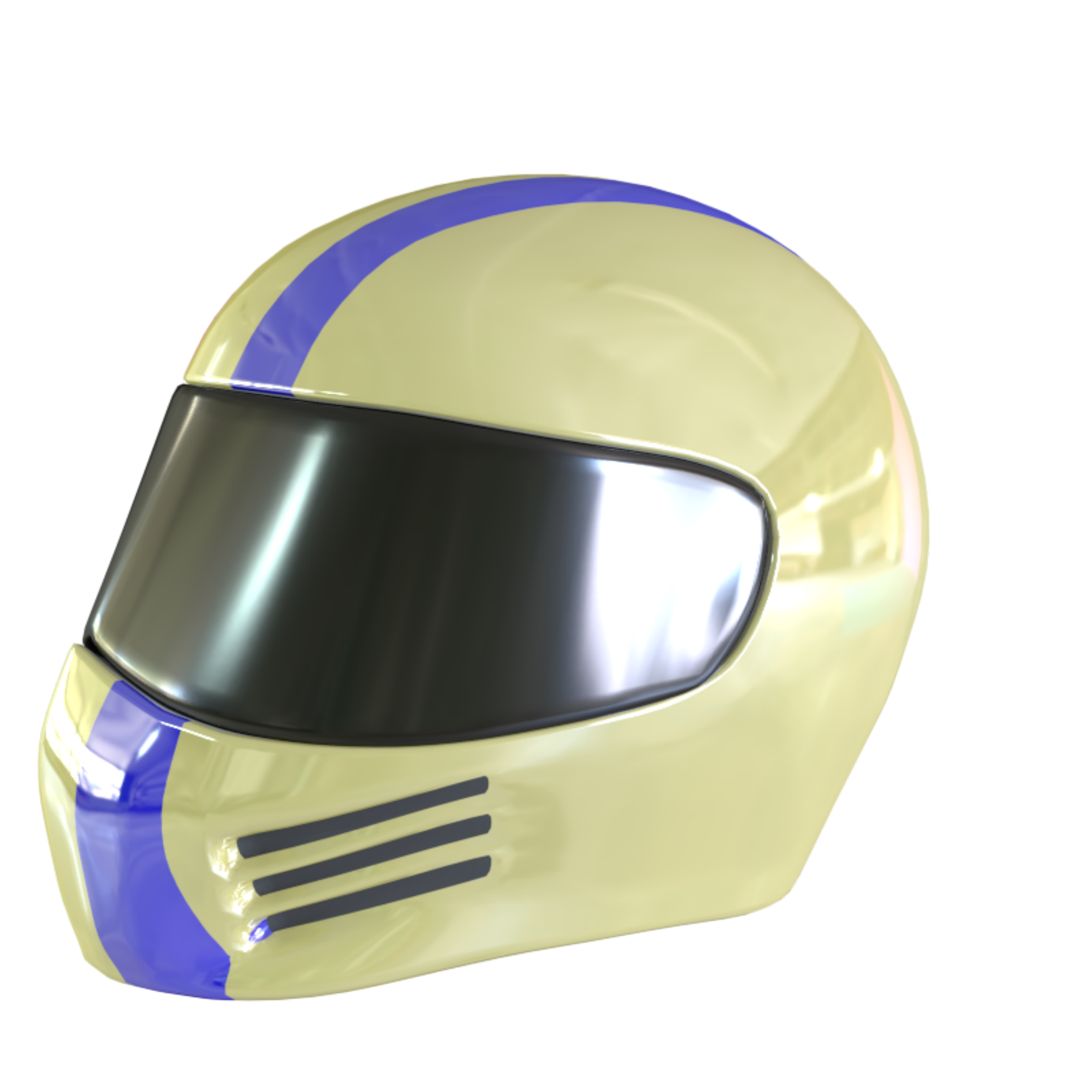}
    \includegraphics[width=0.24\columnwidth]{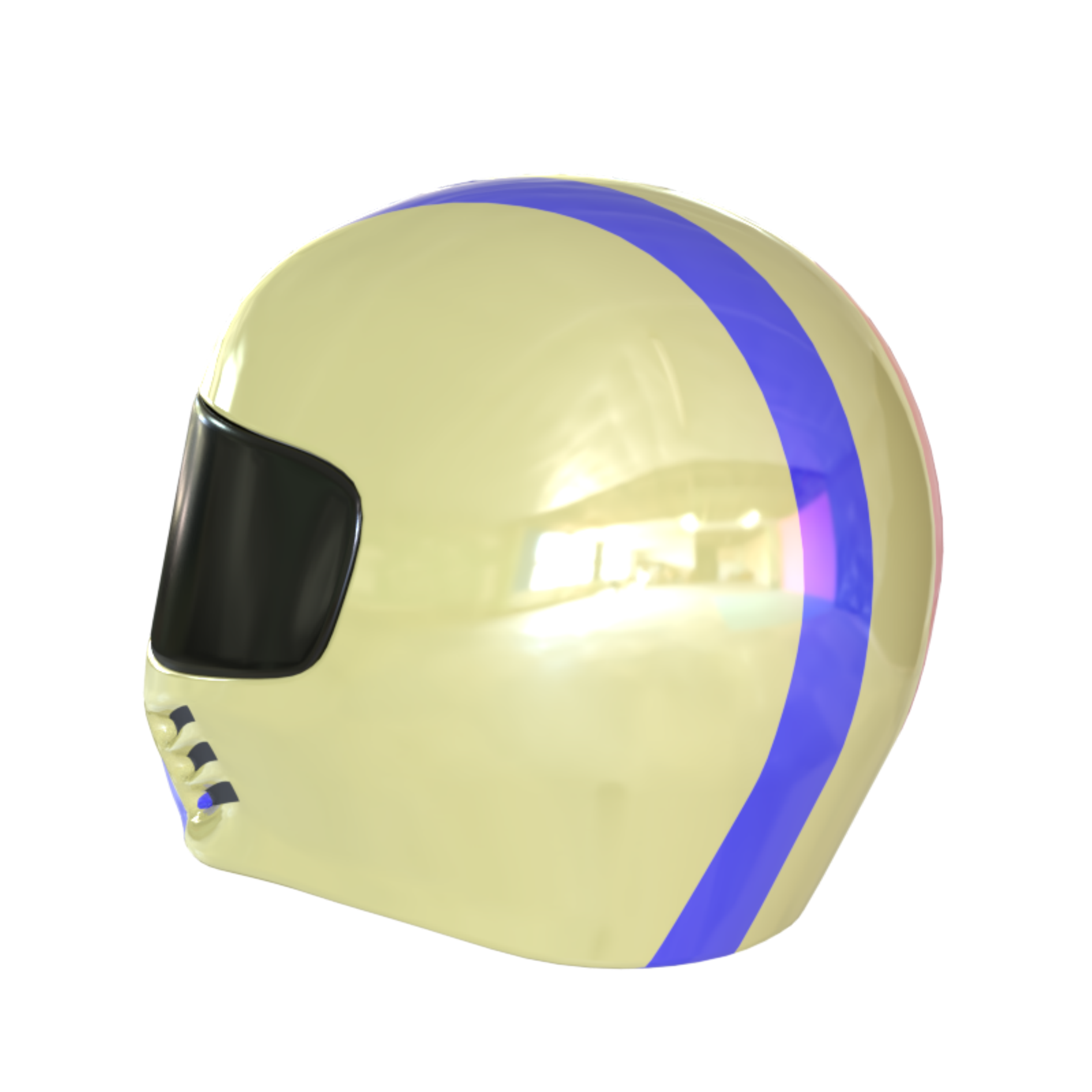}
    \end{minipage}
    \vspace{-0.3cm}
    
    \begin{minipage}{0.48\textwidth}
    \includegraphics[width=0.24\columnwidth]{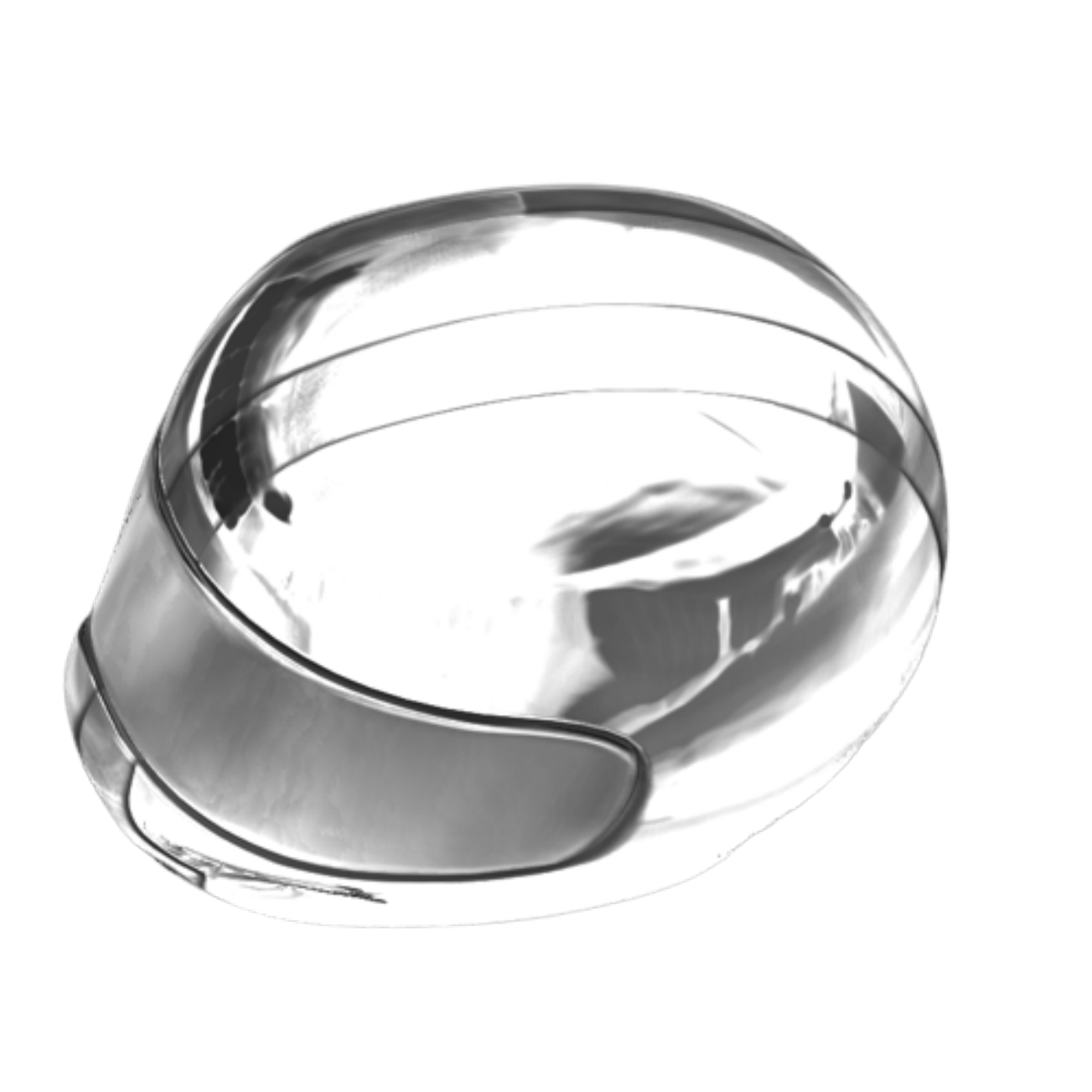}
    \includegraphics[width=0.24\columnwidth]{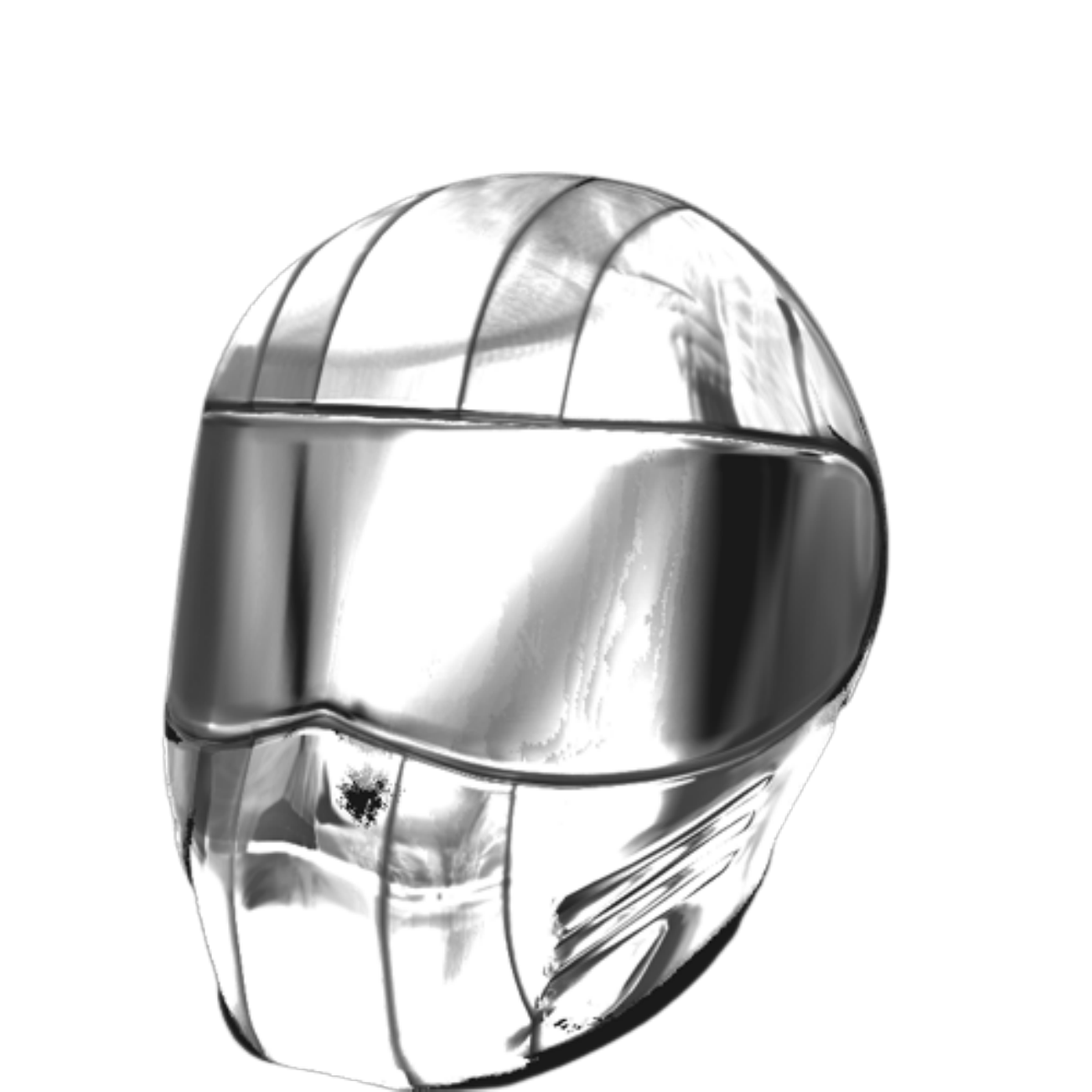}
    \includegraphics[width=0.24\columnwidth]{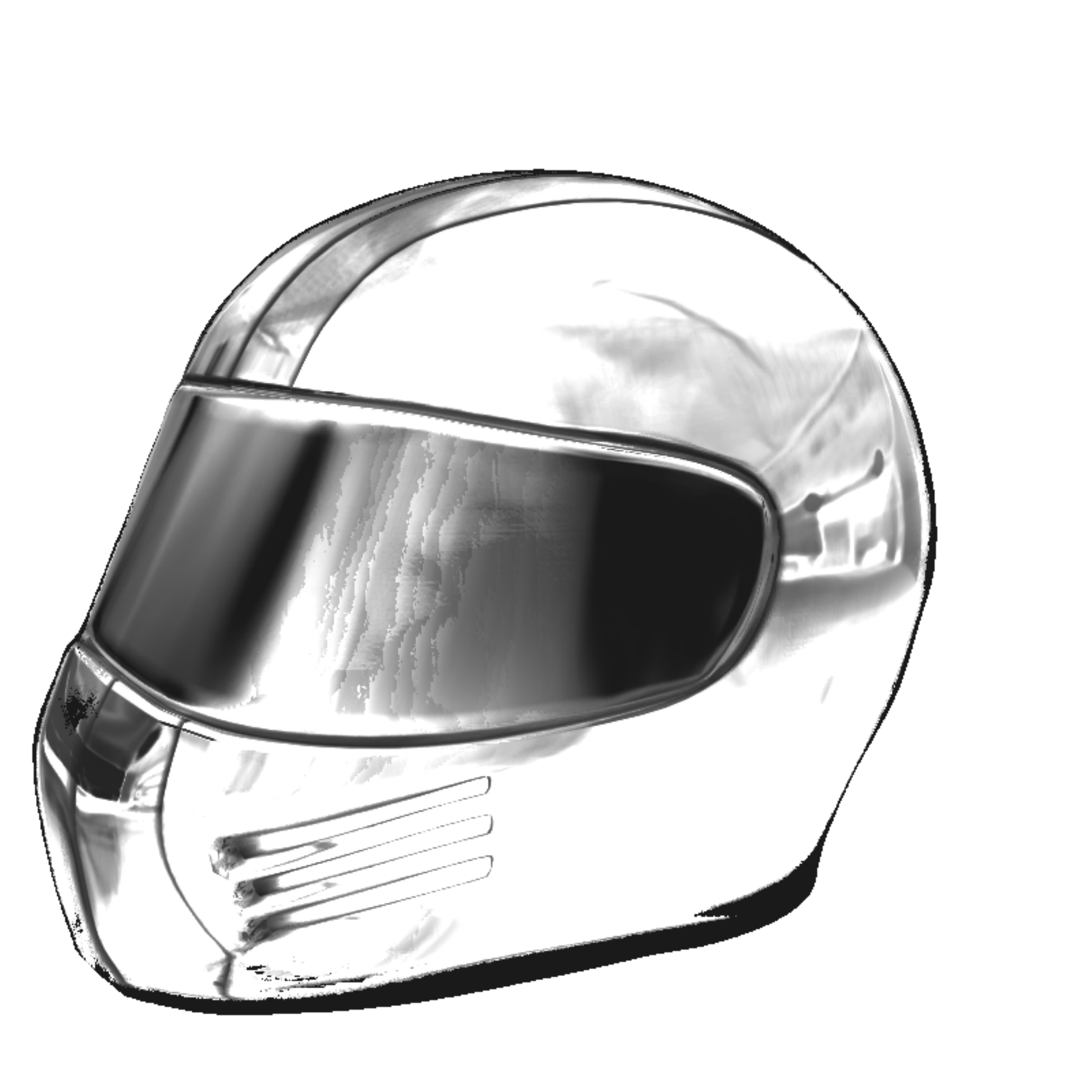}
    \includegraphics[width=0.24\columnwidth]{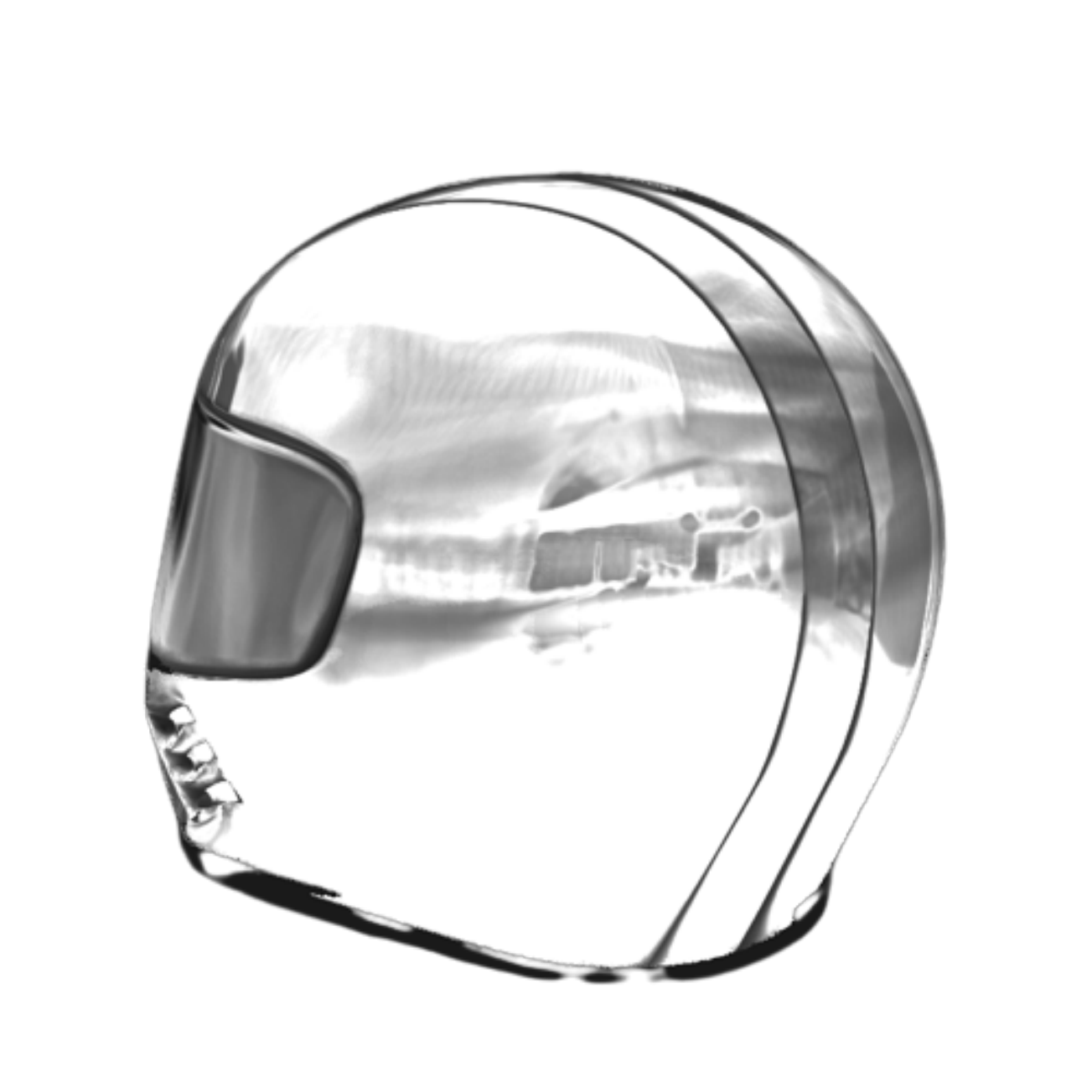}
    \end{minipage}
    \vspace{0.1cm}
    \caption{Visualization of the fraction of the reflection score, which can  localize reflective surfaces. Black indicates a high score.}
    \label{ucc}
\end{figure}

After capturing the predicted surface point $\mathbf{x}_{i}^{*}$, we can calculate the distances $\left\{d_j^*\right\}_{j=1}^{N}$ between this point and all camera locations $\left\{\mathbf{o}_j\right\}_{j=1}^{N}$ as follows:

\begin{equation}
    d_j^* = \| \mathbf{x}_{i}^{*} - \mathbf{o}_j \|.
\end{equation}
Meanwhile, we compute the distances $\left\{d_j\right\}_{j=1}^{N}$ from all  camera locations $\left\{\mathbf{o}_j\right\}_{j=1}^{N}$ to the first intersection of the intermediate reconstructed meshes by ray casting \cite{roth1982ray}. Based on these two distances, the visibility of images $\mathbf{I}_j$ can be estimated by
\begin{equation}
    \label{visi}
    v_j = \mathbb{I}(d_j^* \leq d_j )
\end{equation}
where $\mathbb{I}(\cdot)$ is an indicator function.
Based on the approximation of visibility, we eliminate the invisible pixel colors $\{\mathbf{C}_j(\mathbf{r}) \mid v_j=0\}$ that are used in the calculation of the reflection score in Eq. \eqref{ucc_view}. We then refine the reflection score as follows:
\begin{multline}\label{ucc_form}
    \small
    {\overline{\beta}^{2}_{i}(\mathbf{r})} = 
    \gamma \frac{1}{\sum_{j=1}^{N} v_j} \sum_{j=1}^{N} v_j\text{Mdis},
     \\
    \small
    \text{Mdis} = \sqrt{\left( \mathbf{C}_i(\mathbf{r}) -  \mathbf{C}_j(\mathbf{r})\right)^T \mathbf{\Sigma} ^{-1}\left(  \mathbf{C}_i(\mathbf{r}) -  \mathbf{C}_j(\mathbf{r})\right)}.
\end{multline}
We provide some examples in Fig. \ref{ucc} to illustrate the estimated reflection score.

\subsection{Reflection Direction Dependent Radiance}
\label{radiance}

\wenhang{
As suggested in Ref-NeRF \cite{verbin2022ref}, conditioning the radiance on reflection direction in reflective scenes can lead to a more accurate radiance fields, which has been shown to be beneficial for reconstruction in NeuralWarp \cite{darmon2022improving}. Inspired by this, we reparameterize the radiance network as a function of the reflection direction about the surface normal, and the formulation in Eq. \eqref{view-dep radiance} becomes
\begin{equation}\label{col}
    \mathbf{c} = c(\mathbf{x}, \boldsymbol{\omega}_r),
\end{equation}
where $\hat{\mathbf{d}}$ is the reflection direction calculated by
\begin{equation}
    \boldsymbol{\omega}_r = 2(-\mathbf{d} \cdot \mathbf{\hat{n}} )\mathbf{\hat{n}} + \mathbf{d},
\end{equation}
where $\mathbf{n}$ is the surface normal formulated by 
\begin{equation}
    \mathbf{\hat{n}} = \frac{\nabla f(\mathbf{x})}{||\nabla f(\mathbf{x})||}.
\end{equation}
}

\wenhang{Compared to Ref-NeRF \cite{verbin2022ref}, the reflection direction is more precise in our framework due to the well-estimated surface normal, which leads to a more accurate radiance fields. Compared to NeuralWarp \cite{darmon2022improving} that ignores reflection in their framework, we take the view-dependent radiance into consideration and estimate a more accurate radiance fields. As a result, our method is more reliable and promising for multi-view reconstruction of objects with reflection.}

\subsection{Optimization}
\label{optimization}

During reconstructing a  scene, our total loss function is
\begin{equation}
    \mathcal{L} = \mathcal{L}_{\text{color}} + \alpha \mathcal{L}_{\text{eik}}.
\end{equation}
$\mathcal{L}_{\text{color}}$ is the reflection-aware photometric loss in Eq. \eqref{rew_rgb_loss} with reflection score estimated by Eq. \eqref{ucc_form}. The rendered color is computed by Eq. \eqref{render color} with radiance parameterized in Eq. \eqref{col}.
$\mathcal{L}_{\text{eik}}$ is an eikonal term \cite{eik} to regularize the gradients of geometry network formualated as
\begin{equation}
    \mathcal{L}_{\text{eik}}=\frac{1}{P} \sum_{i=1}^{P}\left(\left|\nabla f\left(\boldsymbol{x}_{i}\right)\right|-1\right)^{2}.
\end{equation}
In our experiments, we choose $\alpha$ as 0.1.

\section{Experimetns}

\subsection{Datasets and Evaluation Protocol}

To evaluate the effectiveness of our model, we conducted experiments on objects from several datasets, including Shiny Blender \cite{verbin2022ref}, Blender \cite{mildenhall2021nerf}, SLF \cite{wood2000surface}, and Bag of Chips \cite{park2020seeing}. The selected objects in these datasets are glossy.


\begin{table*}[t]
    \centering
   \small
    \begin{tabular}{c|cc|cc|cc|cc||cc}
         \hline
           \multirow{2}*{Methods}& \multicolumn{2}{c|}{helmet}&   \multicolumn{2}{c|}{toaster}&  \multicolumn{2}{c|}{coffee} &\multicolumn{2}{c||}{car} &\multicolumn{2}{c}{Mean} \\ \cline{2-11}
             &Acc$\downarrow$   &MAE$\downarrow$ &Acc$\downarrow$   &MAE$\downarrow$  &Acc$\downarrow$  &MAE$\downarrow$ &Acc$\downarrow$   &MAE$\downarrow$ &Acc$\downarrow$  &MAE$\downarrow$ \\  \hline\hline
          $\text{PhySG}$ \cite{zhang2021physg} &- &2.32 &- &9.75  &-    &22.51 &- &8.84 &- &10.86 \\  
          Ref-NeRF \cite{verbin2022ref}   &- &29.48 &- &42.87  &-    &12.24 &- &14.93 &- &24.88 \\  \hline
          $\text{IDR}$ \cite{IDR}      &-  &- &- &- &3.79  &3.68 &\underline{0.55}  &1.25 &- &- \\ \hline
          UNISURF \cite{oechsle2021unisurf}  &1.69  &1.78  &3.75 &5.51   &2.88 &3.15   &2.34 &1.98  &2.67 &3.11\\ 
          VolSDF \cite{volsdf}    &1.55  &1.37  &\underline{2.02} &\underline{2.53} &2.23   &2.28 &0.58 &\underline{1.04}  &\underline{1.60}  &1.81\\ 
          NeuS \cite{wang2021neus}    &1.33   &\underline{1.12} &3.26 &2.87 &\underline{1.42}  &\underline{1.99} &0.73  &1.10  &1.69 &\underline{1.77} \\ \hline
          NeuralWarp \cite{darmon2022improving}  &2.25  &1.94 &5.90  &3.51  &1.54   &2.04 &0.65  &1.07 &2.59 &2.14 \\
        Geo-NeuS \cite{fu2022geo} &\underline{0.74}  &2.36 &6.35  &3.76 &3.85   &6.36 &2.88  &5.67 &3.46 &4.54  \\\hline
        Ours  &\textbf{0.29}  &\textbf{0.38}  &\textbf{0.42} &\textbf{1.47}  &\textbf{0.77}  &\textbf{0.99} &\textbf{0.37} &\textbf{0.80}   &\textbf{0.46} &\textbf{0.91} 
        \\\hline
    
    \end{tabular}
    \vspace{0.1cm}
    \caption{Comparison with state-of-the-art methods on Shiny Blender Dataset.
    Except for Ref-NeRF and PhySG, whose results are taken from the original paper of Ref-NeRF, we implemented the released code on Shiny Blender dataset for other methods. IDR failed to recover meaningful geometry for helmet and toaster, so the results remain empty. Note that NeRF-W failed to produce meshes since it focuses on novel view synthesis in the wild, and COLMAP generated meshes with severe artifacts and missing parts. The quantitative results were not provided.
    $\textbf{Bold}$ results have the best score and $\text{\underline{underlined}}$ the second best. Our method outperforms these methods by a large margin. }
    \label{sota_shiny}  
    \vspace{-0.3cm}
\end{table*}

\vspace{0.1cm}
\noindent\textbf{Shiny Blender.} 
The Shiny Blender dataset is a synthetic dataset introduced in \cite{verbin2022ref}. The dataset includes six different glossy objects that were rendered in Blender with more challenging material properties. The original dataset was created for novel view synthesis evaluation. We selected four objects (i.e., helmet, coffee, toaster, and car) for reconstruction, as the geometry of the ball is too simple (i.e., a sphere of radius 1), and the teapot is less reflective. 
For all reconstruction tasks, we used the original 200 testing images for training, as we found that there are more reflective surfaces, which makes it more challenging for reconstruction. For quantitative evaluation, we provide ground-truth dense point clouds for each scene by upsampling points from the ground-truth meshes exported from Blender.
Since rendering quality was also compared on ShinyBlender, we emphasize that for fair comparison with existing methods, we used the original 100 training images for training. 

\vspace{0.1cm}
\noindent\textbf{Blender.}
We used the glossy drums from Blender dataset \cite{mildenhall2021nerf}. This dataset
was rendered in Blender as Shiny Blender dataset with 100 images for training.

\vspace{0.1cm}
\noindent\textbf{SLF.}
We used the glossy fish from SLF \cite{wood2000surface}. This dataset was captured in a lab-controlled environment. The cameras are distributed on a hemisphere around the center object. 

\vspace{0.1cm}
\noindent\textbf{Bag of Chips.}
We used the glossy cans and corncho1 from \cite{park2020seeing}, which provides meshes scanned by RGBD sensors for evaluation. Since there are more than 2000 images, we only sampled half of them for training due to limited memory. \footnote{The number of training images does not necessarily affect the reconstruction results when there are hundreds of uniformly-distributed views.}



\vspace{0.1cm}
\noindent\textbf{Evaluation Protocol.}
The evaluation metric is the Chamfer Distance, provided by the DTU evaluation metrics \cite{dtu_eval}. The metric can be divided into two parts: \emph{accuracy} and \emph{completeness} (see supplement for details). For the Shiny Blender dataset, the ground-truth meshes are double-layered, which results in many redundant points in the ground truth. Therefore, we only reported the accuracy in Table \ref{sota_shiny}, as it is more meaningful than completeness. Additionally, for the Shiny Blender dataset, mean angular error (MAE) and PSNR were used for evaluating the estimated normals and rendering quality, respectively.

\begin{table}[t]
    \centering
    \small
    \begin{tabular}{c|cccc}
        \hline
        Method &drums &fish &cans &corncho1 \\ \hline 
        $\text{IDR}^{*}$ \cite{IDR} &1.91 &\textbf{0.77} &1.45 &0.87   \\
        NeuS \cite{wang2021neus} &2.29 &1.14 &1.89 &0.92   \\
        Geo-NeuS \cite{fu2022geo} &- &0.87 &2.80 &1.19   \\
        Ours &\textbf{1.35}  &0.81  &\textbf{1.21} &\textbf{0.82}  \\ \hline
    
    \end{tabular}
    \vspace{0.1cm}
     \caption{Comparison with state-of-the-art methods on drums, fish, cans and corncho1. $\text{IDR}^{*}$ requires images masks for training. Geo-Neus failed to recover drums. The average of accuracy and completeness is reported.}
    \label{extra_data}
\end{table}

\begin{table}[t]
    \centering
    \small
    \begin{tabular}{c|cccc||c}
        \hline
        Method &helmet &toaster &coffee &car &mean\\ \hline  
        NeuS \cite{wang2021neus} &27.78 &23.51 &28.82 &26.34   &26.61  \\
         Ref-NeRF \cite{verbin2022ref}&29.68 &25.70 &\textbf{34.21} &\textbf{30.82} &30.10    \\ 
        Ours &\textbf{32.85}  &\textbf{26.97}  &31.05 &29.92  &\textbf{30.20}\\ \hline
    
    \end{tabular}
    \vspace{0.1cm}
     \caption{Rendering quality comparison on Shiny Blender dataset. PSNR is adopted as evluation metric.  }
    \label{psnr}
\end{table}

\subsection{Implementation Details}
We implemented our model based on NeuS \cite{wang2021neus}. The structure of the geometry network and radiance network was the same as that of NeuS. Please refer to our supplement for more details. To estimate visibility in an on-the-fly manner, the intermediate reconstruction result was updated every 500 iterations with a resolution of 128 to decrease the computation cost. $\gamma$ was set to 5. We trained our model for 200k iterations, which took approximately 7 hours on a single NVIDIA RTX 3090 Ti GPU for the reconstruction task. After convergence, a mesh can be extracted from the signed distance functions (SDFs) in a predefined bounding box using Marching Cubes \cite{lorensen1987marching} with a resolution of 512.

\subsection{Comparison with State-of-the-Art Methods}

We compared the reconstruction quality of our method with several other methods, including IDR \cite{IDR}, three baseline methods for multi-view reconstruction (UNISURF \cite{oechsle2021unisurf}, VolSDF \cite{volsdf}, and NeuS \cite{wang2021neus}), two warp-based consistency learning methods (NeuralWarp \cite{darmon2022improving} and Geo-NeuS \cite{fu2022geo}), and two methods specifically designed for modeling objects with reflection (Ref-NeRF \cite{verbin2022ref} and PhySG \cite{zhang2021physg}). The quantitative results are shown in Tables \ref{sota_shiny} and \ref{extra_data}, where \emph{accuracy} is indicated with the term ``Acc'' \cite{dtu_eval}. As COLMAP and NeRF-W fail to recover reflective surfaces, we did not compare with them. Instead, 
we presented the reconstructed meshes  with severe artifacts of COLMAP in the supplement. For NeRF-W, it focuses on novel view synthesis in the wild and failed to produce meshes.

\begin{table}[t]  
    \centering
    \small
    \begin{tabular}{c|ccc|cc}
        \hline
        Method &RS  &Vis &Ref  &Acc$\downarrow$ &MAE$\downarrow$ \\ \hline 
        NeuS  & & &  &1.69 &1.77  \\ 
        NeuS w/ RS  &\checkmark & &  &1.17 &1.43  \\ 
       NeuS w/ Ref & & &\checkmark  &1.36 &1.50  \\
       Ref-NeuS w/o Ref &\checkmark &\checkmark &  &0.80 &1.21  \\ 
        Ours (full) &\checkmark & \checkmark& \checkmark  &\textbf{0.46} &\textbf{0.91}  \\\hline 
    
    \end{tabular}
    \vspace{0.1cm}
     \caption{Ablation study on Shiny Blender dataset. ``RS'' indicates reflection score. ``Vis'' indicate visibility identification. ``Ref'' indicates reflection direction-dependent radiance.}
    \label{ablation}
    \vspace{-0.3cm}
\end{table}

\begin{figure*}
    
    \centering
    \begin{minipage}{0.98\textwidth}
    \centering
    \includegraphics[width=0.16\columnwidth]{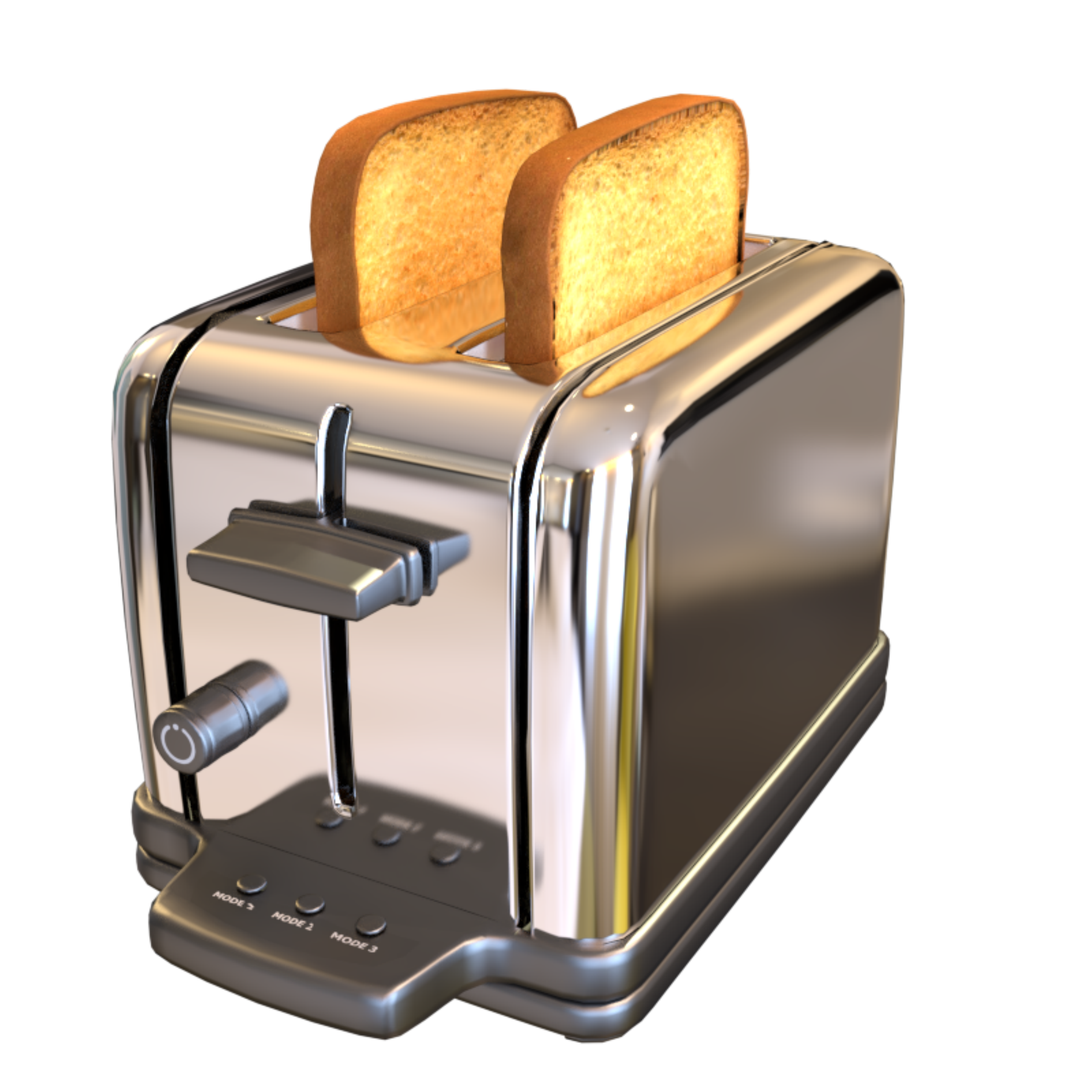}
     \includegraphics[width=0.16\columnwidth]{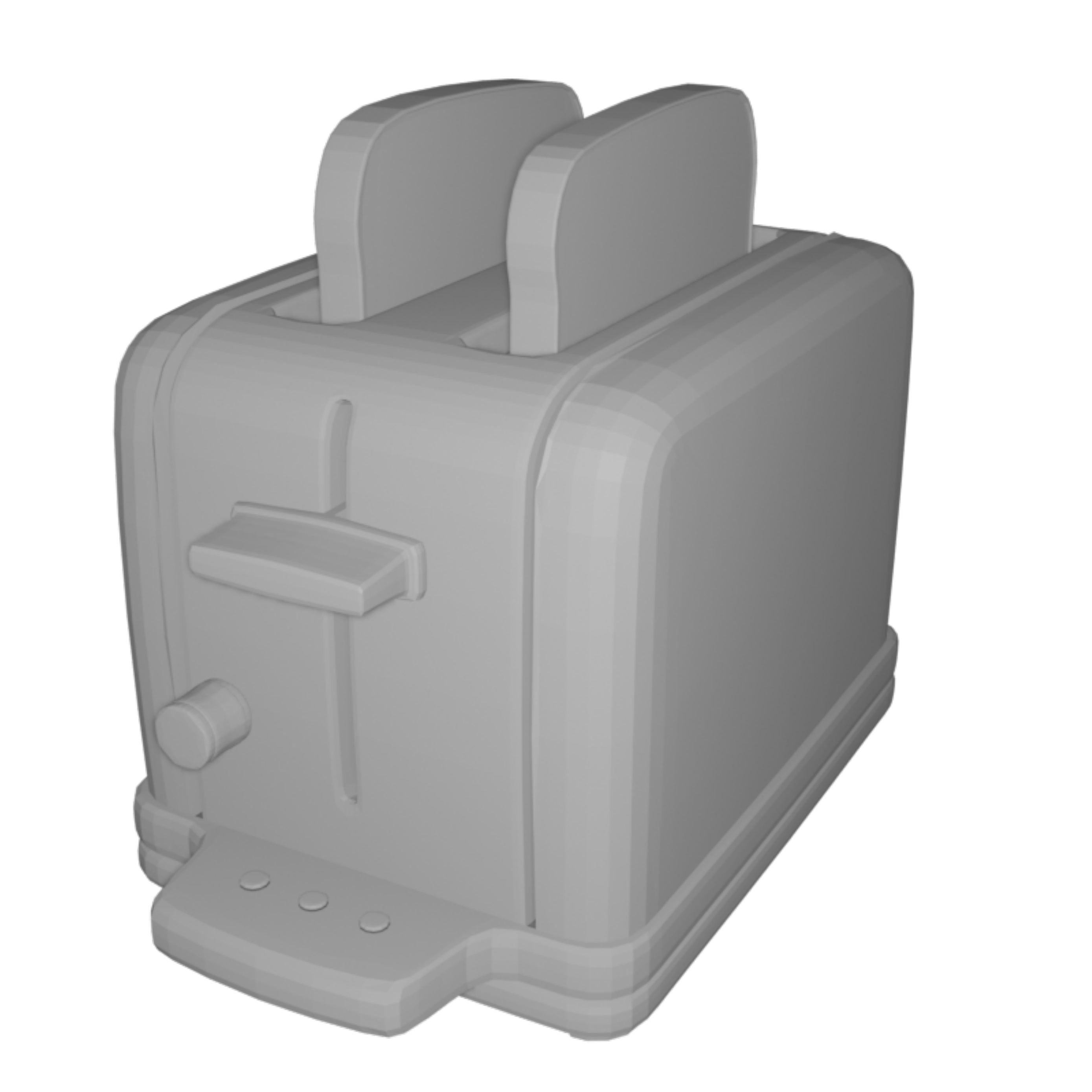}
     \includegraphics[width=0.16\columnwidth]{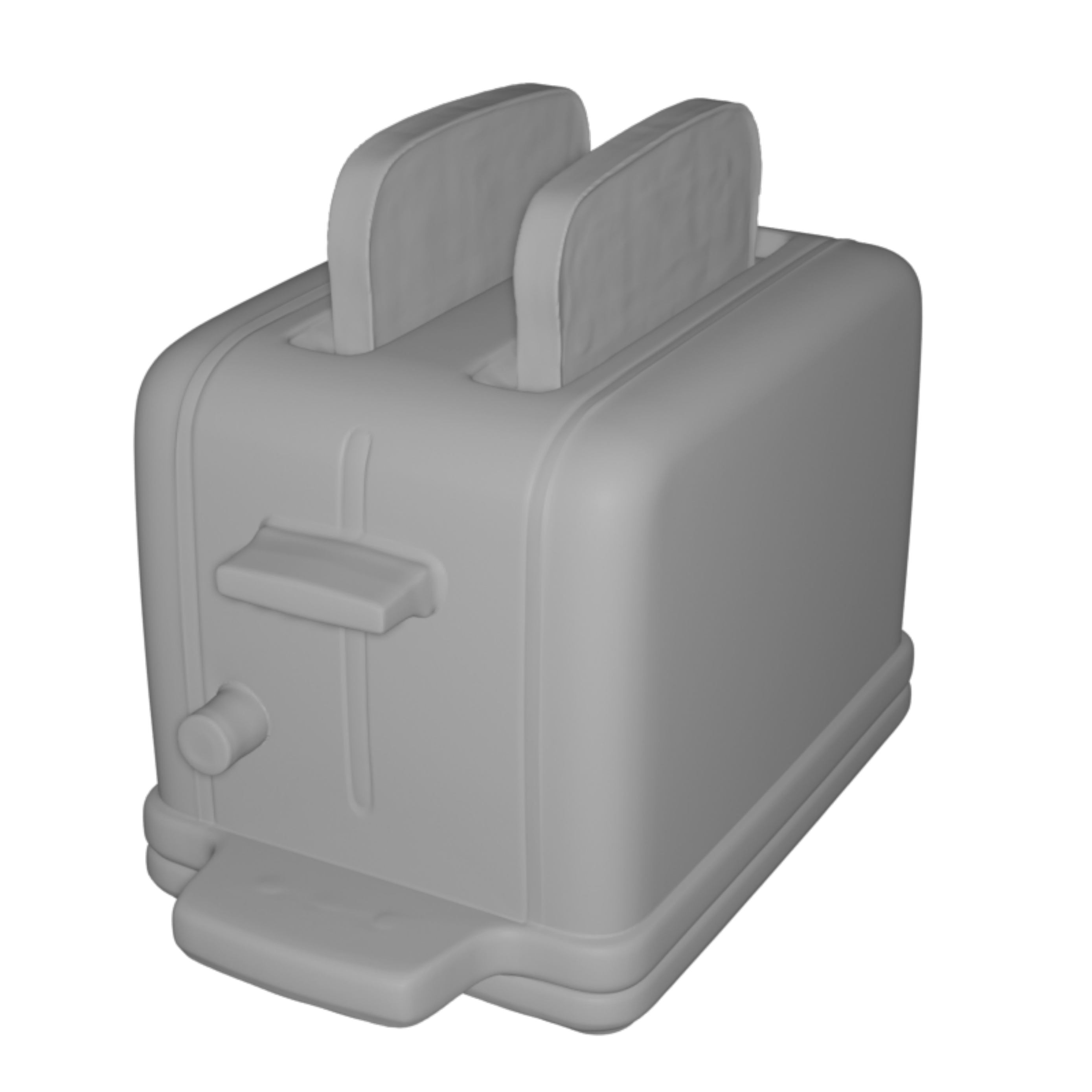}
    \includegraphics[width=0.16\columnwidth]{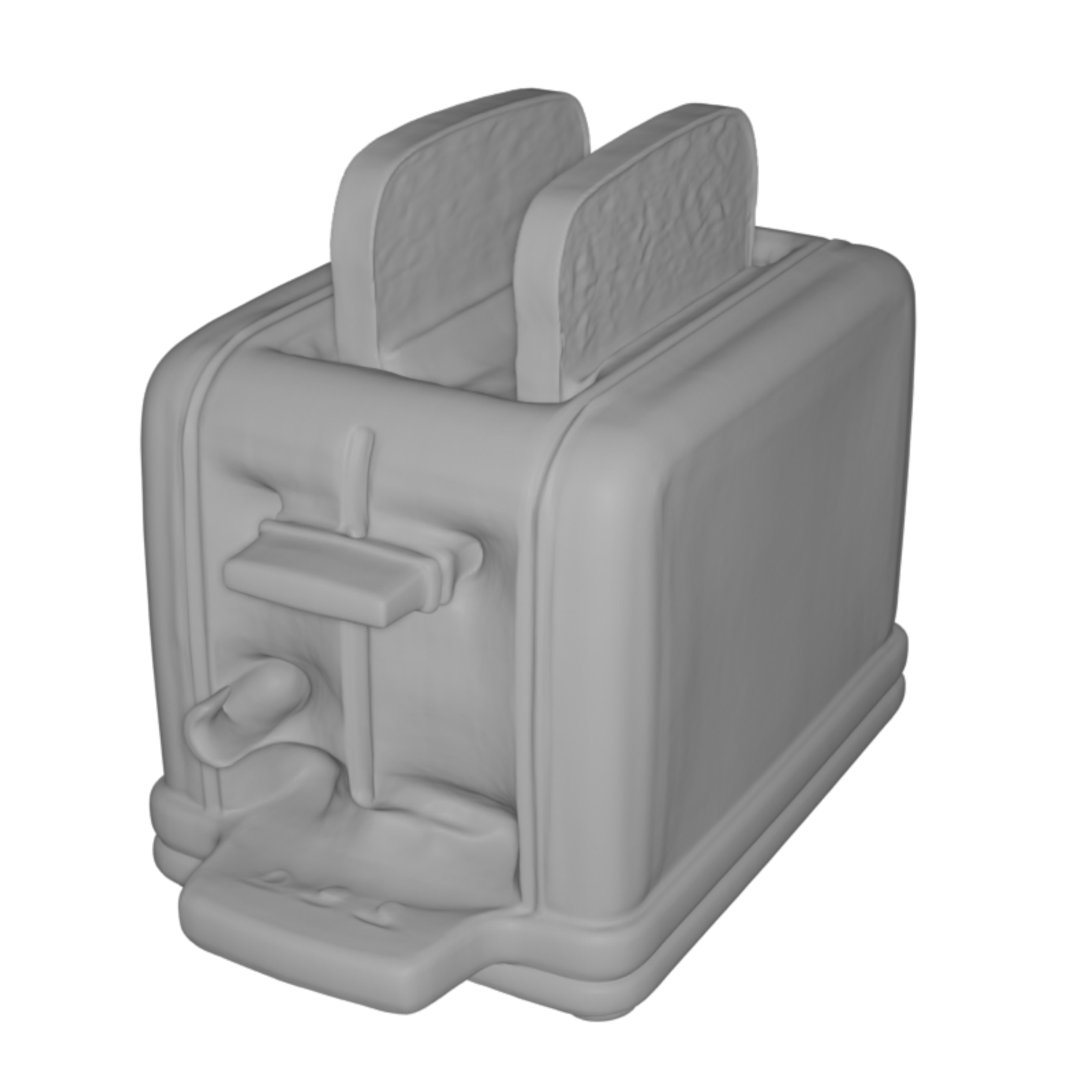}
    \includegraphics[width=0.16\columnwidth]{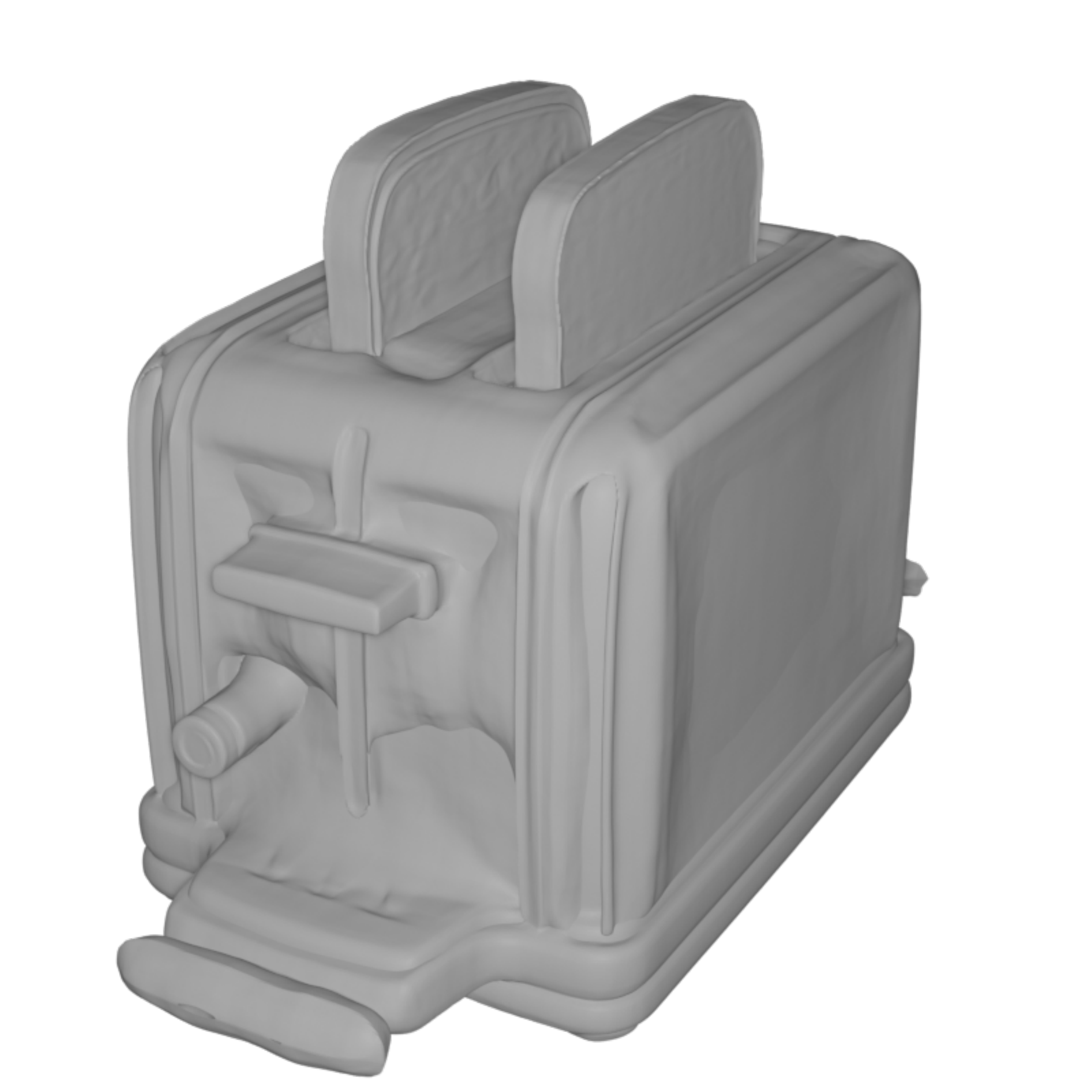}
    \includegraphics[width=0.145\columnwidth]{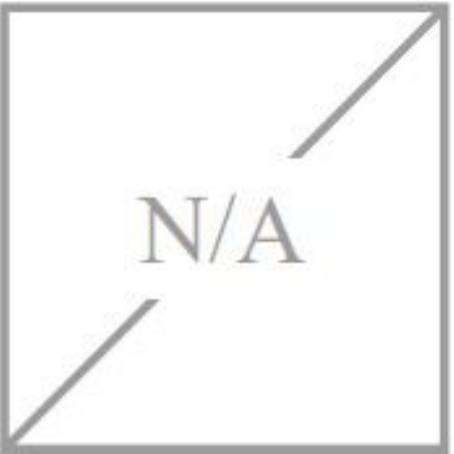}

    \end{minipage}

    \begin{minipage}{0.98\textwidth}
    \centering
     \includegraphics[width=0.16\columnwidth]{figure/r_58.pdf}
    \includegraphics[width=0.16\columnwidth]{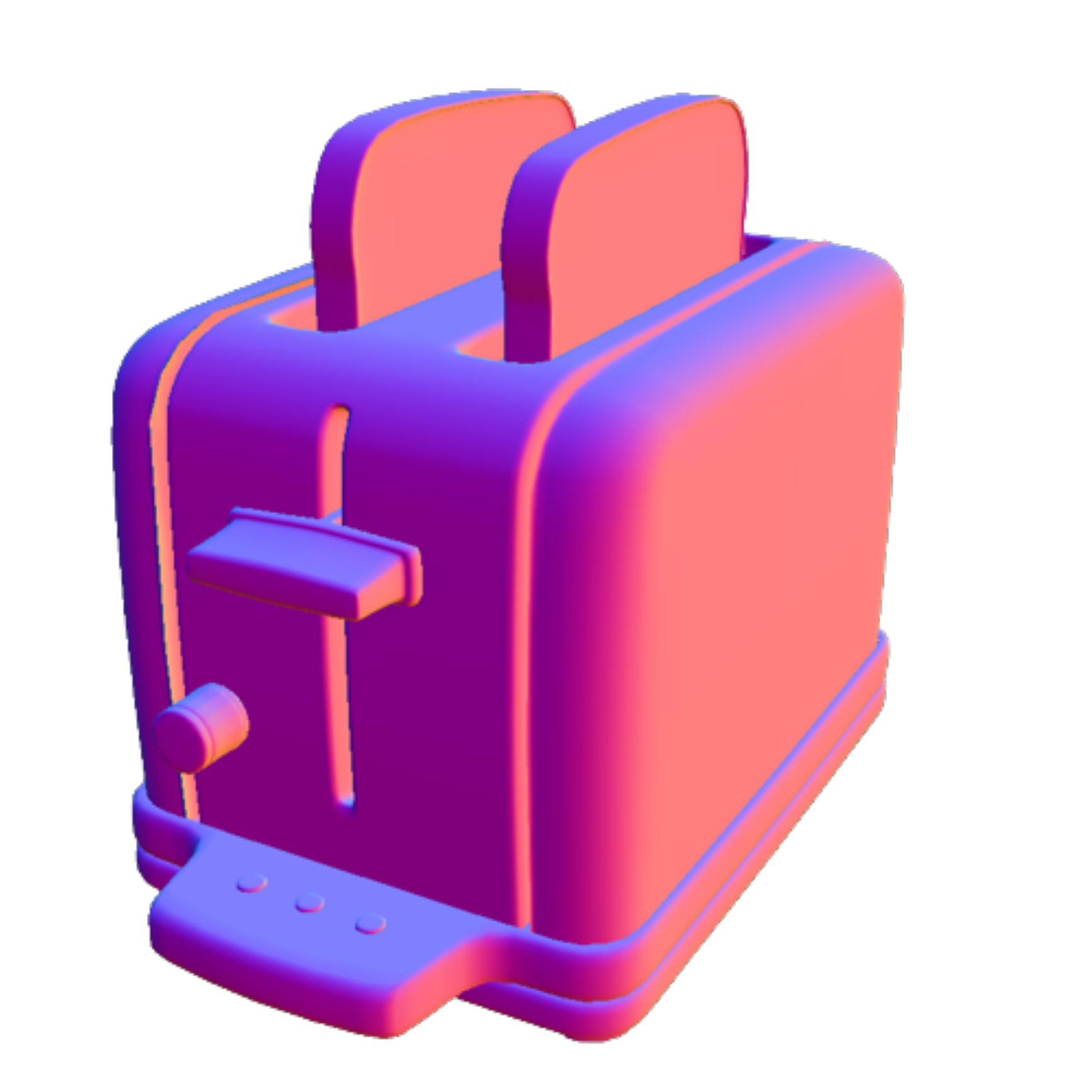}
    \includegraphics[width=0.16\columnwidth]{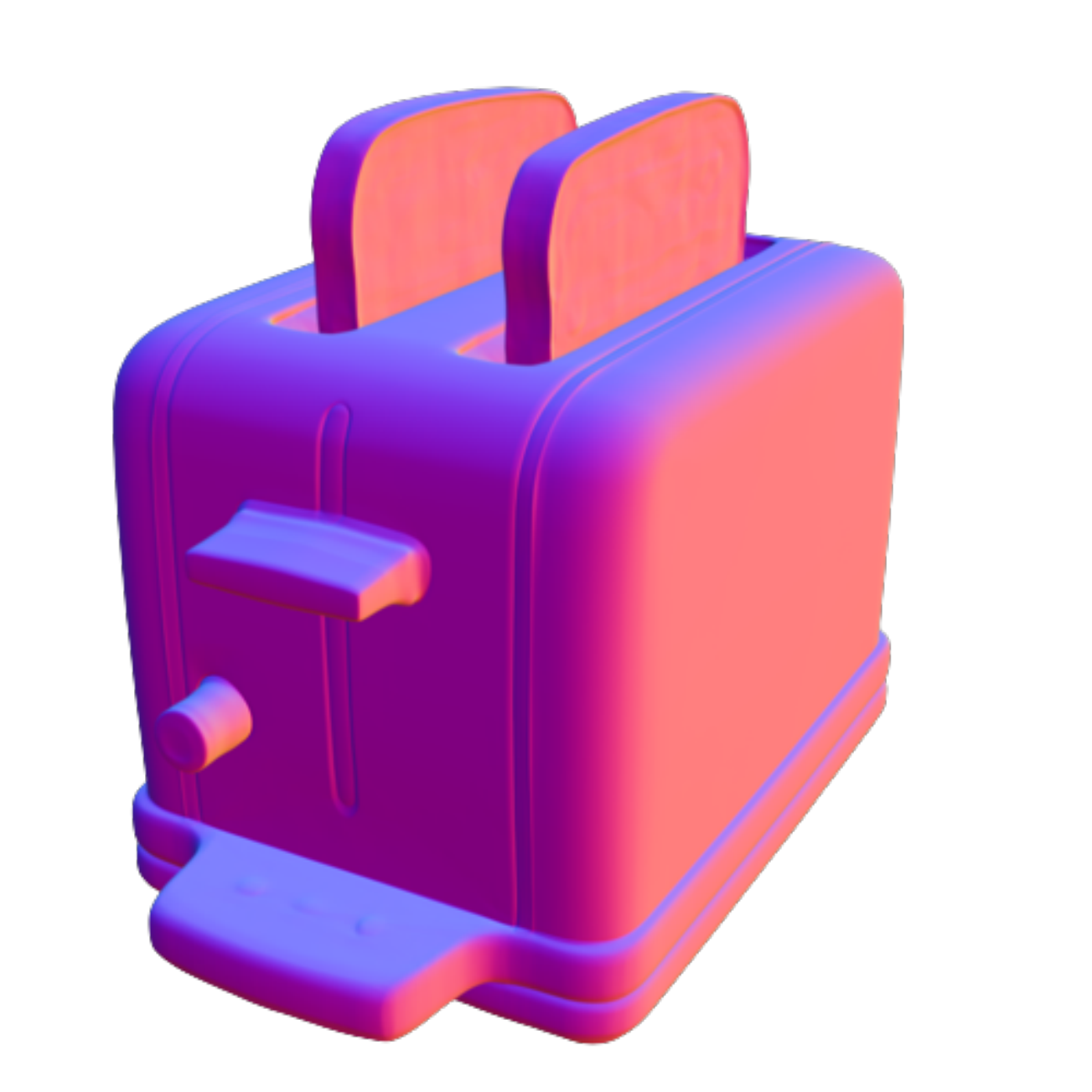}
    \includegraphics[width=0.16\columnwidth]{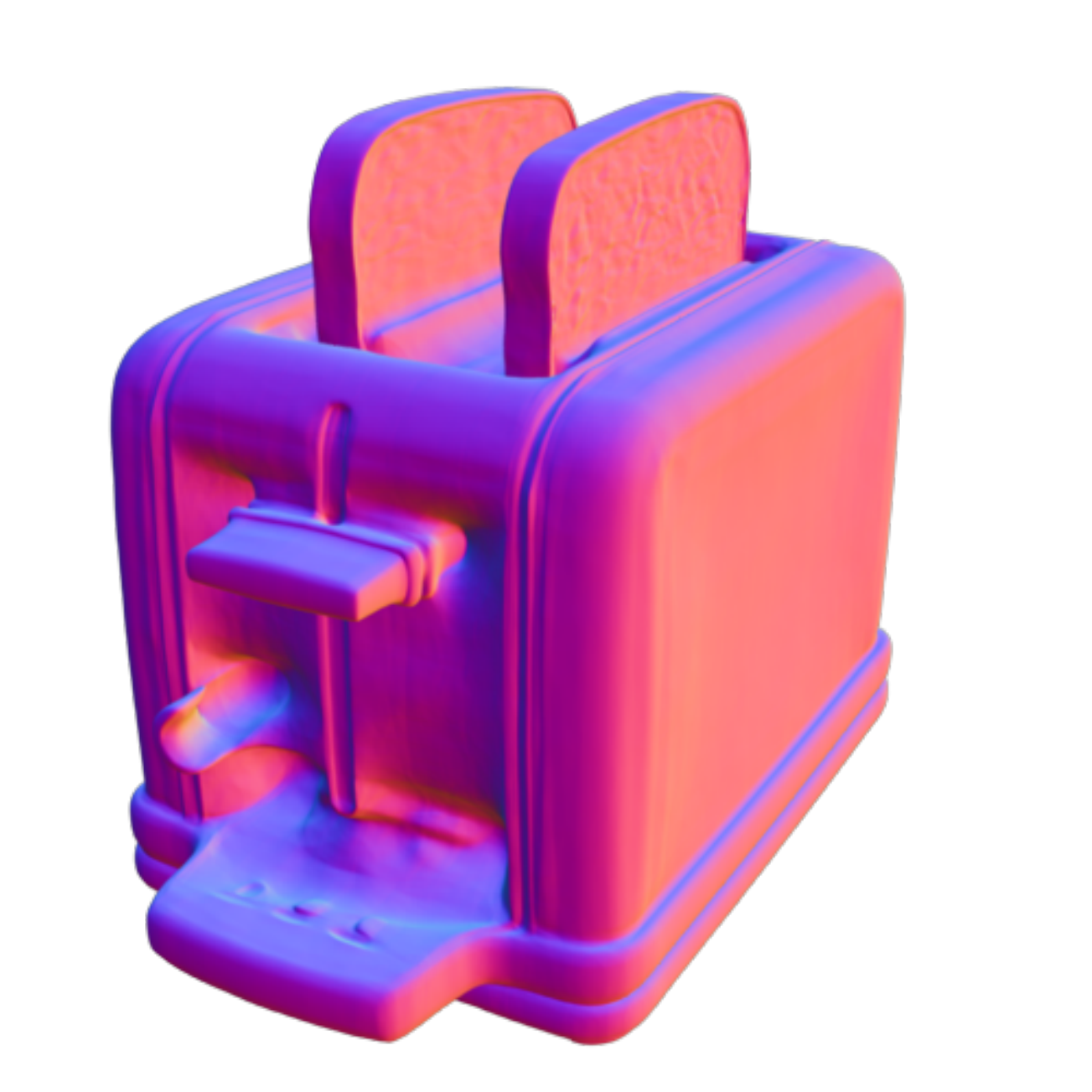}
    \includegraphics[width=0.16\columnwidth]{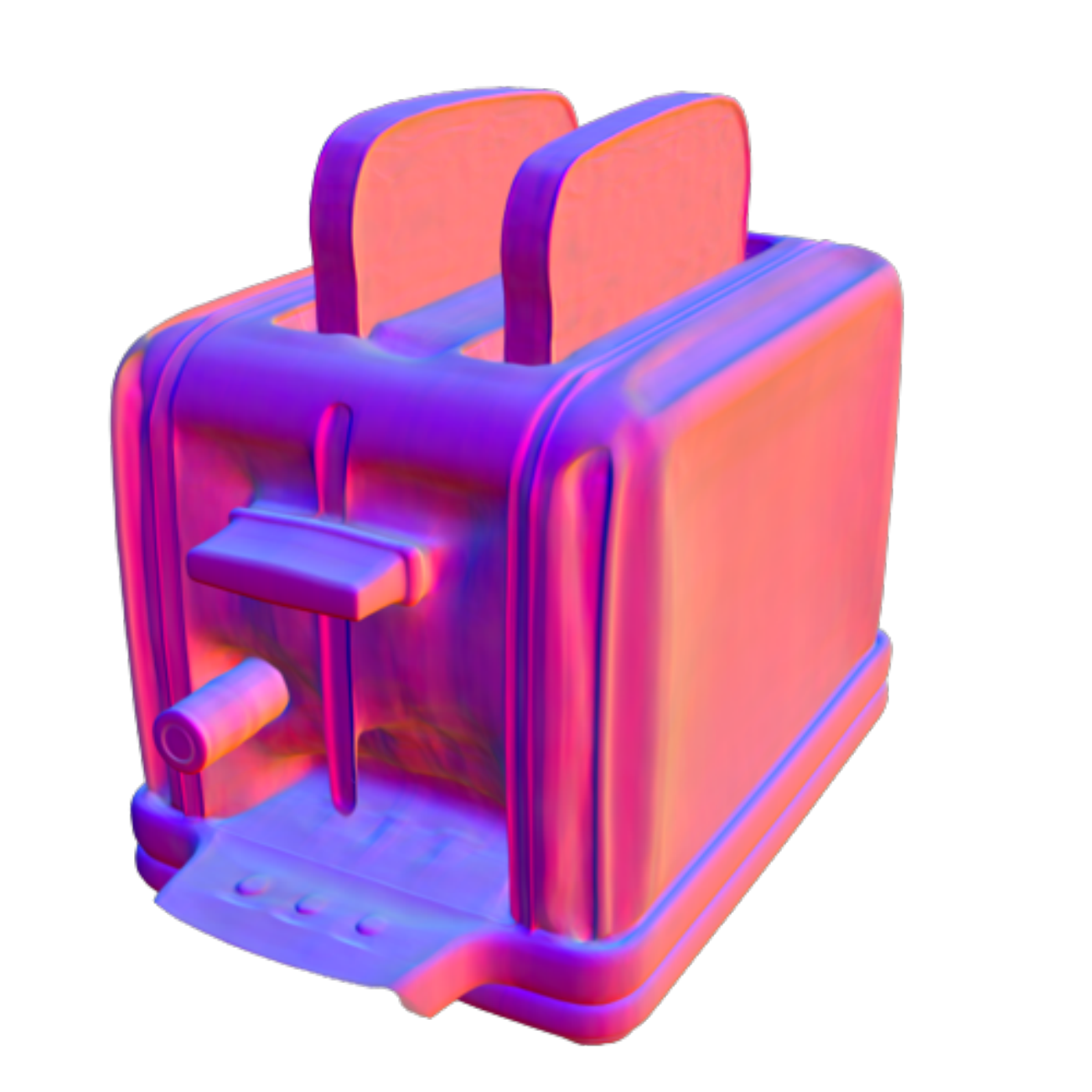}
    \includegraphics[width=0.16\columnwidth]{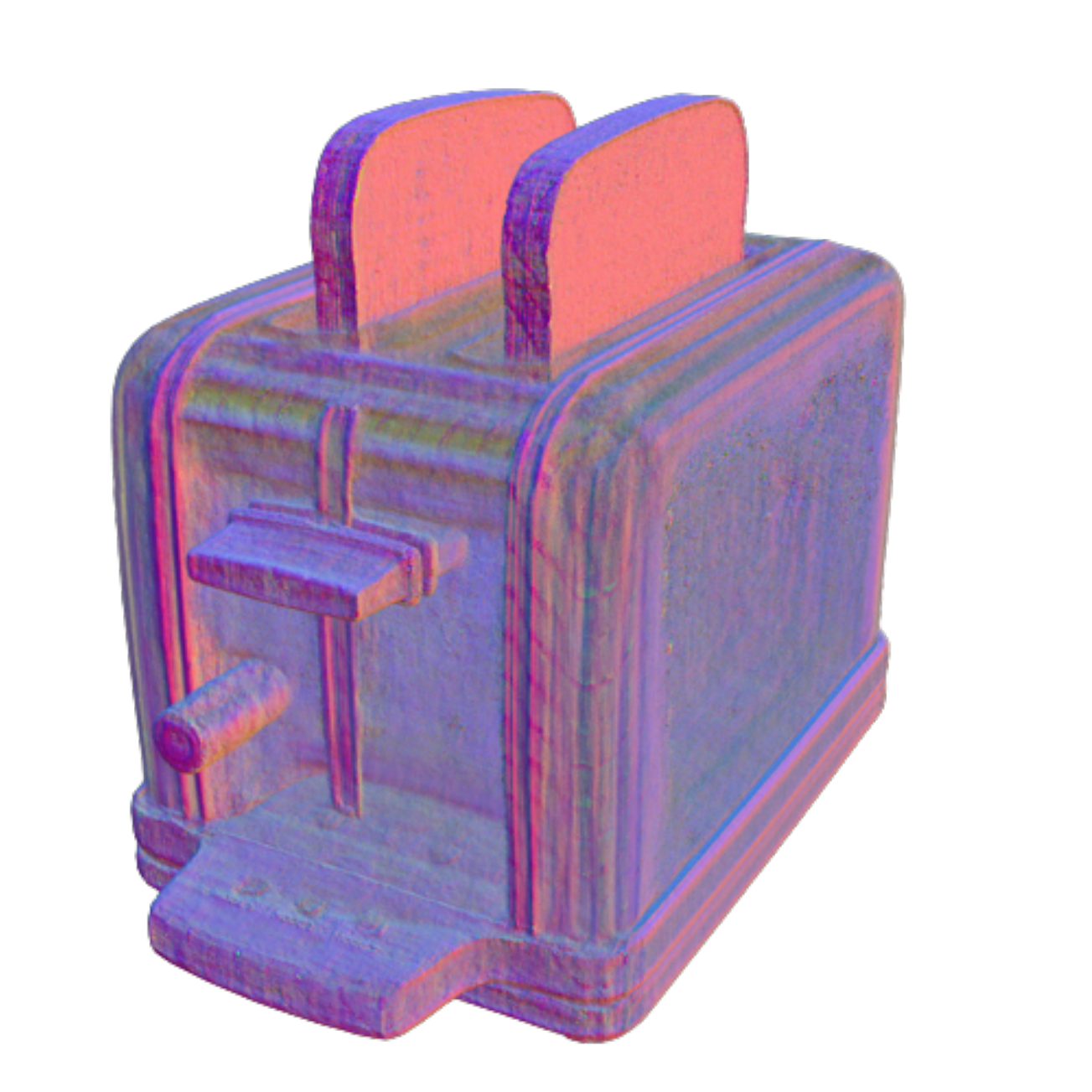}
    \end{minipage}

    \begin{minipage}{0.98\textwidth}
    \centering
     \includegraphics[width=0.16\columnwidth]{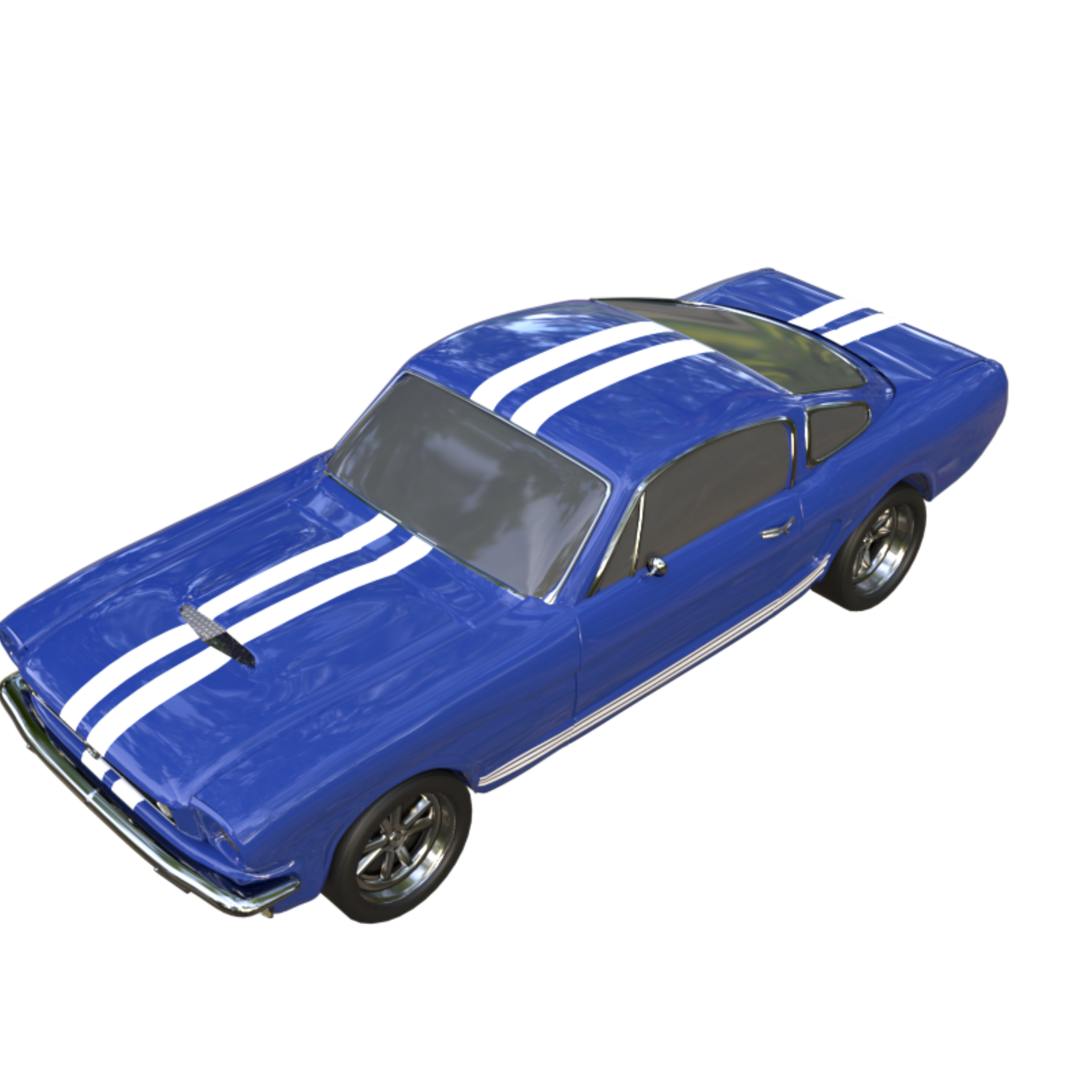}
    \includegraphics[width=0.16\columnwidth]{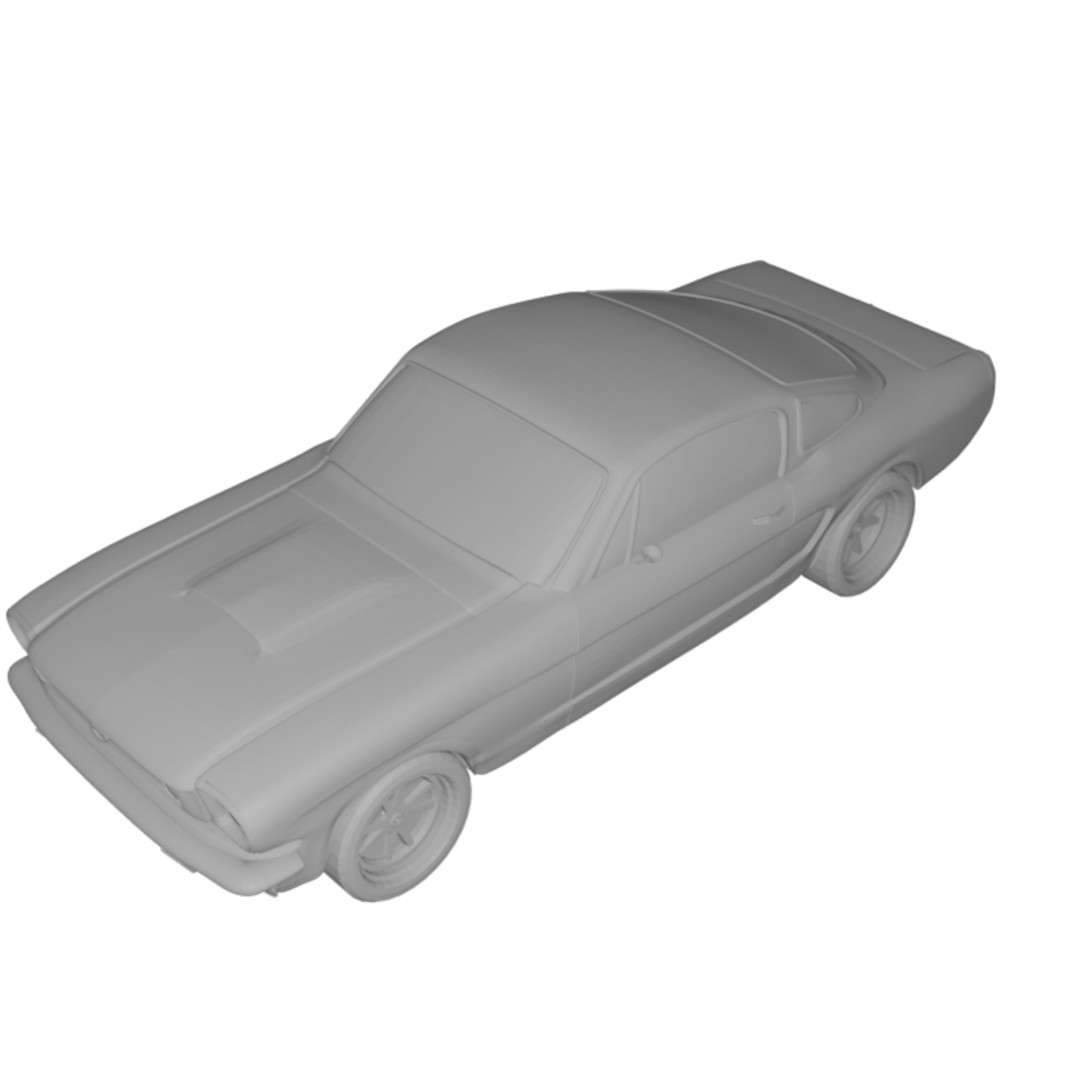}
    \includegraphics[width=0.16\columnwidth]{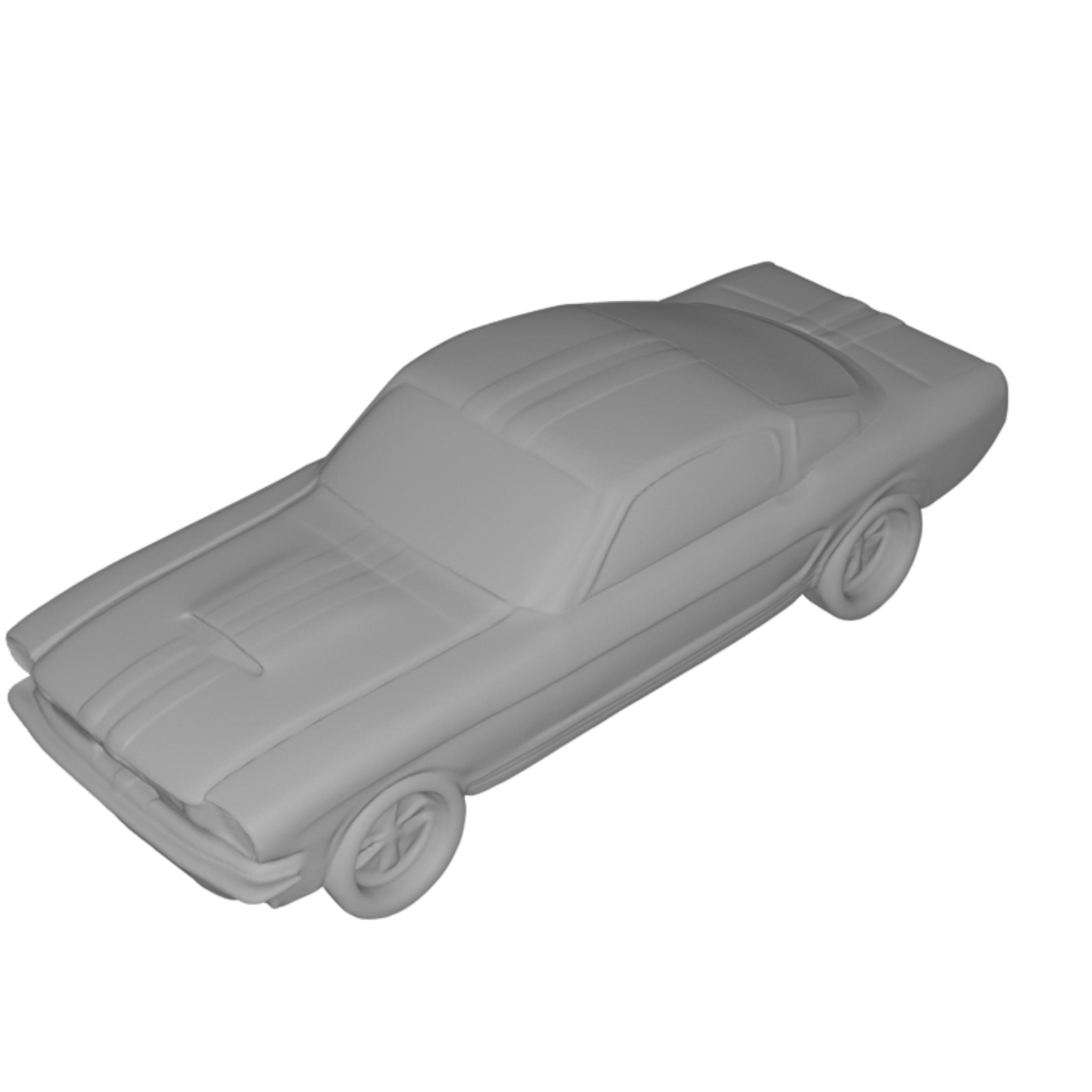}
    \includegraphics[width=0.16\columnwidth]{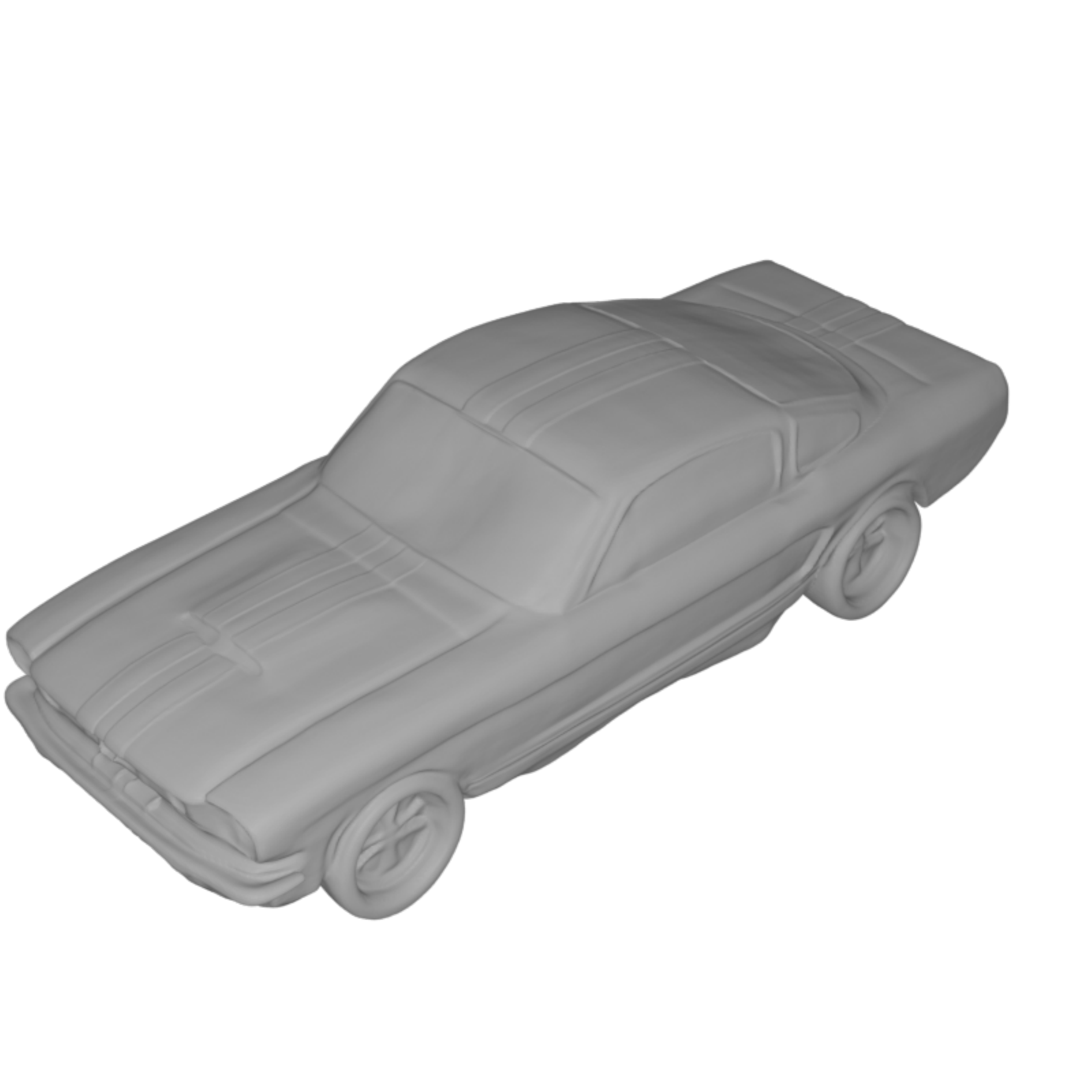}
    \includegraphics[width=0.16\columnwidth]{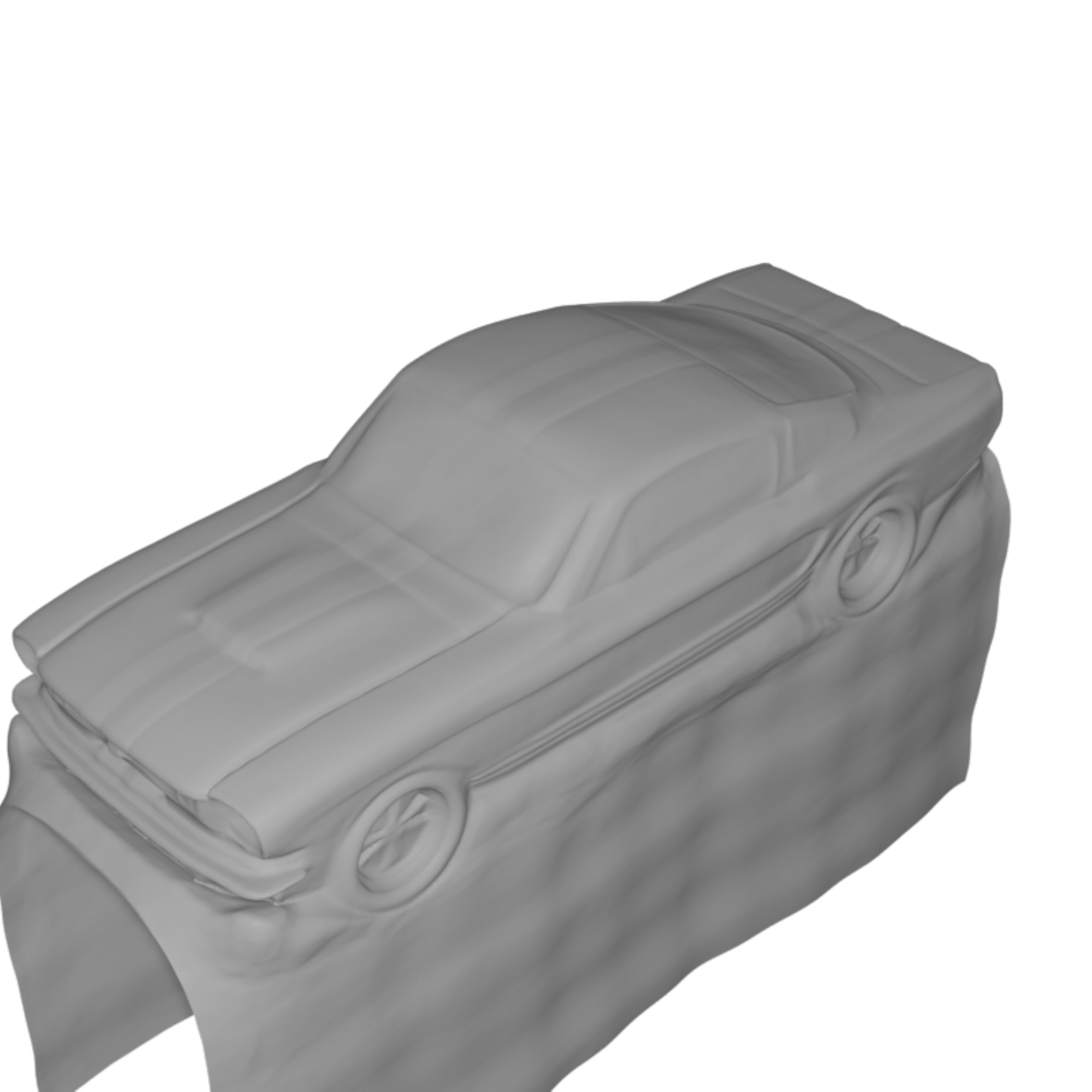}
    \includegraphics[width=0.145\columnwidth]{figure/NA.pic.pdf}
    \end{minipage}
    
    \begin{minipage}{0.16\textwidth}
    \centering
    \includegraphics[width=\columnwidth]{figure/r_163.pdf}
    \centerline{\scriptsize Reference Image}
    \end{minipage}
    \begin{minipage}{0.16\textwidth}
    \centering
    \includegraphics[width=\columnwidth]{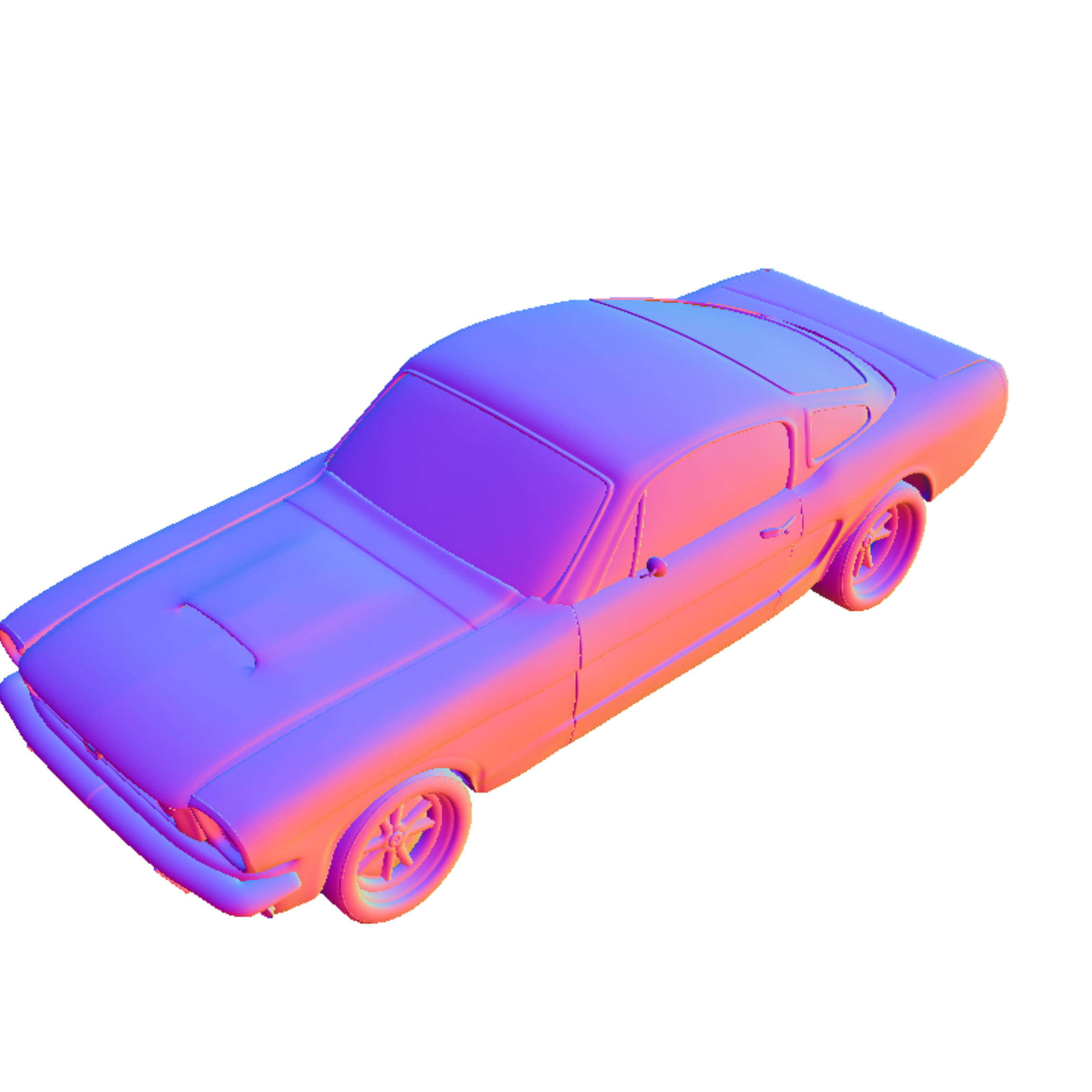}
    \centerline{\scriptsize GT meshes/normals}
    \end{minipage}
    \begin{minipage}{0.16\textwidth}
    \centering
    \includegraphics[width=\columnwidth]{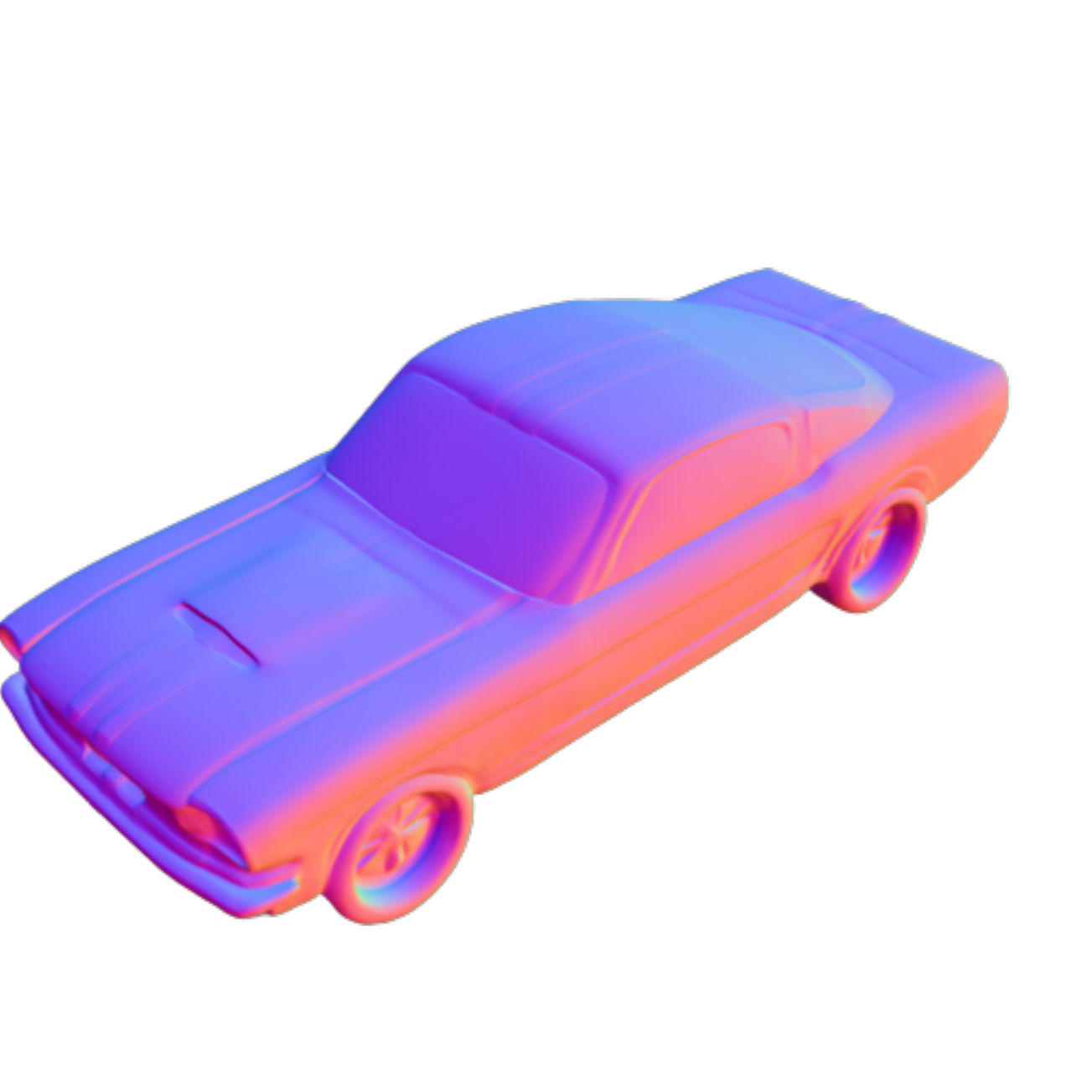}
    \centerline{\scriptsize Ours}
    \end{minipage}
    \begin{minipage}{0.16\textwidth}
    \centering
    \includegraphics[width=\columnwidth]{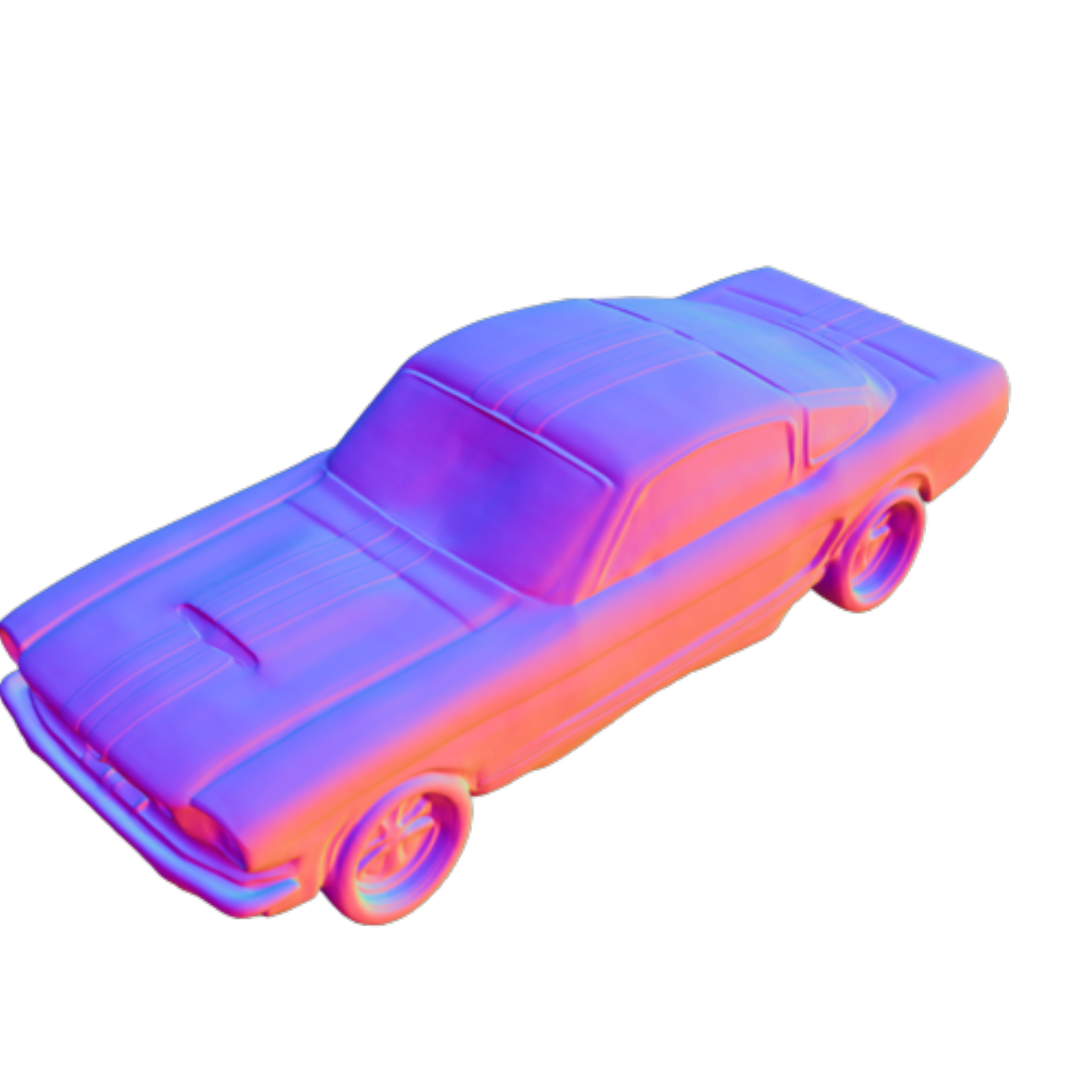}
    \centerline{\scriptsize NeuS}
    \end{minipage}
    \begin{minipage}{0.16\textwidth}
    \centering
    \includegraphics[width=\columnwidth]{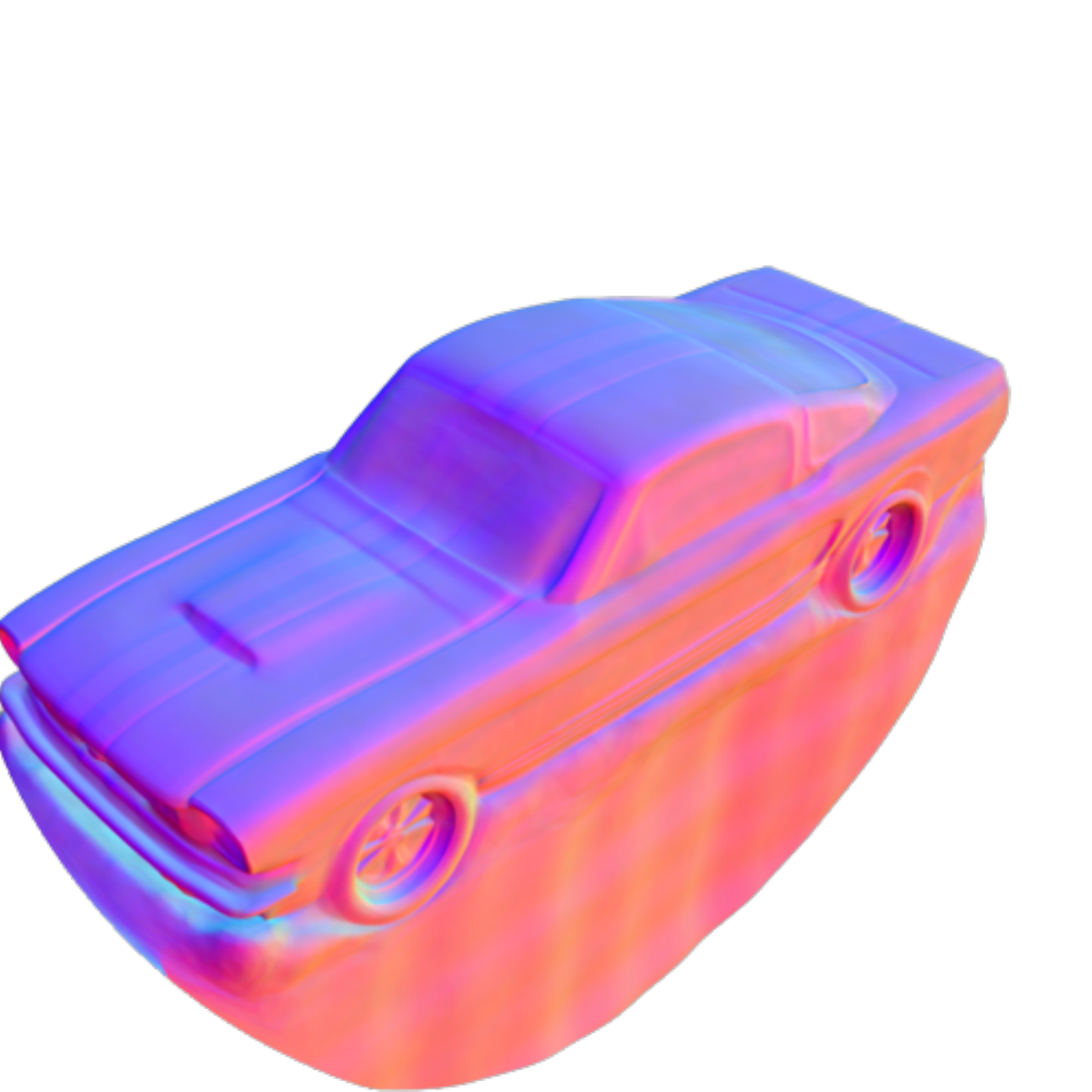}
    \centerline{\scriptsize Geo-NeuS}
    \end{minipage}
    \begin{minipage}{0.16\textwidth}
    \centering
    \includegraphics[width=\columnwidth]{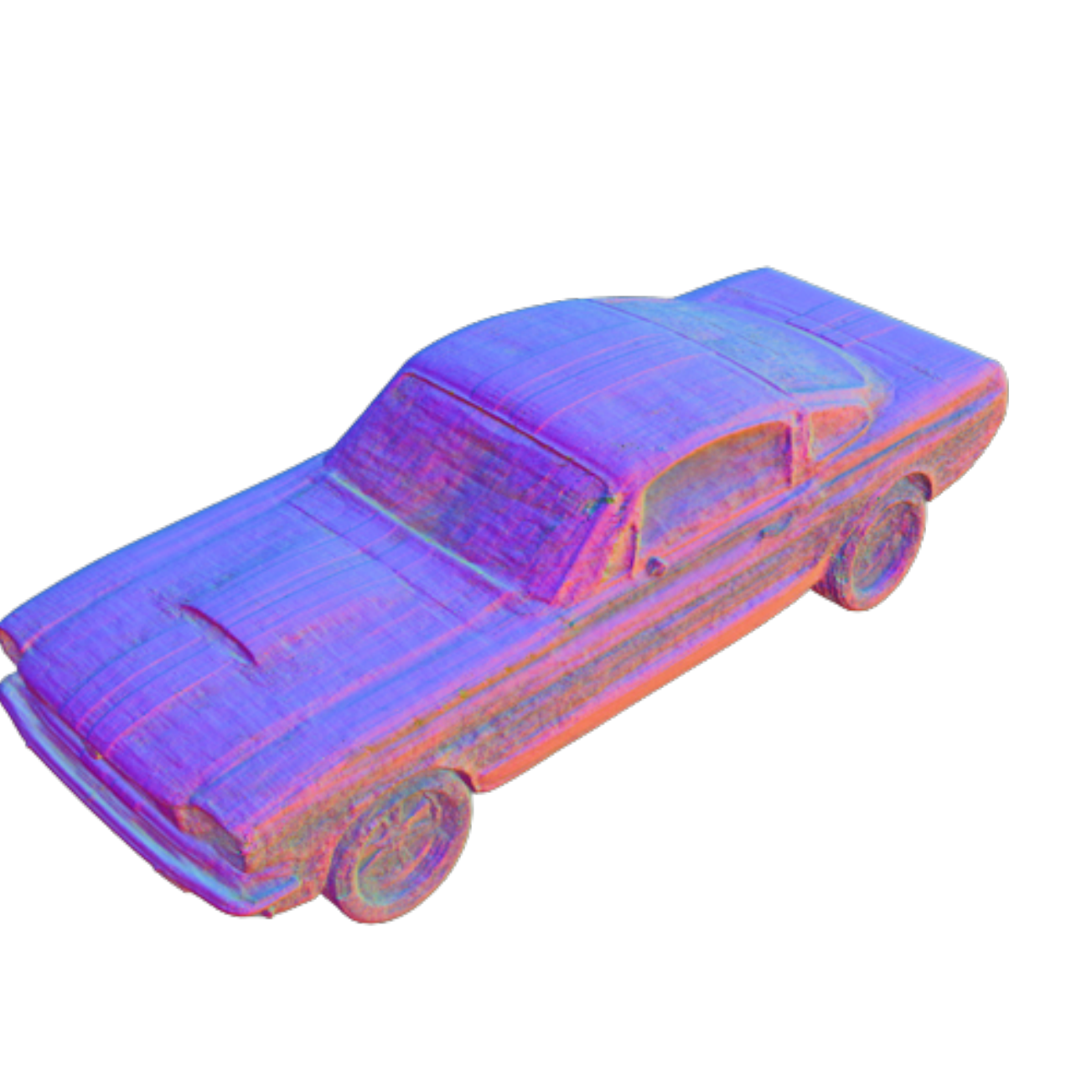}
    \centerline{\scriptsize Ref-NeRF}
    \end{minipage}
    
    \vspace{0.1cm}
    \caption{ The geometry of reconstructed meshes and estimated surface normals on Shiny Blender dataset \cite{verbin2022ref}. We ran NeuS, Geo-Neus and Ref-NeRF official implementations. Our method obviously produces better geometry and surface normals than other methods. }
    \label{toaster-comp}
    \vspace{-0.5cm}
\end{figure*}

For PhySG and Ref-NeRF, we obtained the MAE from the Ref-NeRF paper, as these methods focus on novel view synthesis, the reconstruction accuracy is not reported in Ref-NeRF paper.
For other methods, we implemented the released code on the corresponding datasets. IDR \cite{IDR} was unable to generate meaningful reconstructions for the helmet and toaster objects, so we only reported the results for the car and coffee objects. Our method significantly outperformed all other compared methods by a large margin.
As can be seen qualitatively in Figures \ref{demo} and \ref{toaster-comp}, our method yields promising improvements in both geometry accuracy and surface normals. Ref-NeRF adopts Integrated Positional Encoding similar to mip-NeRF \cite{barron2021mip}, which makes its geometry parameters dependent on view direction, making the meshes inaccessible. 
Furthermore, our results achieve higher rendering quality (PSNR) compared to Ref-NeRF except for a failure case as reported in Table \ref{psnr}. This indicates that more accurate surface normals can lead to improved novel view synthesis quality since the conditioned reflection direction is calculated more accurately compared to Ref-NeRF. 
Note that, due to memory limitations, we only sampled 1024 rays instead of 4096 $\times$ 4 as in Ref-NeRF.

\subsection{Ablation Study}
We conducted an ablation study on the Shiny Blender dataset to evaluate the effectiveness of each component in our model. The experimental results are reported in Table \ref{ablation}.
``NeuS w/ RS'' indicates that we computed the reflection-aware photometric loss with the reflection score estimated by Eq. \eqref{ucc_view}, without accounting for visibility. This leads to a smaller improvement in performance, suggesting that using the reflection score as a source of variance can still be beneficial in improving geometry.
``NeuS w/ Ref'' only replaces the radiance dependency with the reflection direction without considering the reflection score. The performance improved over the baseline, except for one failure case, which we discussed and provided per-scene results for in the supplementary material.
``Ref-NeuS w/o Ref'' indicates that we further identified visibility for reflection score estimation using Eq. \eqref{ucc_form}. The performance significantly improved, highlighting the importance of an unbiased reflection score in accurate geometry recovery.
Finally, we used both reflection-aware photometric loss and reflection direction-dependent radiance in our model. The performance significantly improved compared to using only one of these methods, indicating that both methods are complementary for achieving accurate surface reconstruction.

\section{Limitation and Conclusion}
\noindent\textbf{Limitation.}
Although our method shows promising results in multi-view reconstruction with reflection, several limitations remain. Firstly, estimating the reflection score inevitably increases the computational cost. Secondly, simply replacing the dependency of the radiance network with the reflection direction, regardless of the object material, can lead to artifacts in certain circumstances. 
We present an example of such an artifact in the supplementary material.

\vspace{0.1cm}
\noindent\textbf{Conclusion.}
In this paper, we investigate the issue of multi-view reconstruction for objects with reflective surfaces, which is an important but under-explored problem. Reflection-induced ambiguity can significantly disrupt multi-view consistency, but our proposed Ref-NeuS method addresses this issue by introducing a reflection-aware photometric loss, where the importance of reflective pixels is attenuated using a Gaussian distribution model. Additionally, our method employs reflection direction-dependent radiance, which further improves the geometry with a better radiance fields, including geometry and surface normals.

{\small
\bibliographystyle{ieee_fullname}
\bibliography{main}

\begin{thebibliography}{10}\itemsep=-1pt

\bibitem{dtu_eval}
Henrik Aan{\ae}s, Rasmus~Ramsb{\o}l Jensen, George Vogiatzis, Engin Tola, and
  Anders~Bjorholm Dahl.
\newblock Large-scale data for multiple-view stereopsis.
\newblock {\em International Journal of Computer Vision (IJCV)}, 2016.

\bibitem{barron2021mip}
Jonathan~T Barron, Ben Mildenhall, Matthew Tancik, Peter Hedman, Ricardo
  Martin-Brualla, and Pratul~P Srinivasan.
\newblock Mip-nerf: A multiscale representation for anti-aliasing neural
  radiance fields.
\newblock In {\em Proceedings of the IEEE/CVF International Conference on
  Computer Vision (ICCV)}, 2021.

\bibitem{boss2021nerd}
Mark Boss, Raphael Braun, Varun Jampani, Jonathan~T Barron, Ce Liu, and Hendrik
  Lensch.
\newblock Nerd: Neural reflectance decomposition from image collections.
\newblock In {\em Proceedings of the IEEE/CVF International Conference on
  Computer Vision (CVPR)}, 2021.

\bibitem{campbell2008using}
Neill~DF Campbell, George Vogiatzis, Carlos Hern{\'a}ndez, and Roberto Cipolla.
\newblock Using multiple hypotheses to improve depth-maps for multi-view
  stereo.
\newblock In {\em European Conference on Computer Vision (ECCV)}, 2008.

\bibitem{cheng2021multi}
Ziang Cheng, Hongdong Li, Yuta Asano, Yinqiang Zheng, and Imari Sato.
\newblock Multi-view 3d reconstruction of a texture-less smooth surface of
  unknown generic reflectance.
\newblock In {\em Proceedings of the IEEE/CVF Conference on Computer Vision and
  Pattern Recognition (CVPR)}, 2021.

\bibitem{curless1996volumetric}
Brian Curless and Marc Levoy.
\newblock A volumetric method for building complex models from range images.
\newblock In {\em Proceedings of the 23rd annual conference on Computer
  graphics and interactive techniques}, 1996.

\bibitem{darmon2022improving}
Fran{\c{c}}ois Darmon, B{\'e}n{\'e}dicte Bascle, Jean-Cl{\'e}ment Devaux,
  Pascal Monasse, and Mathieu Aubry.
\newblock Improving neural implicit surfaces geometry with patch warping.
\newblock In {\em Proceedings of the IEEE/CVF Conference on Computer Vision and
  Pattern Recognition (CVPR)}, 2022.

\bibitem{falivene2019towards}
Laura Falivene, Zhen Cao, Andrea Petta, Luigi Serra, Albert Poater, Romina
  Oliva, Vittorio Scarano, and Luigi Cavallo.
\newblock Towards the online computer-aided design of catalytic pockets.
\newblock {\em Nature Chemistry}, 2019.

\bibitem{fu2022geo}
Qiancheng Fu, Qingshan Xu, Yew-Soon Ong, and Wenbing Tao.
\newblock Geo-neus: Geometry-consistent neural implicit surfaces learning for
  multi-view reconstruction.
\newblock {\em arXiv preprint arXiv:2205.15848}, 2022.

\bibitem{fua1995object}
Pascal Fua and Yvan~G Leclerc.
\newblock Object-centered surface reconstruction: Combining multi-image stereo
  and shading.
\newblock {\em International Journal of Computer Vision (IJCV)}, 1995.

\bibitem{furukawa2009accurate}
Yasutaka Furukawa and Jean Ponce.
\newblock Accurate, dense, and robust multiview stereopsis.
\newblock {\em IEEE transactions on pattern analysis and machine intelligence
  (TPAMI)}, 2009.

\bibitem{galliani2015massively}
Silvano Galliani, Katrin Lasinger, and Konrad Schindler.
\newblock Massively parallel multiview stereopsis by surface normal diffusion.
\newblock In {\em Proceedings of the IEEE International Conference on Computer
  Vision (ICCV)}, 2015.

\bibitem{eik}
Amos Gropp, Lior Yariv, Niv Haim, Matan Atzmon, and Yaron Lipman.
\newblock Implicit geometric regularization for learning shapes.
\newblock {\em arXiv preprint arXiv:2002.10099}, 2020.

\bibitem{guo2014robust}
Xiaojie Guo, Xiaochun Cao, and Yi Ma.
\newblock Robust separation of reflection from multiple images.
\newblock In {\em Proceedings of the IEEE/CVF Conference on Computer Vision and
  Pattern Recognition (CVPR)}, 2014.

\bibitem{ji2017surfacenet}
Mengqi Ji, Juergen Gall, Haitian Zheng, Yebin Liu, and Lu Fang.
\newblock Surfacenet: An end-to-end 3d neural network for multiview stereopsis.
\newblock In {\em Proceedings of the IEEE/CVF International Conference on
  Computer Vision (ICCV)}, 2017.

\bibitem{kajiya1984ray}
James~T Kajiya and Brian~P Von~Herzen.
\newblock Ray tracing volume densities.
\newblock {\em ACM SIGGRAPH computer graphics}, 1984.

\bibitem{kar2017learning}
Abhishek Kar, Christian H{\"a}ne, and Jitendra Malik.
\newblock Learning a multi-view stereo machine.
\newblock {\em Advances in neural information processing systems (NeurIPS)},
  2017.

\bibitem{kazhdan2013screened}
Michael Kazhdan and Hugues Hoppe.
\newblock Screened poisson surface reconstruction.
\newblock {\em ACM Transactions on Graphics (ToG)}, 2013.

\bibitem{kendall2017uncertainties}
Alex Kendall and Yarin Gal.
\newblock What uncertainties do we need in bayesian deep learning for computer
  vision?
\newblock {\em Advances in neural information processing systems (NeurIPS)},
  2017.

\bibitem{kennard1969computer}
Ronald~W Kennard and Larry~A Stone.
\newblock Computer aided design of experiments.
\newblock {\em Technometrics}, 1969.

\bibitem{kingma2014adam}
Diederik~P Kingma and Jimmy Ba.
\newblock Adam: A method for stochastic optimization.
\newblock {\em arXiv preprint arXiv:1412.6980}, 2014.

\bibitem{labatut2007efficient}
Patrick Labatut, Jean-Philippe Pons, and Renaud Keriven.
\newblock Efficient multi-view reconstruction of large-scale scenes using
  interest points, delaunay triangulation and graph cuts.
\newblock In {\em Proceedings of the IEEE/CVF International Conference on
  Computer Vision (ICCV)}, 2007.

\bibitem{lasseter1987principles}
John Lasseter.
\newblock Principles of traditional animation applied to 3d computer animation.
\newblock In {\em Proceedings of the 14th annual conference on Computer
  graphics and interactive techniques}, 1987.

\bibitem{lhuillier2005quasi}
Maxime Lhuillier and Long Quan.
\newblock A quasi-dense approach to surface reconstruction from uncalibrated
  images.
\newblock {\em IEEE transactions on pattern analysis and machine intelligence
  (TPAMI)}, 2005.

\bibitem{lorensen1987marching}
William~E Lorensen and Harvey~E Cline.
\newblock Marching cubes: A high resolution 3d surface construction algorithm.
\newblock {\em ACM siggraph computer graphics}, 1987.

\bibitem{mahalanobis1936generalised}
Prasanta~Chandra Mahalanobis.
\newblock On the generalised distance in statistics.
\newblock In {\em Proceedings of the national Institute of Science of India},
  1936.

\bibitem{martin2021nerfw}
Ricardo Martin-Brualla, Noha Radwan, Mehdi~SM Sajjadi, Jonathan~T Barron,
  Alexey Dosovitskiy, and Daniel Duckworth.
\newblock Nerf in the wild: Neural radiance fields for unconstrained photo
  collections.
\newblock In {\em Proceedings of the IEEE/CVF Conference on Computer Vision and
  Pattern Recognition (CVPR)}, 2021.

\bibitem{max1995optical}
Nelson Max.
\newblock Optical models for direct volume rendering.
\newblock {\em IEEE Transactions on Visualization and Computer Graphics (VCG)},
  1995.

\bibitem{mescheder2019occupancy}
Lars Mescheder, Michael Oechsle, Michael Niemeyer, Sebastian Nowozin, and
  Andreas Geiger.
\newblock Occupancy networks: Learning 3d reconstruction in function space.
\newblock In {\em Proceedings of the IEEE/CVF conference on computer vision and
  pattern recognition (CVPR)}, 2019.

\bibitem{mildenhall2021nerf}
Ben Mildenhall, Pratul~P Srinivasan, Matthew Tancik, Jonathan~T Barron, Ravi
  Ramamoorthi, and Ren Ng.
\newblock Nerf: Representing scenes as neural radiance fields for view
  synthesis.
\newblock {\em Communications of the ACM}, 2021.

\bibitem{oechsle2021unisurf}
Michael Oechsle, Songyou Peng, and Andreas Geiger.
\newblock Unisurf: Unifying neural implicit surfaces and radiance fields for
  multi-view reconstruction.
\newblock In {\em Proceedings of the IEEE/CVF International Conference on
  Computer Vision (ICCV)}, 2021.

\bibitem{pan2022activenerf}
Xuran Pan, Zihang Lai, Shiji Song, and Gao Huang.
\newblock Activenerf: Learning where to see with uncertainty estimation.
\newblock In {\em European conference on computer vision (ECCV)}, 2022.

\bibitem{parent2012computer}
Rick Parent.
\newblock {\em Computer animation: algorithms and techniques}.
\newblock Newnes, 2012.

\bibitem{park2019deepsdf}
Jeong~Joon Park, Peter Florence, Julian Straub, Richard Newcombe, and Steven
  Lovegrove.
\newblock Deepsdf: Learning continuous signed distance functions for shape
  representation.
\newblock In {\em Proceedings of the IEEE/CVF conference on computer vision and
  pattern recognition (CVPR)}, 2019.

\bibitem{park2020seeing}
Jeong~Joon Park, Aleksander Holynski, and Steven~M Seitz.
\newblock Seeing the world in a bag of chips.
\newblock In {\em Proceedings of the IEEE/CVF Conference on Computer Vision and
  Pattern Recognition (CVPR)}, 2020.

\bibitem{peng2020convolutional}
Songyou Peng, Michael Niemeyer, Lars Mescheder, Marc Pollefeys, and Andreas
  Geiger.
\newblock Convolutional occupancy networks.
\newblock In {\em European Conference on Computer Vision (ECCV)}, 2020.

\bibitem{roth1982ray}
Scott~D Roth.
\newblock Ray casting for modeling solids.
\newblock {\em Computer graphics and image processing}, 1982.

\bibitem{schonberger2016structure}
Johannes~L Schonberger and Jan-Michael Frahm.
\newblock Structure-from-motion revisited.
\newblock In {\em Proceedings of the IEEE/CVF conference on computer vision and
  pattern recognition (CVPR)}, 2016.

\bibitem{schonberger2016pixelwise}
Johannes~L Sch{\"o}nberger, Enliang Zheng, Jan-Michael Frahm, and Marc
  Pollefeys.
\newblock Pixelwise view selection for unstructured multi-view stereo.
\newblock In {\em European conference on computer vision (ECCV)}, 2016.

\bibitem{schuemie2001research}
Martijn~J Schuemie, Peter Van Der~Straaten, Merel Krijn, and Charles~APG Van
  Der~Mast.
\newblock Research on presence in virtual reality: A survey.
\newblock {\em Cyberpsychology \& behavior}, 2001.

\bibitem{seitz1999photorealistic}
Steven~M Seitz and Charles~R Dyer.
\newblock Photorealistic scene reconstruction by voxel coloring.
\newblock {\em International Journal of Computer Vision (IJCV)}, 1999.

\bibitem{srinivasan2021nerv}
Pratul~P Srinivasan, Boyang Deng, Xiuming Zhang, Matthew Tancik, Ben
  Mildenhall, and Jonathan~T Barron.
\newblock Nerv: Neural reflectance and visibility fields for relighting and
  view synthesis.
\newblock In {\em Proceedings of the IEEE/CVF Conference on Computer Vision and
  Pattern Recognition (CVPR)}, 2021.

\bibitem{verbin2022ref}
Dor Verbin, Peter Hedman, Ben Mildenhall, Todd Zickler, Jonathan~T Barron, and
  Pratul~P Srinivasan.
\newblock Ref-nerf: Structured view-dependent appearance for neural radiance
  fields.
\newblock In {\em Proceedings of the IEEE/CVF Conference on Computer Vision and
  Pattern Recognition (CVPR)}, 2022.

\bibitem{wang2021patchmatchnet}
Fangjinhua Wang, Silvano Galliani, Christoph Vogel, Pablo Speciale, and Marc
  Pollefeys.
\newblock Patchmatchnet: Learned multi-view patchmatch stereo.
\newblock In {\em Proceedings of the IEEE/CVF Conference on Computer Vision and
  Pattern Recognition (CVPR)}, 2021.

\bibitem{wang2021neus}
Peng Wang, Lingjie Liu, Yuan Liu, Christian Theobalt, Taku Komura, and Wenping
  Wang.
\newblock Neus: Learning neural implicit surfaces by volume rendering for
  multi-view reconstruction.
\newblock {\em arXiv preprint arXiv:2106.10689}, 2021.

\bibitem{wood2000surface}
Daniel~N Wood, Daniel~I Azuma, Ken Aldinger, Brian Curless, Tom Duchamp,
  David~H Salesin, and Werner Stuetzle.
\newblock Surface light fields for 3d photography.
\newblock In {\em Proceedings of the 27th annual conference on Computer
  graphics and interactive techniques}, 2000.

\bibitem{xu2019multi}
Qingshan Xu and Wenbing Tao.
\newblock Multi-scale geometric consistency guided multi-view stereo.
\newblock In {\em Proceedings of the IEEE/CVF Conference on Computer Vision and
  Pattern Recognition (CVPR)}, 2019.

\bibitem{xu2020marmvs}
Zhenyu Xu, Yiguang Liu, Xuelei Shi, Ying Wang, and Yunan Zheng.
\newblock Marmvs: Matching ambiguity reduced multiple view stereo for efficient
  large scale scene reconstruction.
\newblock In {\em Proceedings of the IEEE/CVF Conference on Computer Vision and
  Pattern Recognition (CVPR)}, 2020.

\bibitem{yamashita2023nlmvs}
Kohei Yamashita, Yuto Enyo, Shohei Nobuhara, and Ko Nishino.
\newblock nlmvs-net: Deep non-lambertian multi-view stereo.
\newblock In {\em Proceedings of the IEEE/CVF Winter Conference on Applications
  of Computer Vision (WACV)}, 2023.

\bibitem{yao2018mvsnet}
Yao Yao, Zixin Luo, Shiwei Li, Tian Fang, and Long Quan.
\newblock Mvsnet: Depth inference for unstructured multi-view stereo.
\newblock In {\em Proceedings of the European conference on computer vision
  (ECCV)}, 2018.

\bibitem{yao2020blendedmvs}
Yao Yao, Zixin Luo, Shiwei Li, Jingyang Zhang, Yufan Ren, Lei Zhou, Tian Fang,
  and Long Quan.
\newblock Blendedmvs: A large-scale dataset for generalized multi-view stereo
  networks.
\newblock In {\em Proceedings of the IEEE/CVF Conference on Computer Vision and
  Pattern Recognition (CVPR)}, 2020.

\bibitem{volsdf}
Lior Yariv, Jiatao Gu, Yoni Kasten, and Yaron Lipman.
\newblock Volume rendering of neural implicit surfaces.
\newblock {\em Advances in Neural Information Processing Systems (NeurIPS)},
  2021.

\bibitem{IDR}
Lior Yariv, Yoni Kasten, Dror Moran, Meirav Galun, Matan Atzmon, Basri Ronen,
  and Yaron Lipman.
\newblock Multiview neural surface reconstruction by disentangling geometry and
  appearance.
\newblock {\em Advances in Neural Information Processing Systems (NeurIPS)},
  2020.

\bibitem{mvsdf}
Jingyang Zhang, Yao Yao, and Long Quan.
\newblock Learning signed distance field for multi-view surface reconstruction.
\newblock In {\em Proceedings of the IEEE/CVF International Conference on
  Computer Vision (ICCV)}, 2021.

\bibitem{zhang2021physg}
Kai Zhang, Fujun Luan, Qianqian Wang, Kavita Bala, and Noah Snavely.
\newblock Physg: Inverse rendering with spherical gaussians for physics-based
  material editing and relighting.
\newblock In {\em Proceedings of the IEEE/CVF Conference on Computer Vision and
  Pattern Recognition (CVPR)}, 2021.

\bibitem{zhang2020nerf++}
Kai Zhang, Gernot Riegler, Noah Snavely, and Vladlen Koltun.
\newblock Nerf++: Analyzing and improving neural radiance fields.
\newblock {\em arXiv preprint arXiv:2010.07492}, 2020.

\bibitem{zhang2021nerfactor}
Xiuming Zhang, Pratul~P Srinivasan, Boyang Deng, Paul Debevec, William~T
  Freeman, and Jonathan~T Barron.
\newblock Nerfactor: Neural factorization of shape and reflectance under an
  unknown illumination.
\newblock {\em ACM Transactions on Graphics (TOG)}, 2021.

\end{thebibliography}
}

\appendix

\section*{Supplementary
}

\section{Optimization and Additional Model Details}

\noindent \textbf{Optimization Details.} 
We used Adam \cite{kingma2014adam} as our optimizer. For the first 5,000 iterations, the learning rate was linearly increased from 0 to $5 \times 10^{-4}$ using a warm-up strategy. After that, we controlled it using the cosine decay schedule to the minimum learning rate of $2.5 \times 10^{-5}$. We trained each model for 200,000 iterations, which took a total of 7 hours on a single NVIDIA RTX3090Ti GPU.
For the novel view synthesis task, we trained each model for 1,000,000 iterations over 80 hours using a smaller batch size with fewer sampled rays on a single NVIDIA RTX3090Ti GPU. To ensure consistency with the reconstruction baselines, we used single-image batching with 512 sampled rays for all reconstruction tasks. For novel view synthesis, we used single-image batching with 1024 sampled rays, limited to the GPU memory, instead of 4096 $\times$ 4 as used in Ref-NeRF. On each ray, we sampled 64 coarse points, 64 fine points, and 32 points to model the background, as in NeRF++ \cite{zhang2020nerf++}.

\vspace{0.1cm}
\noindent \textbf{Network architecture.} 
Our network architecture is similar to NeuS \cite{wang2021neus}, comprising of a geometry network and a radiance network to encode SDF and view-dependent radiance, respectively. The geometry network parametrizes the signed distance function and consists of 8 hidden layers with a hidden size of 256. Instead of ReLU, we used Softplus with $\beta$ = 100 for all hidden layers. We used a skip connection \cite{mildenhall2021nerf} to connect the input with the output of the fourth layer. The geometry network takes the spatial position $\mathbf{x}$ of points as input and outputs the signed distances to the object. In addition, the geometry network produces a geometry feature with dimension 256, which is further used as input to the radiance network to acquire view-dependent radiance.
The radiance network comprises 4 hidden layers of size 256, which parametrize view-dependent radiance. It takes as input the spatial position $\mathbf{x}$, the normal vector $\mathbf{\hat{n}}$, the reflection direction $\boldsymbol{\omega}_r$, and the 256-dimensional geometry feature vector. We applied positional encoding with 6 frequencies to the spatial location $\mathbf{x}$ and 4 frequencies to the view direction $\boldsymbol{\omega}_r$.
\begin{figure}
    \centering
    \begin{minipage}{0.48\textwidth}
        \includegraphics[width=0.48\columnwidth]{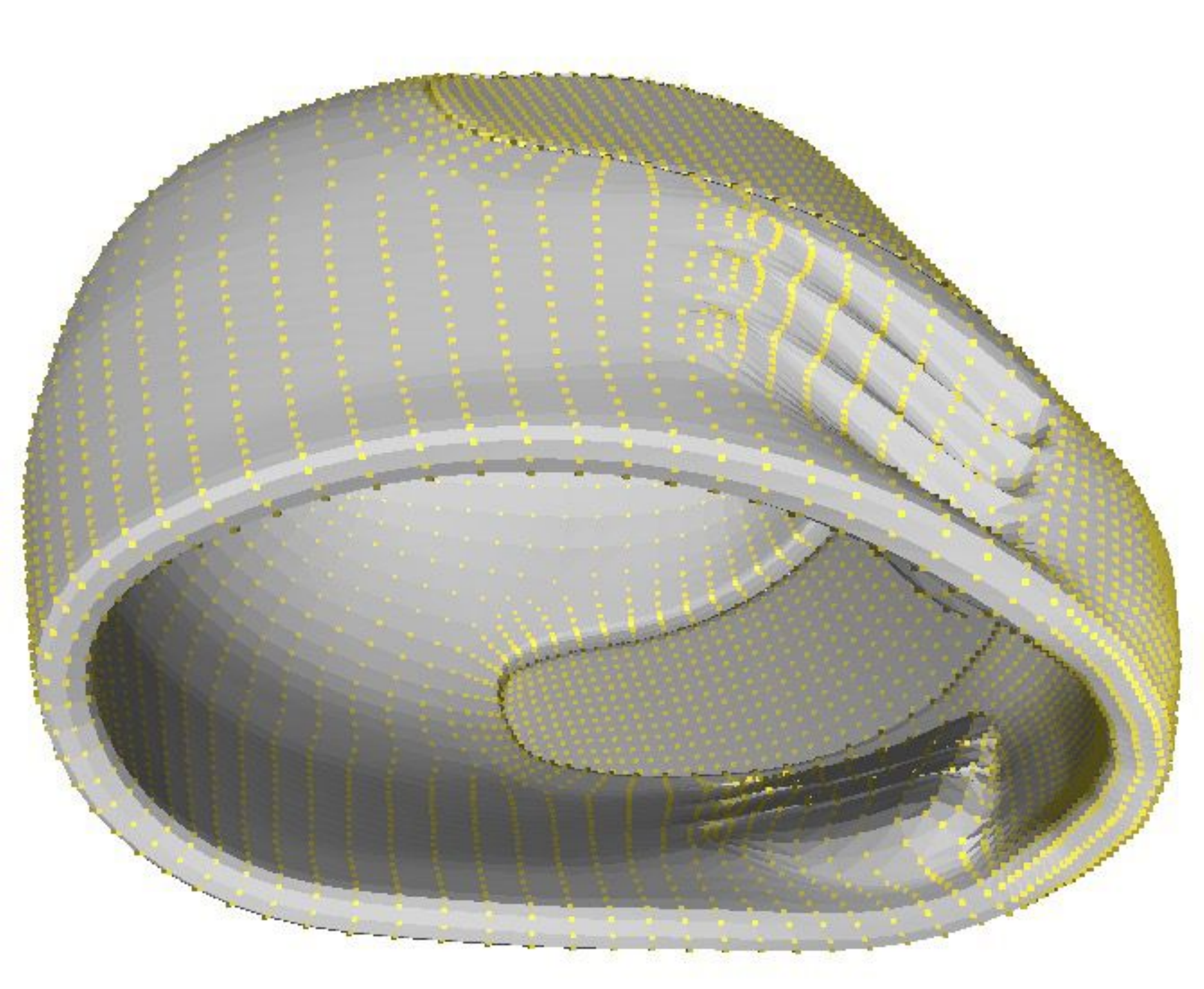}
        \includegraphics[width=0.48\columnwidth]{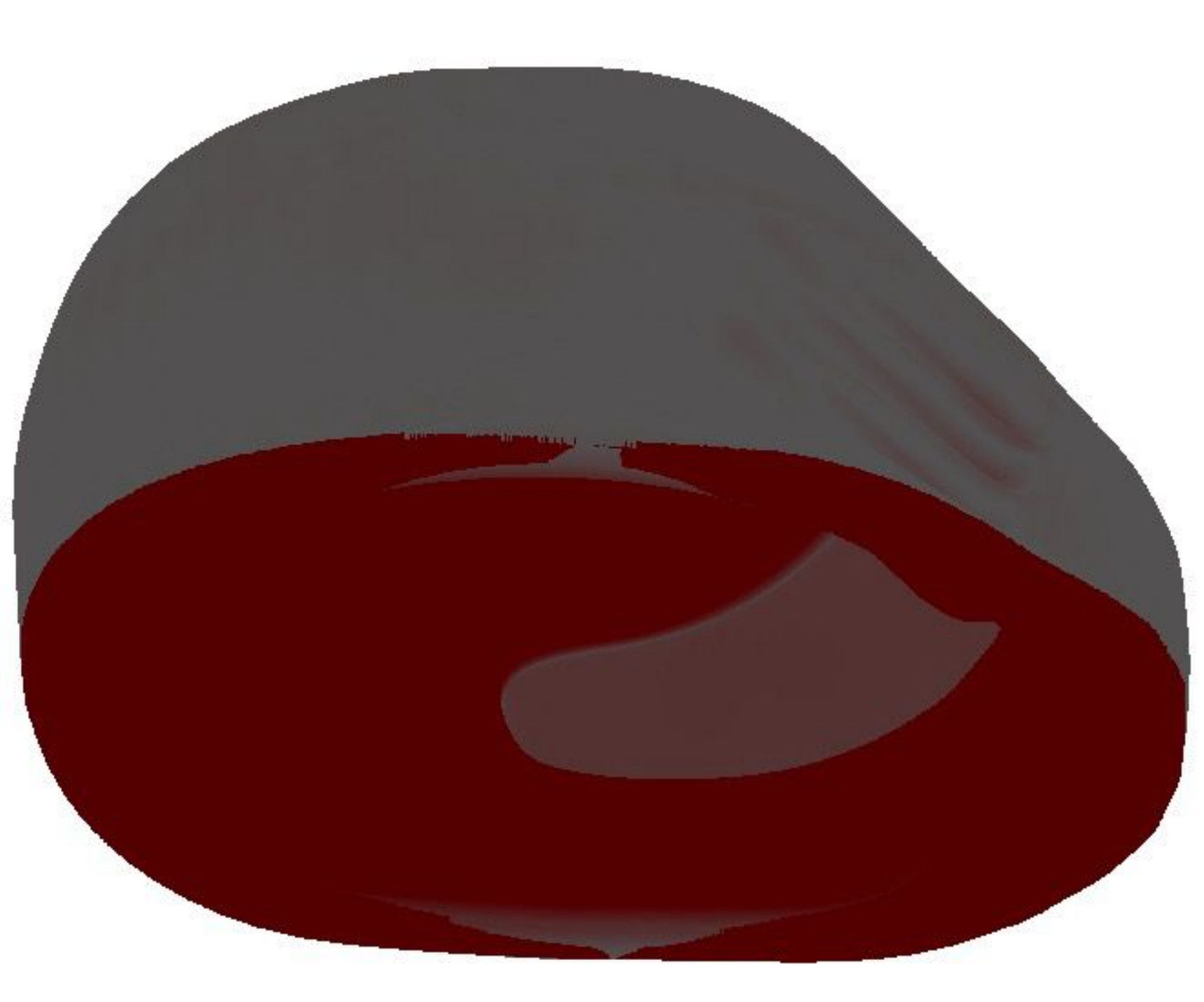}
    \end{minipage}
    \vspace{0.3cm}
    \caption{In the ground truth meshes (left) of ShinyBlender, there are two layers. However, on multi-view images, the inner layer is invisible. This means the completeness is biased, as most of the points in the inner layer contribute to meaningless error (right). Points with large errors are marked in red.}
    \label{shiny}
\end{figure}

\begin{figure}
\centering
   \begin{minipage}{0.49\textwidth}
    \centering
     \includegraphics[width=0.48\columnwidth]{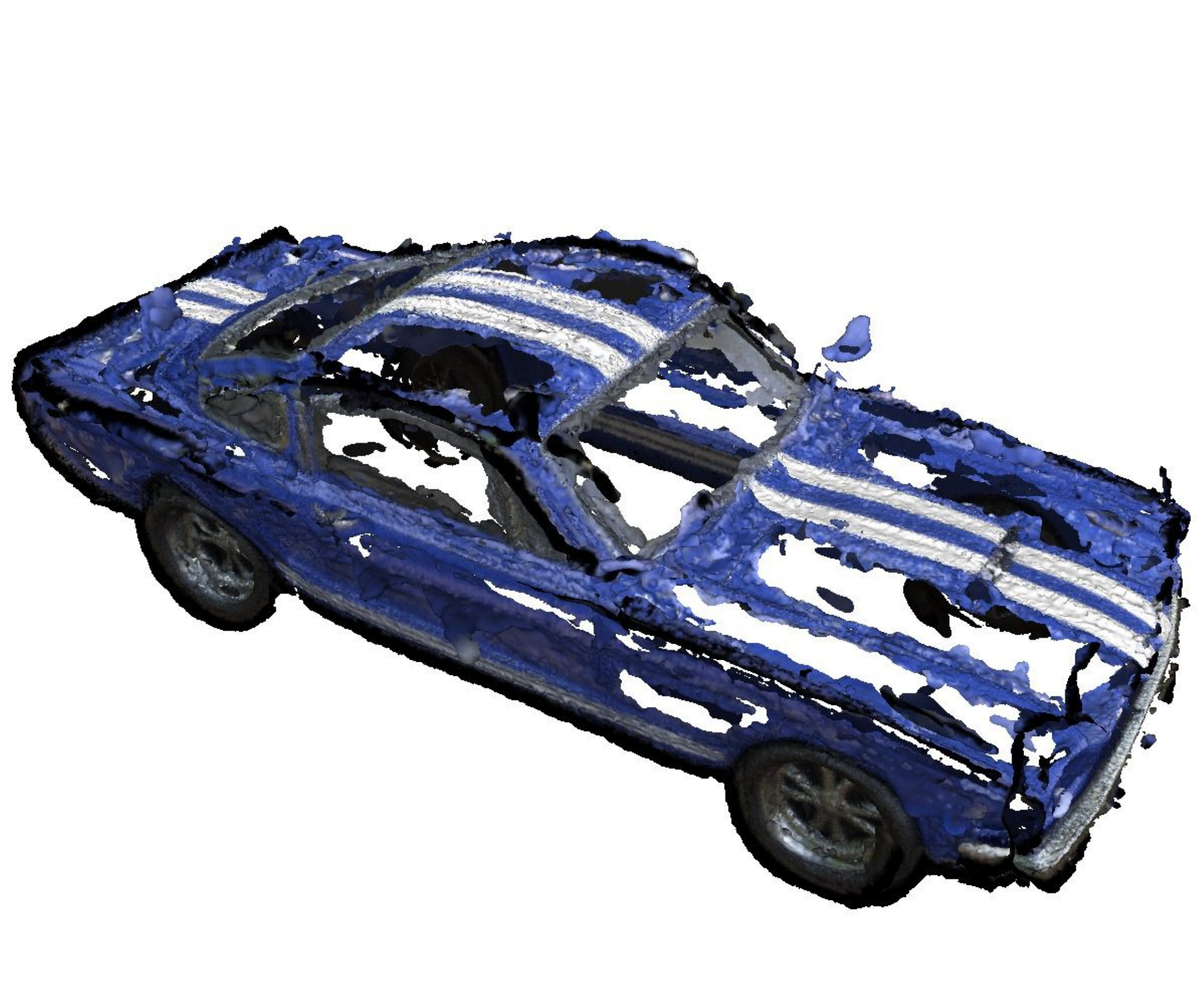}
    \includegraphics[width=0.48\columnwidth]{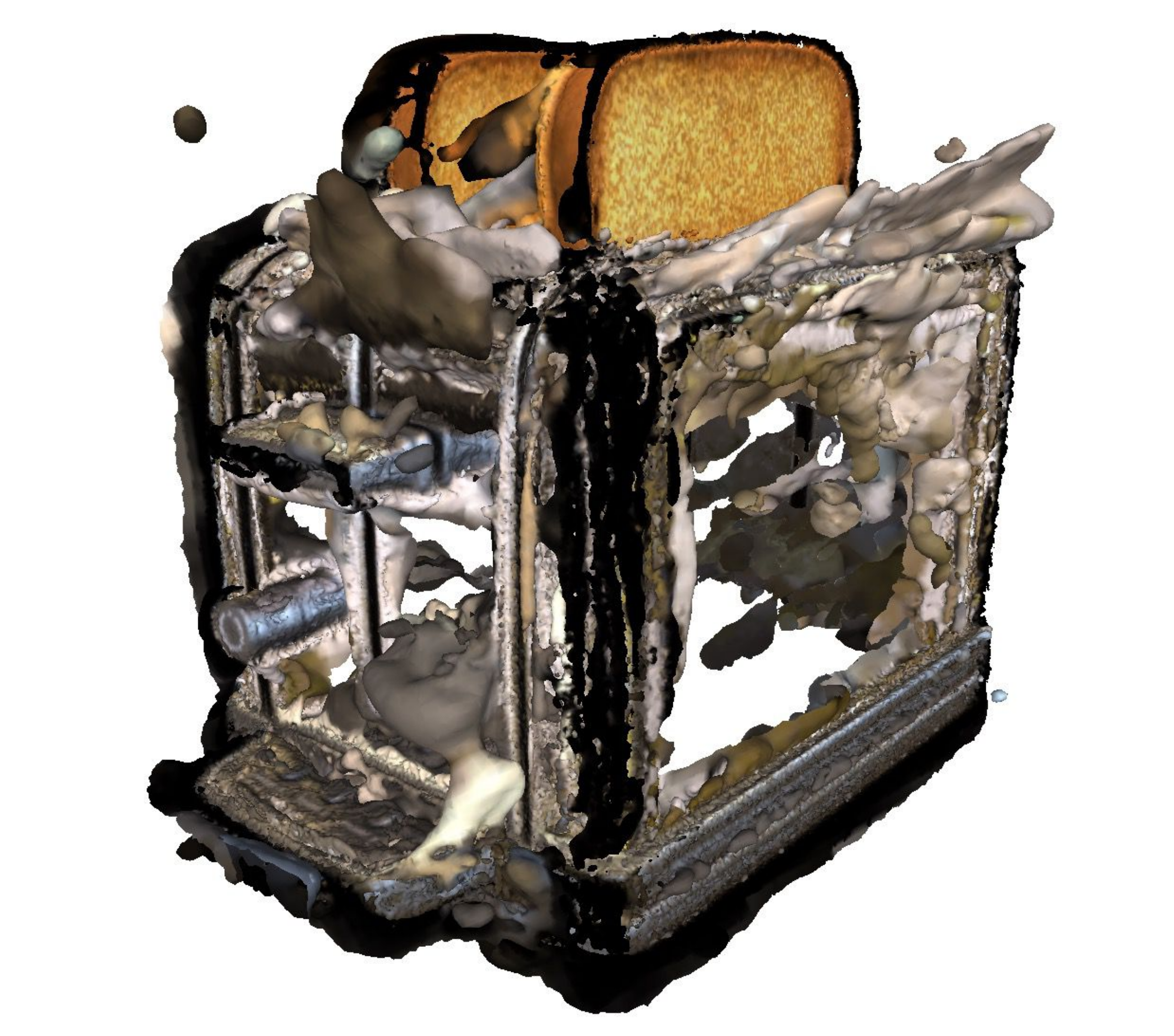}
    
    \includegraphics[width=0.48\columnwidth]{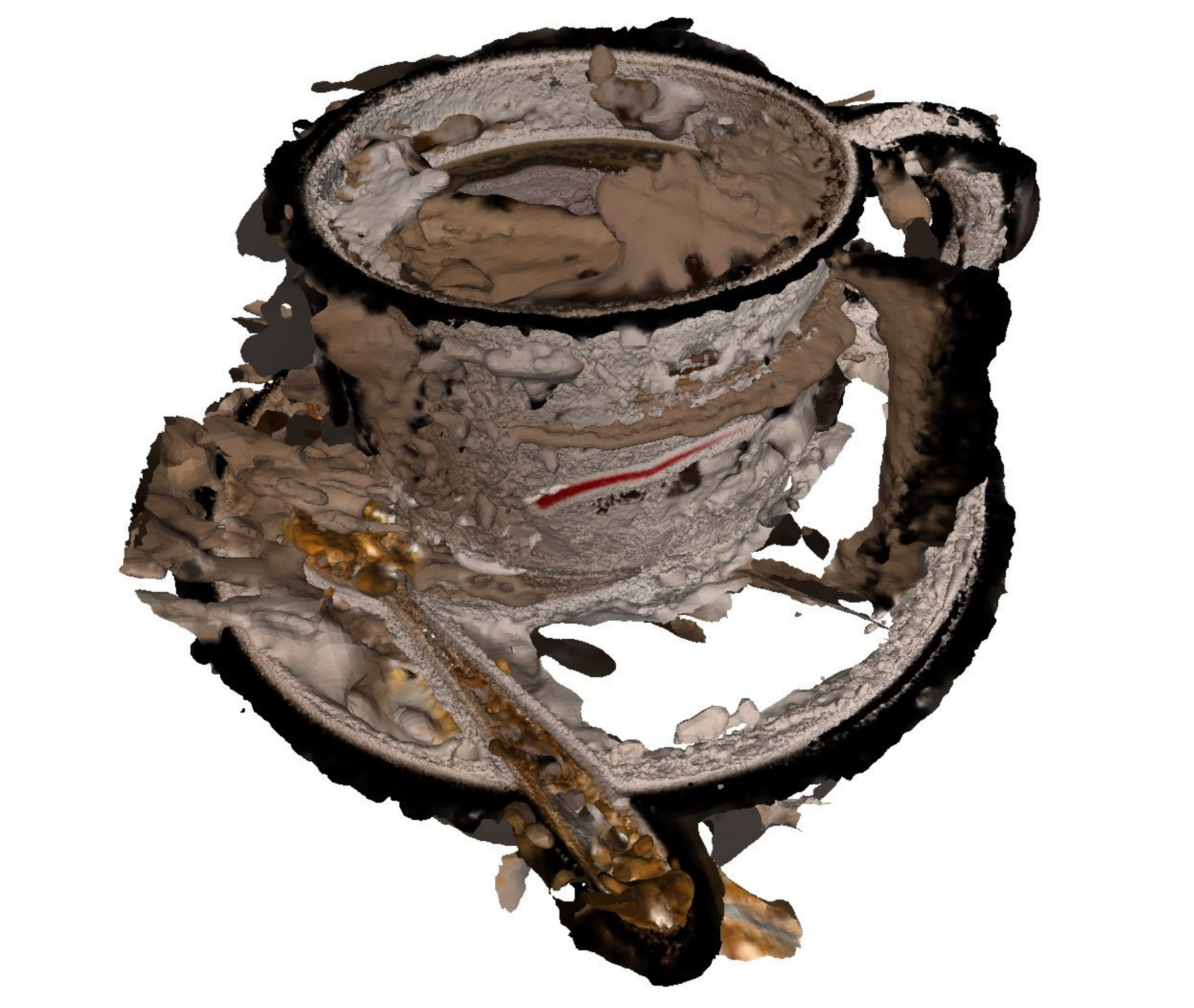}
    \includegraphics[width=0.48\columnwidth]{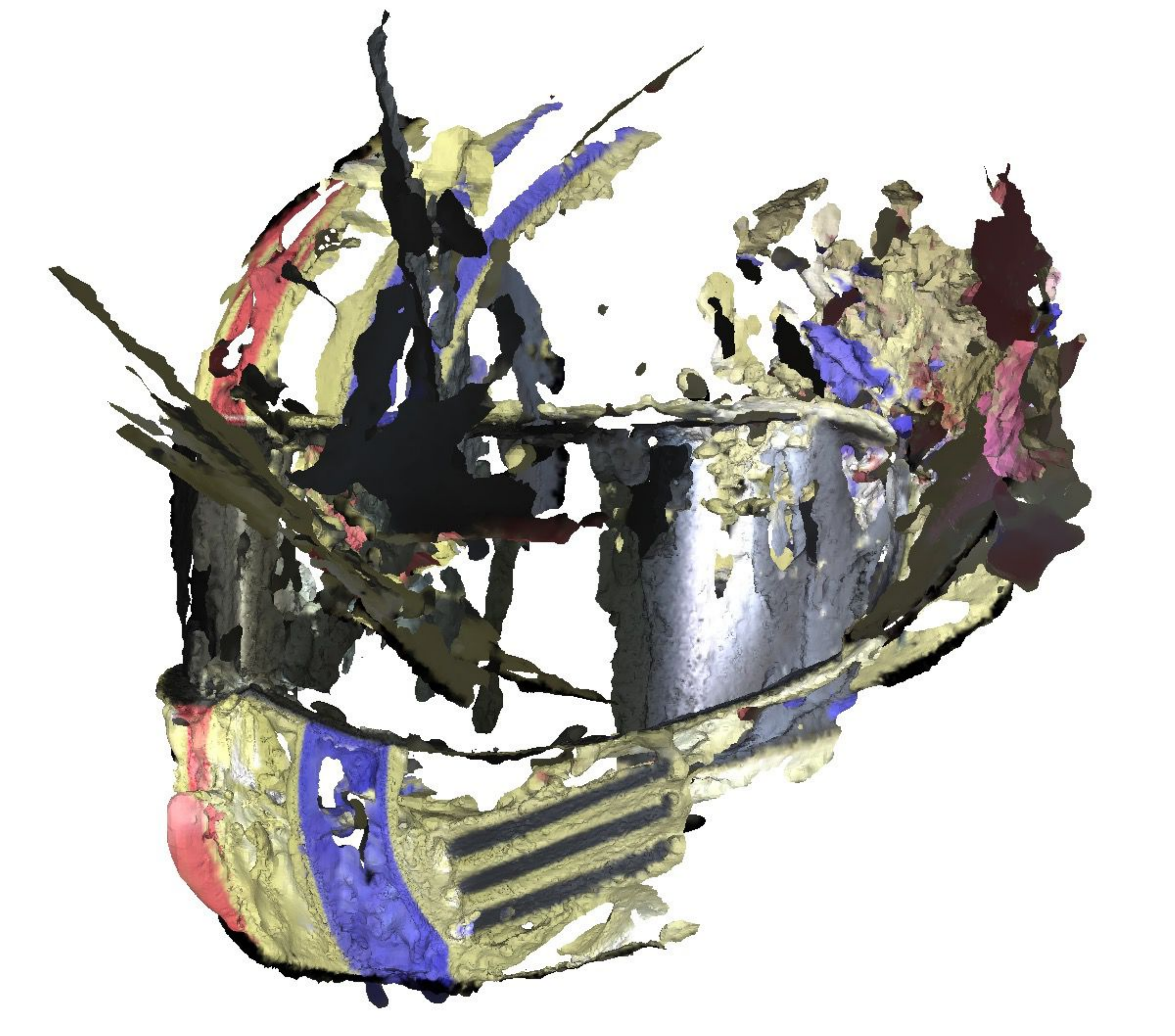}

    \end{minipage}
    \vspace{0.3cm}
    \caption{Visualization of the reconstruction results by COLMAP on the ShinyBlender dataset. COLMAP failed to reconstruct objects with reflective surfaces, as the multi-view consistency was not plausible.}
    \label{colmap}

\end{figure}

\begin{figure*}[htbp]
    \centering
    \begin{minipage}{\textwidth}
    \centering
     \includegraphics[width=0.19\columnwidth]{figure/r_58.pdf} 
    \includegraphics[width=0.19\columnwidth]{figure/r_58_normal-removebg-preview.pdf}
    \includegraphics[width=0.19\columnwidth]{figure/00000058_0-neus.pdf}
    \includegraphics[width=0.19\columnwidth]{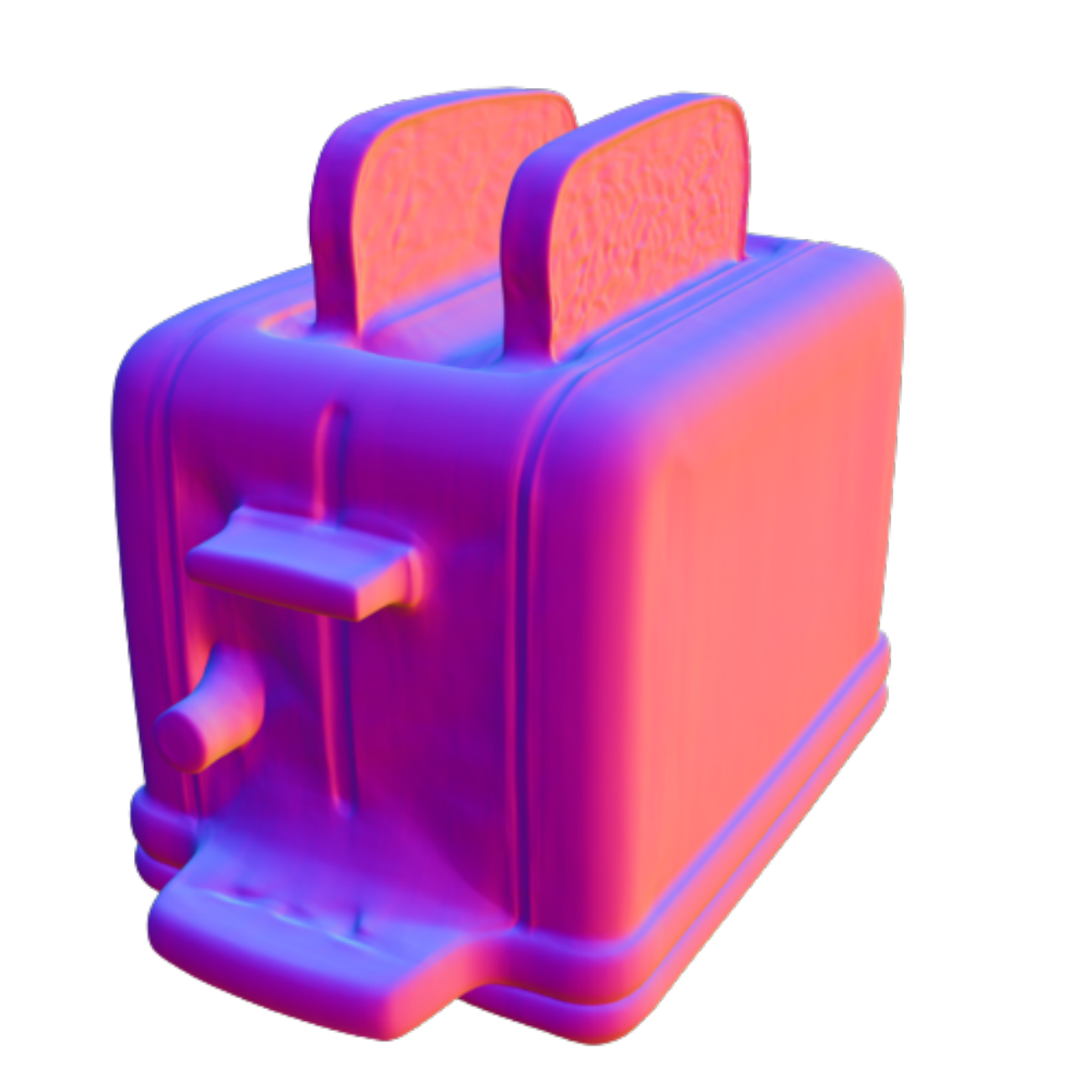}
    \includegraphics[width=0.19\columnwidth]{figure/00140000_0_58-removebg-preview.pdf}
    \end{minipage}
    \centering
    \begin{minipage}{0.19\textwidth}
    \includegraphics[width=\columnwidth]{figure/r_58.pdf}
    \centerline{\scriptsize Reference Image}
    \end{minipage}
    \begin{minipage}{0.19\textwidth}
    \includegraphics[width=\columnwidth]{figure/58_mesh_gt.pdf}
    \centerline{\scriptsize GT Meshes}
    \end{minipage}
    \begin{minipage}{0.19\textwidth}
    \includegraphics[width=\columnwidth]{figure/83_mesh_neus.pdf}
    \centerline{\scriptsize NeuS}
    \end{minipage}
    \begin{minipage}{0.19\textwidth}
    \includegraphics[width=\columnwidth]{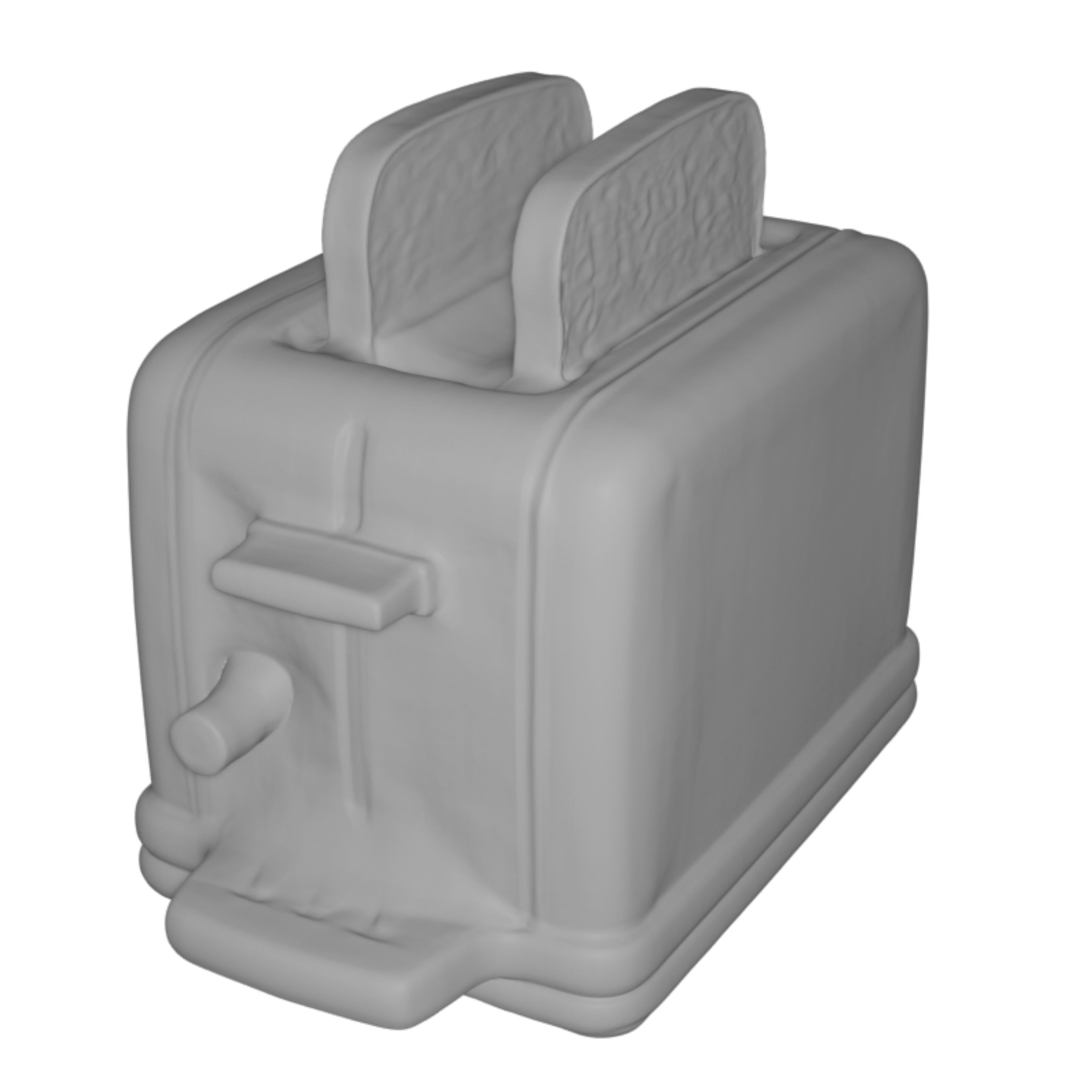}
    \centerline{\scriptsize Ref-NeuS w/o Ref}
    \end{minipage}
    \begin{minipage}{0.19\textwidth}
    \includegraphics[width=\columnwidth]{figure/toaster_mesh_ours.pdf}
    \centerline{\scriptsize Ref-NeuS}
    \end{minipage}

     \vspace{0.3cm}
    \caption{Surface geometry and surface normals of ablation models on toaster of ShinyBlender.}
    \label{ablation_vis}

\end{figure*}

\section{ Evaluation Details}

\noindent \textbf{Meshes.} For the ShinyBlender \cite{verbin2022ref} and Blender \cite{mildenhall2021nerf} datasets, the ground truth meshes were exported from Blender files. Due to the original models' small scales with a radius around 1, we exported them with a scale factor of 150. For the fish from SLF \cite{wood2000surface} and the cans/corncho1 from Bag of Chips \cite{park2020seeing}, we increased the meshes' sizes by 100 and 1000 times, respectively, resulting in similar scales for all ground truth meshes. During training, we normalized the object to a unit sphere. During inference, we transferred the meshes to the original space to compute the Chamfer Distance.

Since the original meshes contain too few points, we upsampled the points in each triangle to obtain dense point clouds for evaluation. Finally, the Chamfer Distance was computed by
\begin{equation}\small
     d (  S_ {1}  ,  S_ {2})=  \frac {1}{S_ {1}}  \sum _ {x\in S_1} {\min_{y \in S_2} ||x-y||_ {2}^ {2}+\frac {1}{S_ {2}}}  \sum _ {y\in S_ {2}}  \min_{x \in S_1}  ||y-  x||_ {2}^ {2},
\end{equation}
where the first term is used to test \emph{accuracy}, and the second term validates \emph{completeness} \cite{dtu_eval}. $S_1$ and $S_2$ are the recovered point clouds upsampled from meshes and ground truth dense point clouds, respectively. For the ShinyBlender dataset, shown in Fig. \ref{shiny}, the ground truth dense point clouds include two layers. However, the inner layer is invisible on multi-view images and cannot be reconstructed, resulting in a biased completeness, so only accuracy is reported.

\noindent \textbf{Surface Normals.}
To compute the surface normal for a pixel $p$, we compute the normals of sampled points along the ray $\textbf{r}$ derived from the SDF as follows:
\begin{equation}
    \mathbf{\hat{n}_i} = \frac{\nabla f(\mathbf{x_i})}{||\nabla f(\mathbf{x_i})||}.
\end{equation}
Then, the volume rendering procedure is performed to aggregate these normals, forming a single surface normal:
\begin{equation}
    \hat{\textbf{n}}(\mathbf{r})=\sum_{i=1}^N T_i \alpha_i \mathbf{\hat{n}}_i.
\end{equation}
We used the normalized normals $\overline{\textbf{n}}(\mathbf{r}) = \frac{\hat{\textbf{n}}(\mathbf{r})}{||\hat{\textbf{n}}(\mathbf{r})||}$ for evaluating MAE for all pixels.

\begin{table*}[t]  
    \centering
    \small
    \begin{tabular}{c|ccc|cc|cc|cc|cc||cc}
        \hline
        \multirow{2}*{Method} &\multirow{2}*{RS}  &\multirow{2}*{Vis} &\multirow{2}*{Ref} &\multicolumn{2}{c|}{helmet} &\multicolumn{2}{c|}{toaster} &\multicolumn{2}{c|}{coffee} &\multicolumn{2}{c||}{car} &\multicolumn{2}{c}{mean}\\ \cline{5-14}
         &&& &Acc &MAE &Acc &MAE &Acc &MAE &Acc &MAE &Acc &MAE\\ \hline
        NeuS & &  & &1.33 &1.12 &3.26 &2.87 &1.42 &1.99 &0.73 &1.10 &1.69 & 1.77 \\ 
        
        NeuS w/ RS &\checkmark & & &0.75 &0.85 &2.14 &2.23 &1.11 &1.58 &0.66 &1.05 &1.17 &1.43 \\ 
         NeuS w/ Ref & &   &\checkmark &0.41 &0.69  &0.59 &1.59 &3.87  &2.74 &0.55 & 0.97 &1.36 &1.50 \\
         Ref-NeuS w/o Ref&\checkmark  &\checkmark  & &0.43 &0.71  &1.43 &2.12 &0.77  &0.99 &0.58 & 1.00 &0.80 &1.21 \\
        Ours (full) &\checkmark  &\checkmark  &\checkmark &\textbf{0.29} &\textbf{0.38} &\textbf{0.42} &\textbf{1.47}&\textbf{0.77} &\textbf{0.99}&\textbf{0.37} &\textbf{0.80}&\textbf{0.46} &\textbf{0.91} \\ \hline
    
    \end{tabular}
    \vspace{0.1cm}
     \caption{Detailed quantitative metrics of ablation study on Shiny Blender dataset.}
    \label{ablationt}

\end{table*}

\section{Results of using training views}
In the main text, we used original 200 test views for  reconstruction training, here we show the results of using 100 training views for reconstruction training in Table \ref{training_view}. The conclusion is similar to that obtained using the test views.

\begin{table*}[ht]
    \small
    \centering
    \begin{tabular}{c|cc|cc|cc}
         \hline
           \multirow{2}*{Methods}& \multicolumn{2}{c|}{helmet}&   \multicolumn{2}{c|}{toaster}   &\multicolumn{2}{c}{car} \\ \cline{2-7}
             &Acc$\downarrow$   &MAE$\downarrow$ &Acc$\downarrow$   &MAE$\downarrow$  &Acc$\downarrow$  &MAE$\downarrow$  \\  \hline\hline

             Neus &0.92 &0.88 &3.34 &2.73 &0.72 &1.08  \\
             Ref-NeuS &0.33 &0.39 &0.45 &1.56 &0.36  &0.77  \\ \hline
    \end{tabular}
    \caption{Results of using training views for reconstruction training.}
    \label{training_view}
\end{table*}

\section{Detailed Results of Ablation Study}
We reported the quantitative metrics (accuracy and MAE) for each scene of ShinyBlender in Table \ref{ablationt}.

\section{Additional Results on diffuse materials}

We also carried out experiments on non-reflective objects to show that reflection score will not cause performance degradation of non-reflective objects, where DTU scenes (i.e., scan55, scan83, scan105, scan106, scan114, scan118) were used. The results are reported in Table \ref{dtu_diffuse}.

\begin{table}[ht]
    \small
    \centering
    \begin{tabular}{c|c|c|c|c|c|c}
    \hline
         scene & 55  & 83 &105 &106 &114 &118  \\ \hline
         NeuS &0.37  &1.45 &0.78 &0.52 &0.36 &0.45  \\ \hline
         NeuS w/ RS &0.36  &1.27 &0.72 &0.51 &0.36 &0.46  \\ \hline
    \end{tabular}
    \caption{The results on non-reflective objects.}
    \label{dtu_diffuse}
\end{table}

\section{Additional Results on scenes with both diffuse and shiny materials}

\begin{figure*}[ht]
    \centering
    \includegraphics[width=0.98\textwidth]{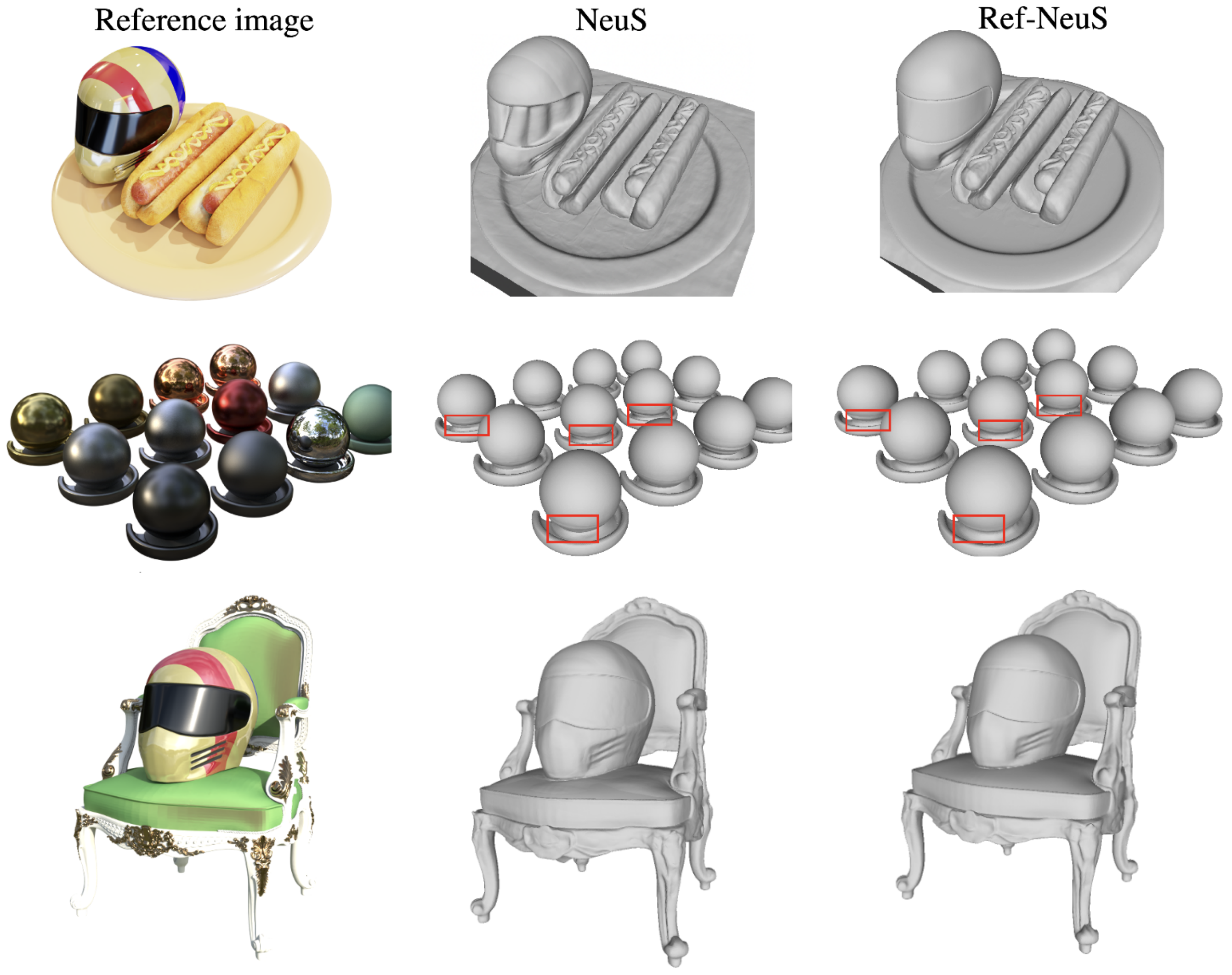}
    \caption{ Comparison on scenes with both diffuse and shiny materials.}
    \label{hel_hot}
\end{figure*}

Previous reconstructed scenes are either shiny or diffuse materials. To show the robustness of our method, we further diverse the scenes with both diffuse and shiny materials. Given that such scenes are uncommon in existing datasets, besides materials for Blender dataset, we further employed Blender to render multi-view images that combine helmet for Shinyblender and hotdog from Blender. The comparison is presented in Fig. \ref{hel_hot}. Our method can reconstruct the shiny objects better, while do not lead to performance drop on diffuse materials.

\section{Additional Visualizations}

\vspace{0.1cm}
\noindent \textbf{Visualizations of COLMAP. }
We visualized the reconstruction results of COLMAP \cite{schonberger2016pixelwise}, an MVS-based method on the Shiny Blender dataset, as shown in Fig. \ref{colmap}. COLMAP fails to recover reflective surfaces, indicating that the multi-view consistency is not reasonable in reflective scenes, leading to severe missing parts and artifacts.

\vspace{0.1cm}
\noindent \textbf{Visualizations of ablation study. }
We visualized how the reflection-aware photometric loss and reflection direction-dependent radiance improve surface quality in Fig. \ref{ablation_vis}. Due to the reflective surfaces, the toaster is extremely challenging even for human perception. It is challenging to distinguish where the real surface lies.
NeuS reconstructs the toaster with severe missing parts due to reflection. ``Ref-Neus w/o Ref'' reconstructs the surface with fewer missing parts, indicating that reflection-aware photometric loss can localize the reflective surfaces and alleviate the ambiguity. Our full model, Ref-NeuS, achieves better reconstruction results without missing parts. 

\begin{figure*}
    \centering
    \includegraphics[width=0.98\textwidth]{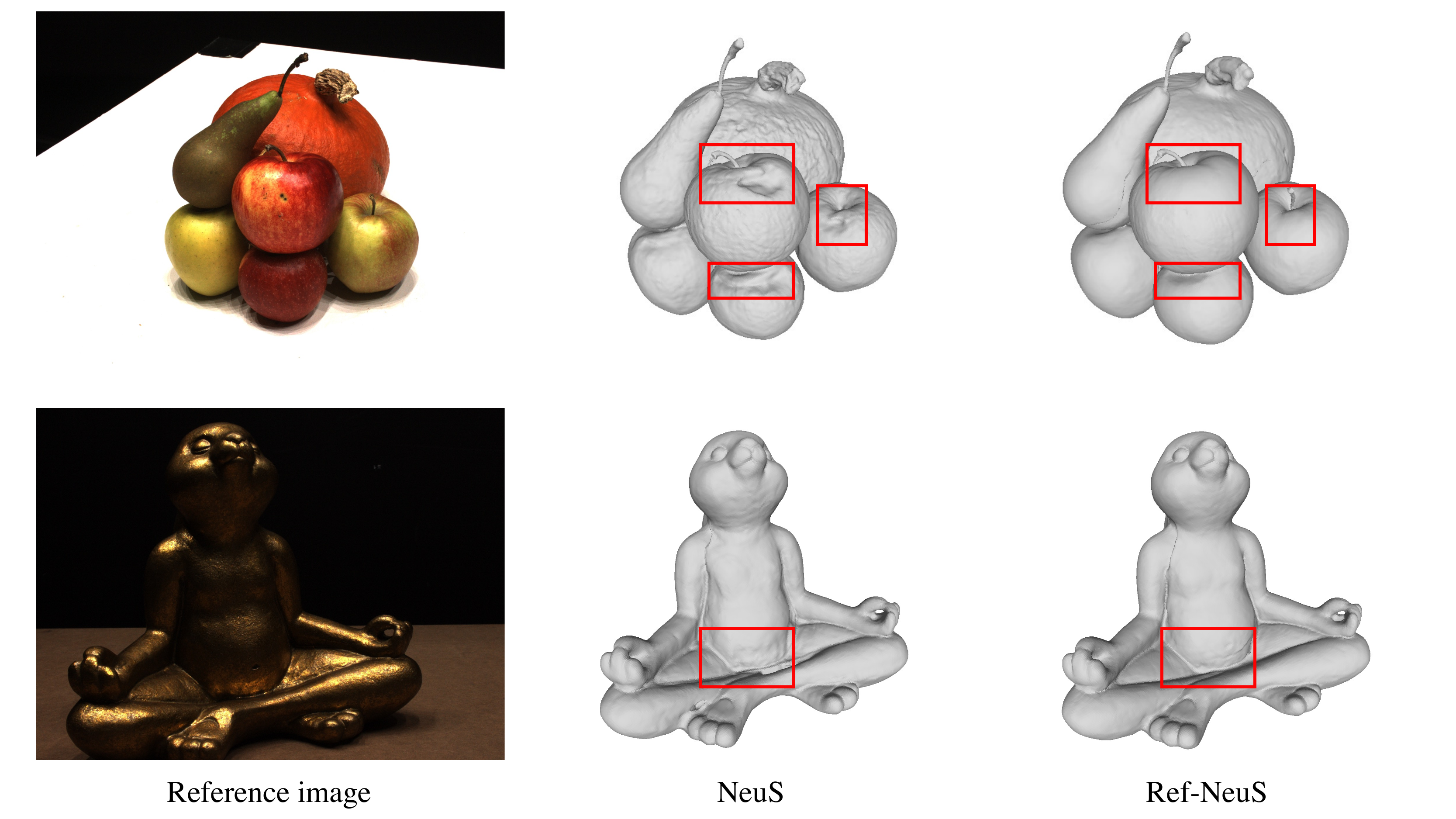}
    \caption{Qualitative comparison with NeuS on scan63 and scan 110 of DTU dataset.}
    \label{dtu}
\end{figure*}

\begin{figure*}
    \centering
    \includegraphics[width=0.95\textwidth]{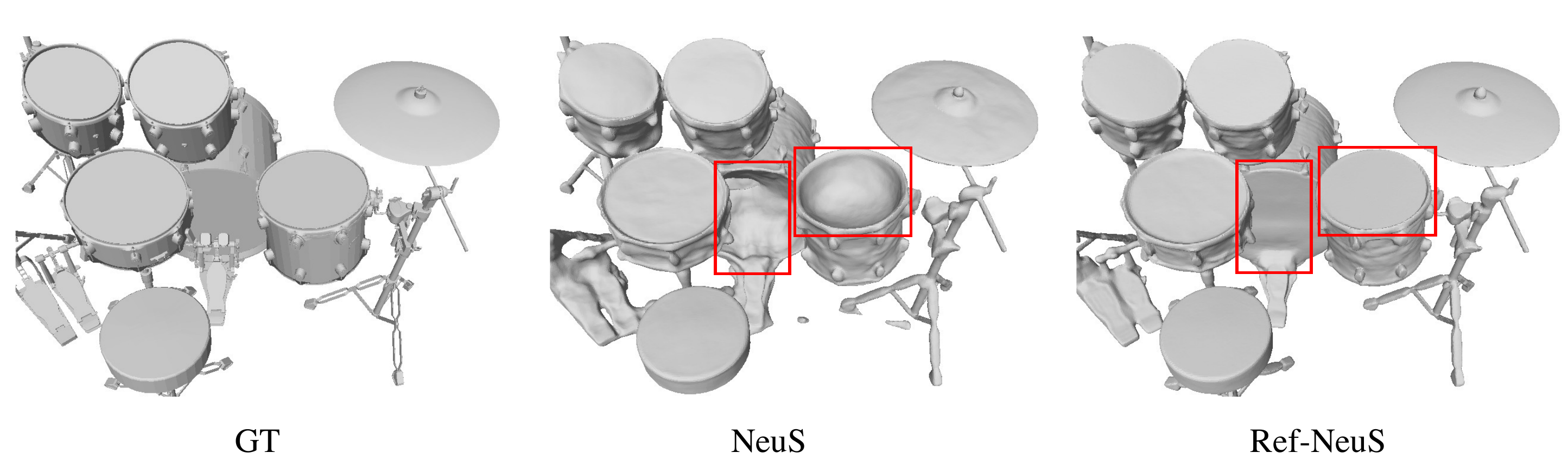}
    \caption{Qualitative comparison with NeuS on drums of Blender dataset.}
    \label{drums}
\end{figure*}

\begin{figure*}[htbp]
    \centering
    \includegraphics[width=0.9\textwidth]{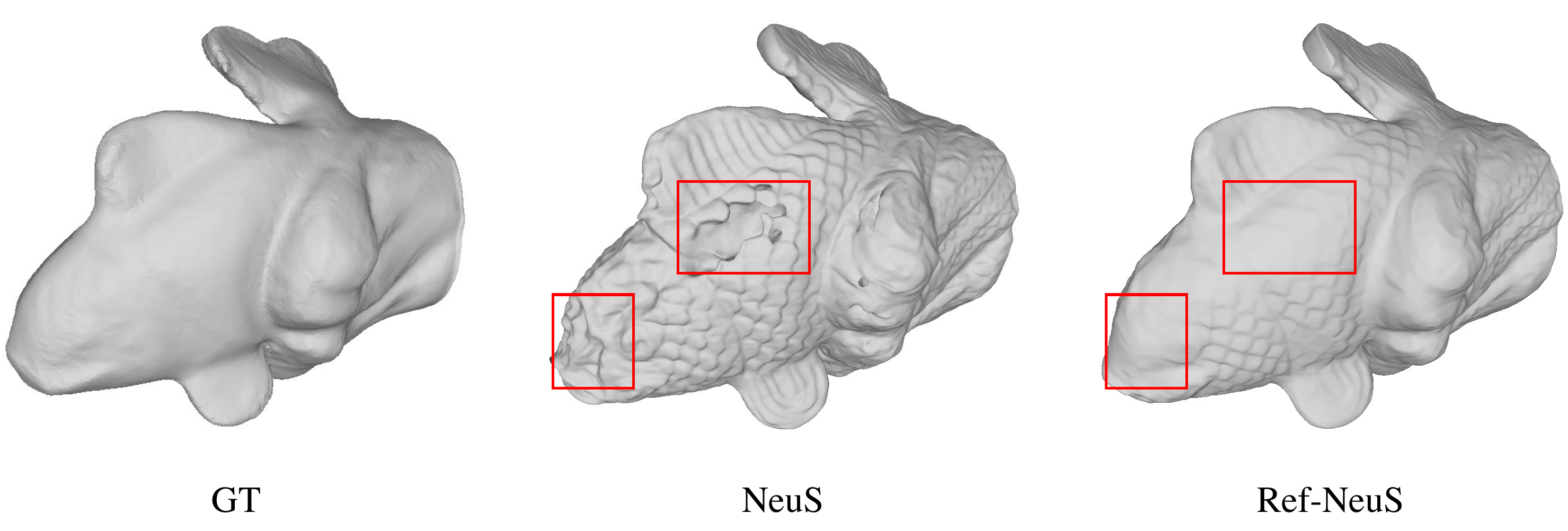}
    \caption{ Qualitative comparison with NeuS on fish of SLF dataset.}
    \label{fish}
\end{figure*}

\begin{figure*}[htbp]
    \centering
    \includegraphics[width=0.9\textwidth]{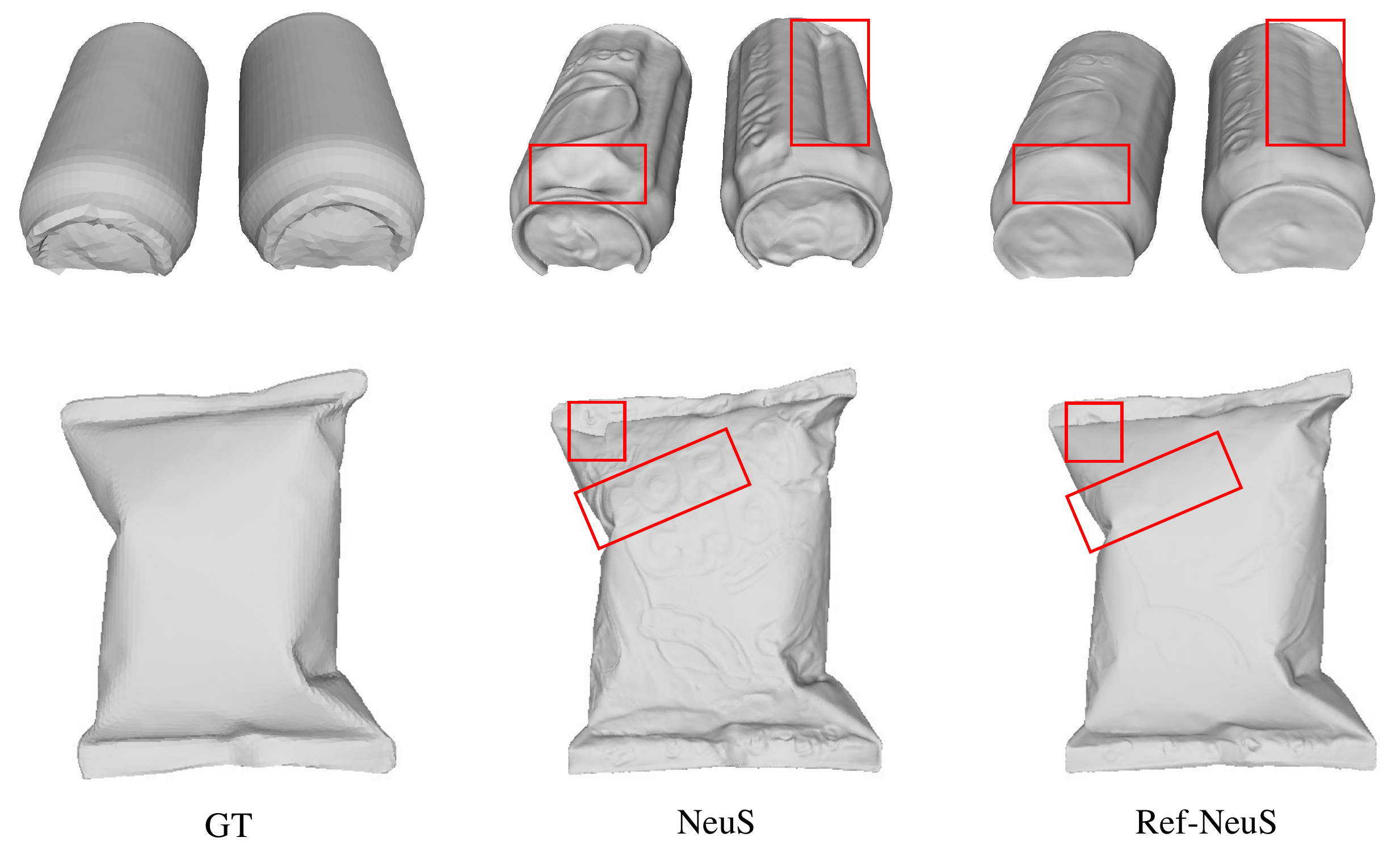}
    \caption{ Qualitative comparison with NeuS on cans and corncho1 of Bag of Chips dataset.}
    \label{chip}
\end{figure*}

\vspace{0.1cm}
\noindent \textbf{Visualizations of Ref-NeuS.  }
We present additional visualization results of different objects in Fig. \ref{dtu}, Fig. \ref{drums}, Fig. \ref{fish}, and Fig. \ref{chip} to demonstrate the effectiveness of our Ref-NeuS. Our Ref-NeuS achieves better reconstruction quality compared to NeuS.

\begin{figure*}[htbp]
    \centering
    \includegraphics[width=0.9\textwidth]{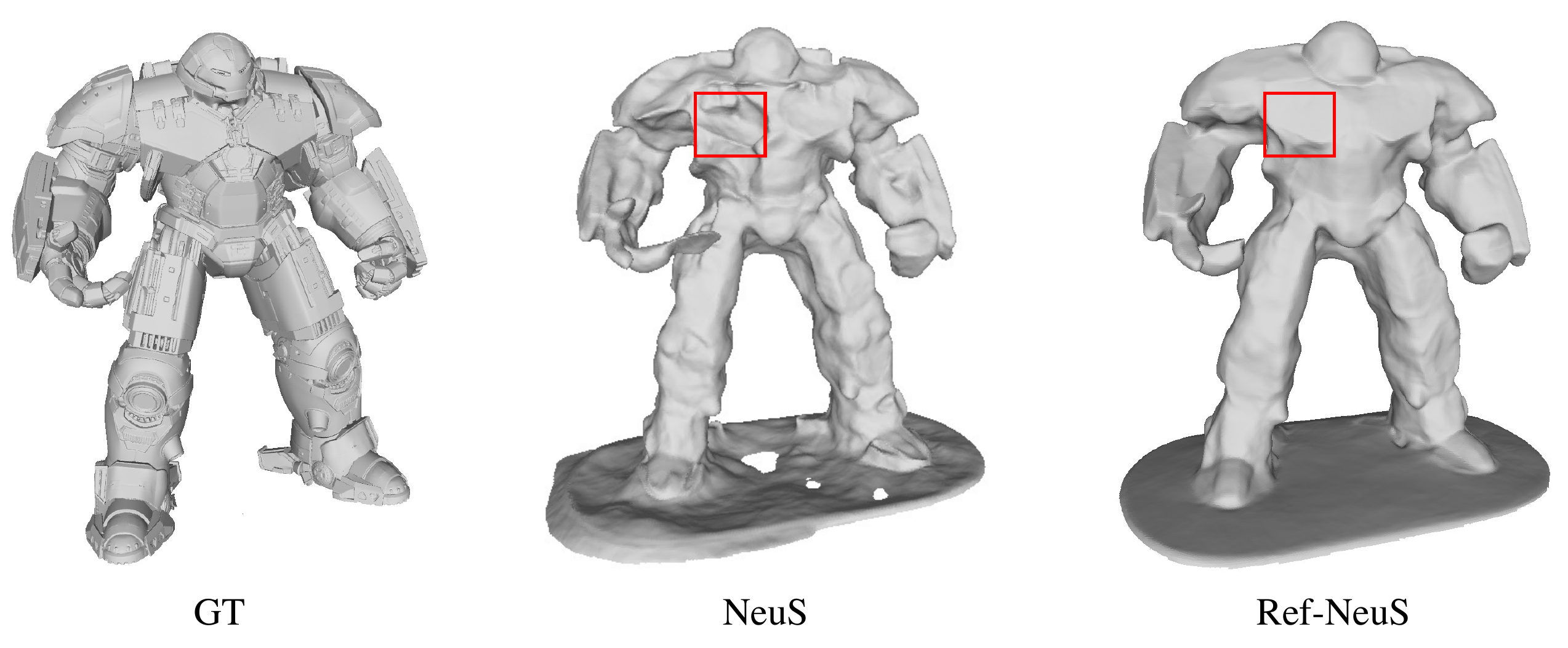}
    \caption{ Qualitative comparison with NeuS on Hulk.}
    \label{hulk}
\end{figure*}

\vspace{0.1cm}
\noindent \textbf{Visualizations of results on Hulk. } 
The real-world objects used in our experiments were captured under strict conditions in a lab-controlled environment. Differently, we captured the Hulk with glossy surfaces using an iPad in a natural environment, capturing both the object and its surroundings with lighting illumination and ambient light. As we captured the object with the iPad moving around it, the light source may have been occluded, resulting in shadows on the surfaces. We show the results in Fig. \ref{hulk}, we can still achieve better performance than NeuS.

\section{Running Time}
We demonstrate that estimating the reflection score will not significantly increase the running time. There are two main steps that contribute to the increase in running time. The first step involves the intermediate meshes. Since we extract intermediate meshes with a resolution of 128, it only takes approximately 0.35 seconds for each mesh extraction. The second step involves projection and distance computation for visibility identification. To obtain pixel colors, we project the predicted surface point onto visible source images. This step does not incur notable extra computational cost, with only 0.012 seconds per step. In Table \ref{runtime}, we present the total running time.

\begin{table}[ht]
    \centering
    \begin{tabular}{c|c}
        setting &Running time [h]  \\ \hline
        NeuS &7 \\ 
        Ref-NeuS &7.5
        
    \end{tabular}
    \vspace{0.1cm}
    \caption{Comparison of running time.}
    \label{runtime}
\end{table}

\begin{figure*}[htbp]
    \centering
    \includegraphics[width=0.9\textwidth]{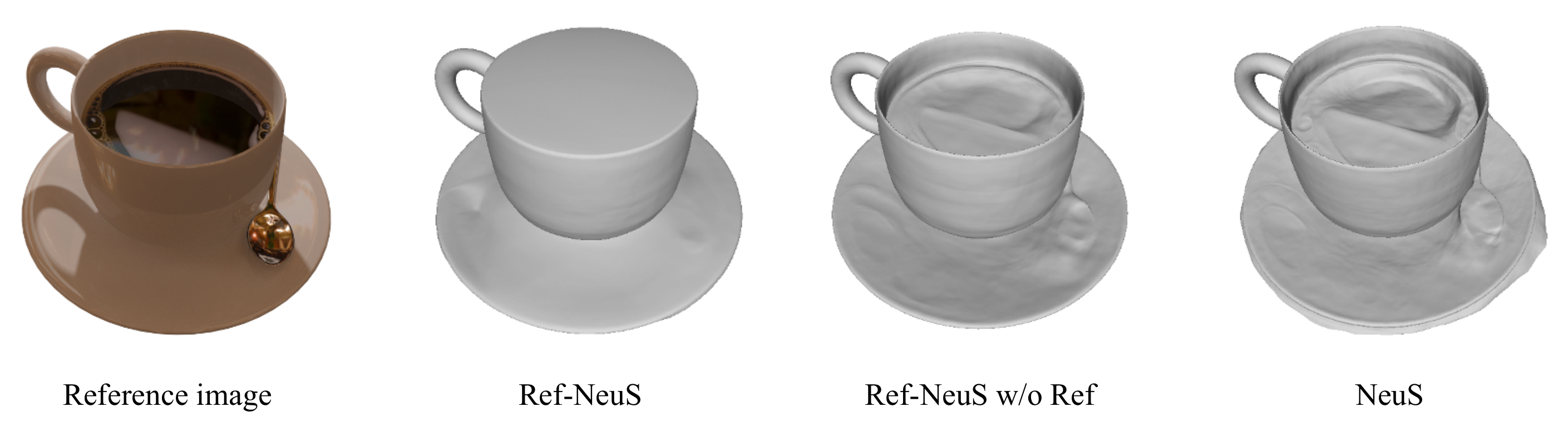}
    \caption{A failure case on coffee of ShinyBlender.}
    \label{coffee}
\end{figure*}

\section{Failure Case}
Figure \ref{coffee} displays the reconstruction results of the coffee object using ShinyBlender, which is a failure case of our method. This object contains water surfaces that possess different reflection coefficients compared to solid objects. Merely substituting the dependency of the radiance network with reflection direction, without considering the object material, can result in artifacts. This motivates future work on how to better model view-dependent radiance while taking the material into consideration. However, incorporating reflection-aware photometric loss can still improves the reconstruction quality over NeuS. We present the results of ``Ref-NeuS w/o Ref'' for Ref-NeuS on the coffee object of ShinyBlender.

\end{document}